\pdfoutput=1

\documentclass[11pt]{article}

\usepackage{acl}

\usepackage{times}
\usepackage{latexsym}

\usepackage[T1]{fontenc}

\usepackage[utf8]{inputenc}

\usepackage{microtype}

\usepackage{inconsolata}

\usepackage{amsmath}
\usepackage{amsfonts}
\usepackage{graphicx}
\usepackage{subcaption}
\usepackage{tikz}
\usepackage{multirow}
\usepackage{float}
\usepackage{booktabs}
\usepackage{tcolorbox}
\usepackage{multicol}
\usepackage{algorithm}
\usepackage{algpseudocode}
\usepackage{soul}
\usepackage{array}
\usepackage{arydshln}
\usepackage{bbm}
\usetikzlibrary{bayesnet}
\usepackage[table]{xcolor}
\usepackage{colortbl}
\usepackage[edges]{forest}
\usepackage{pifont}
\usepackage{makecell}
\usepackage{tabularx}
\usepackage{hyperref}
\usepackage{fontawesome5}
\usepackage{array}
\usepackage{booktabs}
\usepackage{makecell}
\usepackage{multirow}
\usepackage[table]{xcolor}
\usepackage{colortbl}

\definecolor{benchbg}{RGB}{235,242,255}
\definecolor{modelbg}{RGB}{235,247,237}
\definecolor{rulegray}{RGB}{95,95,95}
\definecolor{textgray}{RGB}{90,90,90}

\newcommand{\probetarget}[1]{\textit{\textcolor{textgray}{#1}}}

\newcolumntype{Y}{>{\raggedright\arraybackslash}X}

\title{VLA-Trace: Diagnosing Vision-Language-Action Models Through Representation and Behavior Tracing}

\newcommand{\pifive}{\pi_{0.5}}

\author{
\bfseries
Haoyuan Shi$^{1,2}$\thanks{Core contributors.}\thanks{Work done during the internship at X-Humanoid.},
Xiancong Ren$^{2}$\footnotemark[1]\thanks{Project leader.},
Yingji Zhang$^{2,3}$\footnotemark[1]\footnotemark[2],
Qinfan Zhang$^{2,4}$\footnotemark[1]\footnotemark[2], \\
\bfseries
Jiayu Hu$^{2}$,
Haozhe Shan$^{2,5}$,
Han Dong$^{2,6}$,
Jinpeng Lu$^{1,2}$, \\
\bfseries
Yinda Chen$^{1,2}$,
Yi Zhang$^{2}$,
Yong Dai$^{2}$,
Xiaozhu Ju$^{2}$\thanks{Corresponding author.} \\
$^{1}$ University of Science and Technology of China, $^{2}$ X-Humanoid \\
$^{3}$ Harbin Engineering University, $^{4}$ Beihang University \\
$^{5}$ Fudan University, $^{6}$ University of New South Wales \\
\href{https://vla-trace.github.io/}{
Project Page}
\hspace{1em}
\href{https://github.com/VLA-Trace/VLA-Trace}{
Github Code}
}

\begin{document}
\maketitle

\begin{abstract}
Understanding how Vision-Language-Action (VLA) models transform multimodal knowledge into embodied control remains an open challenge. 
We present VLA-Trace, a progressive diagnostic framework that analyzes VLA models through a unified evidence chain from representation dynamics to causal control attribution and behavioral manifestation. It specifically combines cross-modal and checkpoint-drift centered kernel alignment (CKA) to trace representation evolution, attention knockout interventions to identify modality-specific control pathways, and rollout-level behavioral probes to examine grounding, shortcut dependence, and semantic following.
Experiments on $\pi_{0.5}$ and OpenVLA reveal three key findings. First, the two models exhibit distinct modality-specific adaptation dynamics during VLA finetuning. Second, they rely on different multimodal routing strategies and layer-wise dependencies during action decoding. Third, although VLA policies excel at visually grounded trajectory generation, they remain limited in fine-grained semantic following. These findings highlight future directions for representation-preserving adaptation, causal VLA circuits, and compositional semantic control.
\end{abstract}
\section{Introduction}

Vision-Language-Action (VLA) models have become a promising paradigm for advancing embodied intelligence in real-world environments.  
Built upon large-scale pretrained vision-language models (VLMs)~\cite{paligemma,llama2,florence2,shan2026epic,bai2025qwen3}, recent systems, including X-VLA~\cite{x-vla}, OpenVLA-style models~\cite{openvla-oft,openvla}, and $\pi$-style models~\cite{pi05,pi0,pi0fast,pi06}, achieve strong robotic control performance by unifying visual-language perception and action generation within a single policy framework. However, despite these successes, the internal mechanisms underlying action generation remain poorly understood.
This issue is critical because, without understanding how robotic policy learning reshapes internal representations, it is difficult to diagnose model failures and, consequently, to design more effective VLA architectures. 
In particular, it remains unclear whether multimodal knowledge is preserved, how visual and linguistic signals are aligned during policy learning, and which modalities govern action decoding.

\begin{figure*}[ht!]
    \centering
    \includegraphics[width=\textwidth]{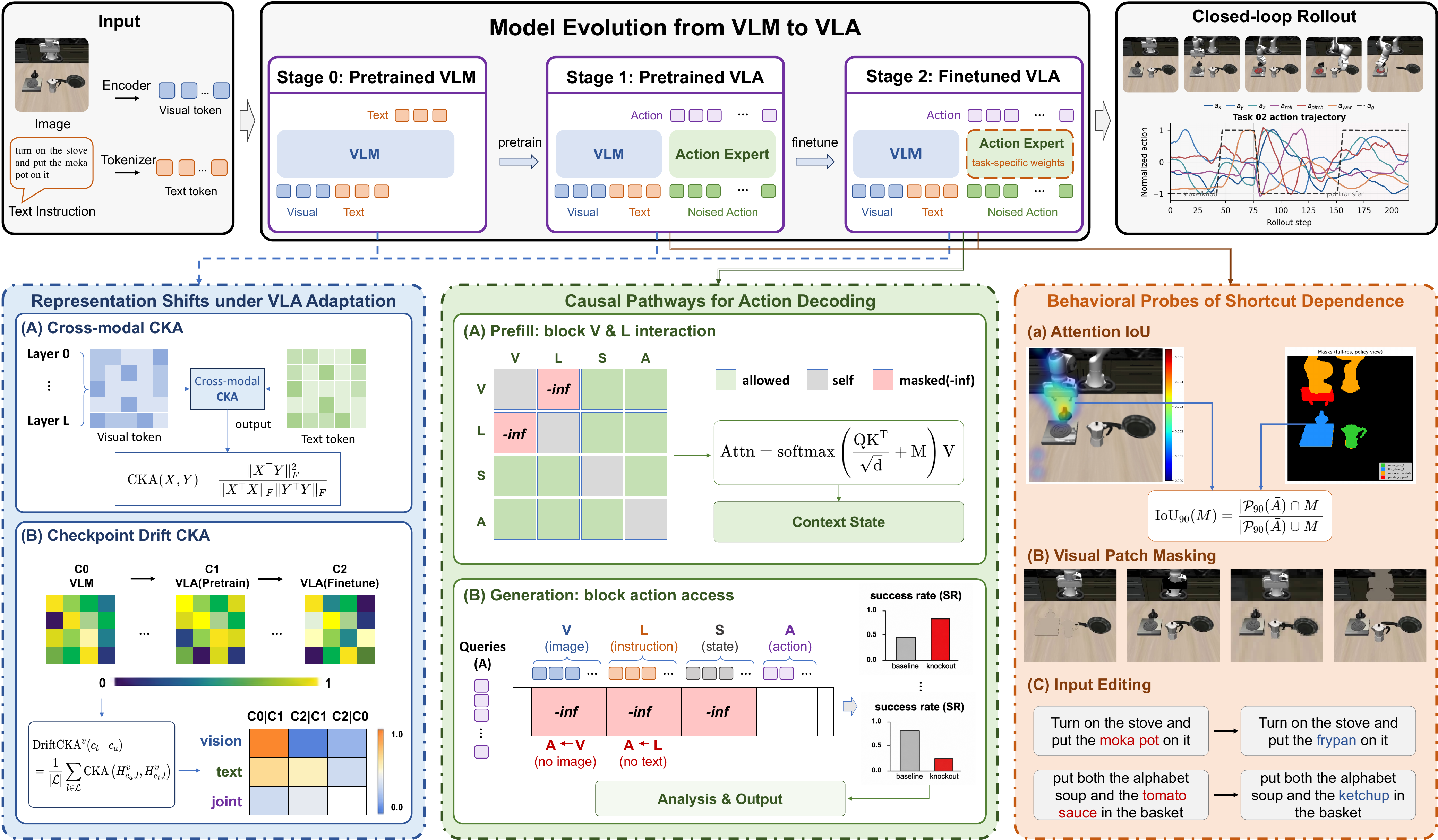}
    \vspace{-15pt}
    \caption{Overview of VLA-Trace. The framework progressively diagnoses VLA models by tracing representation dynamics, identifying causal pathways for action decoding, and probing behavioral reliance on shortcut dependence.}
    \vspace{-10pt}
    \label{fig:overview}
\end{figure*}

To answer these questions, we propose \textbf{VLA-Trace}, a systematic three-stage analysis framework that progressively investigates VLA models from latent representation evolution to causal control attribution and finally to behavioral dependence. As illustrated in Fig.~\ref{fig:overview}, instead of treating the policy as an end-to-end system, we construct a unified evidence chain for interpreting VLA adaptation. First, we employ cross-modal and checkpoint-drift centered kernel alignment (CKA) \cite{CKA} to identify whether vision-language prior is preserved or reshaped during fine-tuning. Second, we perform targeted attention knockouts to examine whether vision-language information is functionally utilized. Third, we combine attention localization and input editing to assess whether the resulting routing patterns reflect robust vision-language grounding or shallow visual shortcuts. Through this framework, we uncover the following key observations:
\vspace{-7pt}
\paragraph{$\pi_{0.5}$ and OpenVLA exhibit distinct modality-specific adaptation dynamics during VLA finetuning.} Specifically, our CKA analysis reveals that $\pi_{0.5}$ exhibits highly unstable layer-wise cross-modal fusion dynamics, primarily reorganizing textual representations into task-conditioned control features. In contrast, OpenVLA exhibits smoother but often weaker intermediate-layer image--text alignment, while preserving its text-pooled representations and reorganizing primarily vision-pooled and joint-pooled subspaces.

\paragraph{$\pi_{0.5}$ and OpenVLA exhibit distinct multimodal routing strategies and layer-wise dependency during action decoding.} Attention knockout experiment across visual and textual pathways illustrates that $\pi_{0.5}$ is dominated by a concentrated visual-to-action routing pathway during decoding, which limits the functional role of language. Conversely, OpenVLA distributes control-relevant information across both visual and textual access.

\paragraph{VLA policies excel at visually grounded trajectory but remain limited in fine-grained semantic following.}
Attention localization and input editing reveal a gap between visual grounding and semantic following. Although policies localize to manipulation-relevant regions, they often ignore fine-grained semantic modifications, suggesting strong visually grounded trajectory imitation and weak compositional language control.

These observations highlight future directions for effective representation preservation and adaptation, causal VLA circuits, and compositional semantic control in VLA systems. Overall, this work proposes a systematic framework for probing how vision-language prior knowledge influences action control in VLA models. By identifying these bottlenecks in multimodal adaptation, causal routing, and semantic grounding, our progressive diagnosis provides guidance for developing more interpretable and robust next-generation VLA models.

\section{Related Work} \label{sec:related}

\vspace{-5pt}
In this section, we review related work around two topics: \textit{VLA models} and \textit{mechanistic and representation analysis of VLA models}. These areas motivate the design of VLA-Trace as a progressive analysis framework.

\vspace{-5pt}
\subsection{Vision-Language-Action Models}

While large multimodal models have powered diverse agentic applications ranging from web navigation~\cite{guo2026web} and data visualization~\cite{lu2026multivis} to video generation~\cite{li2024anim}, VLA models specifically adapt the vision-language priors of VLMs to embodied control. Unlike world foundation models~\cite{cosmoswfm,genieenvisioner,dreamzero,lu2026current}, which model environment dynamics through high-dimensional video generation, VLAs offer a more direct and lightweight paradigm by mapping multimodal representations to robot actions~\cite{reed2022generalist,brohan2022rt1,zitkovich2023rt,li2024vision}. Recent advances, including Octo~\cite{team2024octo}, OpenVLA~\cite{openvla}, and $\pi$-style policies~\cite{pi05,pi0,pi06,pi07}, have shown strong generalization across tasks and embodiments, while systems such as \textit{Gemini Robotics}~\cite{Gemini_Robotic_1.5,Gemini_Robotic} and GR00T N1~\cite{gr00tn12025} further extend VLA capabilities to complex manipulation and control.


Despite these advances, it remains unclear how VLM representations are reshaped during VLA adaptation and whether pretrained knowledge is preserved and functionally utilized during action generation. Prior work shows that direct VLM-to-VLA finetuning can degrade visual and language representations~\cite{grover2025preserving,hancock2025actions}, motivating a closer examination of how representation changes influence embodied behavior.

\vspace{-5pt}
\subsection{Mechanistic Analysis of VLA Models}
Recent studies investigate the underlying mechanisms of VLA models, which can be broadly categorized into three directions:

\textbf{Representation-level analysis} investigates the information encoded in hidden states. Probing analyses show that hidden states can encode symbolic properties, relations, action states, and latent transition information, suggesting that semantically meaningful state information is present in VLA representations~\cite{fang2025intention,hancock2025actions,grant2026features}.
\textbf{Causal mechanism analysis} moves toward mechanistic interpretability by steering activations, identifying task-relevant attention heads, and extracting sparse features that influence policy behavior~\cite{mitra2025mechanistic,haon2025mechanistic,swann2026sparse}.
\textbf{Input intervention analysis} examines grounding and robustness at input levels, revealing that VLA policies remain sensitive to distractors, background changes, object grounding errors, and failures to follow language instructions~\cite{kerr2023lerf,li2024shapegrasp,hancock2024byovla,xie2026strong,fei2025libero,zhang2026a2eval}.



Nevertheless, existing analyses remain fragmented. Representation-level studies reveal what information is encoded, but not whether it is causally used during action decoding. Input intervention analyses expose behavioral failures, yet often leave the internal pathway from representation shifts to failure modes unclear. To bridge this gap, we propose \textbf{VLA-Trace}, a unified diagnostic framework that integrates representation geometry analysis, causal analysis, and input intervention into a stage-wise progressive pipeline to understand embodied control behavior.

\section{Representation Analysis Framework}

\definecolor{benchbg}{RGB}{235,242,255}
\definecolor{modelbg}{RGB}{235,247,237}

\newcommand{\bmark}[1]{\colorbox{benchbg}{\scriptsize\strut B#1}}
\newcommand{\mmark}[1]{\colorbox{modelbg}{\scriptsize\strut M#1}}

\begin{table}[t]
\centering
\small
\renewcommand{\arraystretch}{1.18}
\setlength{\tabcolsep}{4pt}
\arrayrulecolor{rulegray}

\begin{tabular}{@{}p{0.5cm}p{3.2cm}cc@{}}
\toprule
\textbf{Stage} & \textbf{Analysis Probe} & \textbf{Benchmarks} & \textbf{Models} \\
\midrule

\multirow{2}{*}{S1}
& \makecell[l]{Cross-Modal CKA\\
\probetarget{vision--language alignment}}
& \colorbox{benchbg}{LI/CO}
& \colorbox{modelbg}{P/O} \\

& \makecell[l]{Checkpoint-Drift CKA\\
\probetarget{stage-wise drift}}
& \colorbox{benchbg}{LI}
& \colorbox{modelbg}{P/O} \\
\midrule

\multirow{2}{*}{S2}
& \makecell[l]{Attention Knockout\\
\probetarget{modality reliance}}
& \colorbox{benchbg}{LI/CV/RT}
& \colorbox{modelbg}{P/O/F} \\

& \makecell[l]{Layer-Wise Knockout\\
\probetarget{layerwise control}}
& \colorbox{benchbg}{LI/CV/RT}
& \colorbox{modelbg}{P/O/F} \\
\midrule

\multirow{3}{*}{S3}
& \makecell[l]{Attention IoU\\
\probetarget{region overlap}}
& \colorbox{benchbg}{L10}
& \colorbox{modelbg}{P/O} \\

& \makecell[l]{Visual Patch Mask\\
\probetarget{object/spatial/background}}
& \colorbox{benchbg}{LI/CV/RT/SP}
& \colorbox{modelbg}{P/O/F/X} \\

& \makecell[l]{Input Editing\\
\probetarget{image/prompt sensitivity}}
& \colorbox{benchbg}{L10}
& \colorbox{modelbg}{P/O} \\
\bottomrule
\end{tabular}
\vspace{-5pt}
\caption{Overview of the VLA-Trace framework. Probes progress from representation geometry to causal attribution and behavioral manifestation; italic text summarizes each probe's target. Blue benchmark tags: \colorbox{benchbg}{LI}=LIBERO, \colorbox{benchbg}{CO}=COCO, \colorbox{benchbg}{CV}=CALVIN, \colorbox{benchbg}{RT}=RoboTwin2.0, \colorbox{benchbg}{SP}=Simpler, \colorbox{benchbg}{L10}=LIBERO-10. Green model tags: \colorbox{modelbg}{P}=$\pi_{0.5}$, \colorbox{modelbg}{O}=OpenVLA, \colorbox{modelbg}{F}=OFT, \colorbox{modelbg}{X}=X-VLA.}
\label{tab:probe}
\vspace{-0.6cm}
\end{table}

\paragraph{Overview.} We investigate three stage-wise representations: C0, the pretrained VLM; C1, the pretrained VLA; and C2, the task-finetuned VLA. 
This staged formulation enables us to trace how general vision-language knowledge is preserved or reorganized during robotic policy learning. Built upon this progression, our framework integrates three complementary analyses. 
First, representation-level analysis uses CKA to examine whether visual, textual, and joint representation geometries are preserved or reshaped across three stages. Second, causal analysis (stage 2) employs attention knockouts to evaluate whether these representations are functionally required for action decoding. Third, input intervention analysis (stages 1-2), including attention localization, visual masking, and input editing, investigates whether the identified pathways lead to robust spatial grounding, shortcut dependence, or limited semantic control in behavior. Tab.~\ref{tab:probe} summarizes the probing design, while Tab.~\ref{tab:vlas} presents the evaluated models and designed benchmarks, with additional results in the Appendix.

\vspace{-5pt}
\paragraph{Experiment Setup.} We primarily focus on $\pifive$ and OpenVLA, as they present contrasting structures that are highly diagnostic for language and vision utilization. $\pifive$ employs a PaliGemma \cite{paligemma} with an action expert under flow-matching action generation. Its transformer input consists of multi-view images, a BOS token, the task instruction, and a newline token, allowing bidirectional visual-language interaction during context formation. Conversely, OpenVLA utilizes an autoregressive Llama-based architecture~\cite{llama2} where image patches are followed by a prompt template (e.g., \textit{``In: What action should the robot take to \{instruction\}? Out:''}). 
For evaluation, we use COCO~\cite{coco} as a generation-domain image-text reference and LIBERO~\cite{liu2023libero} as the robot-domain benchmark. 

\subsection{Representation Shifts under VLA Adaptation}
\label{subsec:cka_representation}
\paragraph{Evaluation Metrics.} First, we investigate the evolution of latent representations across training stages to analyze the impact of robot-domain training on vision--language alignment. Specifically, we measure CKA similarity~\cite{CKA} between pooled visual and textual representations across layers and stages using COCO and LIBERO as evaluation datasets. We focus on two complementary analyses:

\begin{figure*}[ht!]
    \centering
    \includegraphics[width=15cm]{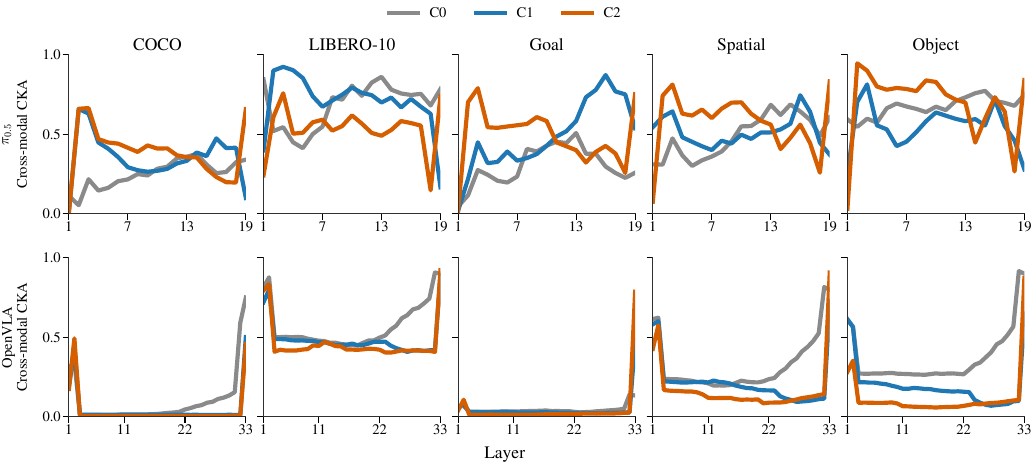}
    \vspace{-8pt}
    \caption{
    Layer-wise image--text CKA across datasets and training stages. The panels compare cross-modal alignment for $\pi_{0.5}$ and OpenVLA. Each curve denotes one checkpoint: C0, the pretrained VLM; C1, the pretrained VLA; and C2, the task-finetuned VLA.
    }
    \label{fig:cross_modal_cka_grid}
\end{figure*}

\textbf{(1) Cross-modal CKA} measures the alignment between visual and textual representations within a model. Given two representation matrices $X \in \mathbb{R}^{N \times d_x}$ and $Y \in \mathbb{R}^{N \times d_y}$, the linear CKA similarity is defined as:
$
\mathrm{CKA}(X,Y)=
\frac{\lVert X^\top Y \rVert_F^2}
{\lVert X^\top X \rVert_F \lVert Y^\top Y \rVert_F} \nonumber
$

\textbf{(2) Checkpoint Drift CKA} characterizes representational changes across training stages. Specifically, for each layer, we keep three pooled views: \textit{vision pooled}, \textit{text pooled},  and \textit{joint pooled}. For a given view $v$, we first compute the CKA between the same layer of an anchor checkpoint $c_a$ and a target checkpoint $c_t$, and then average these layer-wise values:
$
\mathrm{DriftCKA}^{v}(c_t\mid c_a)=
\frac{1}{|\mathcal{L}|}\sum_{l\in\mathcal{L}} \nonumber
\mathrm{CKA}\!\left(H^{v}_{c_a,l}, H^{v}_{c_t,l}\right)
$
where $H^{v}_{c,l}\in\mathbb{R}^{N\times d}$ denotes the sample-by-feature matrix of view $v$ at layer $l$ from checkpoint $c$, and $\mathcal{L}$ denotes the set of evaluated layers. Therefore, Drift CKA measures the geometric similarity of representations across different training stages. Higher values indicate stronger geometric preservation across checkpoints, whereas lower values reflect greater representational reorganization. Further experimental details are provided in the Appendix~\ref{app:cka}.

\subsubsection{CKA Alignment Analysis}
\paragraph{Vision-Language Alignment.} 
Fig.~\ref{fig:cross_modal_cka_grid} reveals different layer-wise cross-modal organization patterns in $\pifive$ and OpenVLA. For $\pifive$, cross-modal CKA is broadly distributed across intermediate layers and varies substantially across checkpoints and datasets, suggesting active cross-modal reorganization during VLA pretraining and task finetuning. OpenVLA, in contrast, often shows smoother but weaker image--text CKA across intermediate layers, with high alignment frequently concentrated near the boundary or terminal layers.
This observation may stem from differences in architectural design, particularly the contrast between bidirectional and unidirectional attention mechanisms. Specifically, \textit{the bidirectional attention mechanism in $\pifive$ may induce more complex and fluctuating cross-modal fusion dynamics, resulting in less stable alignment patterns across layers \textbf{(Finding~1)}.}

\begin{figure}[ht!]
    \includegraphics[width=\columnwidth]{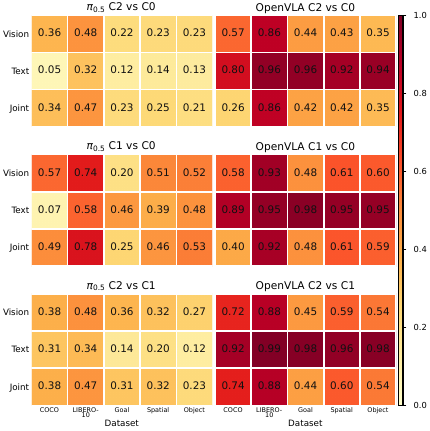}
    \vspace{-15pt}
    \caption{Stage-wise checkpoint-drift CKA for $\pifive$ and OpenVLA. Each cell reports the mean matched-layer CKA for vision, text, and joint representations.}
    \label{fig:drift_cka}
    \vspace{-0.5cm}
\end{figure}

\paragraph{Checkpoint Drift and Adaptation.} 

Next, we analyze checkpoint-drift CKA to identify preserved and reorganized subspaces during robot adaptation. In Fig.~\ref{fig:drift_cka}, $\pifive$ exhibits substantially stronger drift in textual representations, suggesting that language representations are heavily reorganized into task-conditioned control features. In contrast, OpenVLA primarily restructures visual and joint representation spaces. These results reveal \textit{distinct modality-specific preservation and adaptation dynamics across the two models \textbf{(Finding~2)}.}

\subsection{Causal Pathways for Action Decoding}
\label{subsec:attention_knockout}

\begin{figure*}[ht!]
    \includegraphics[width=15cm]{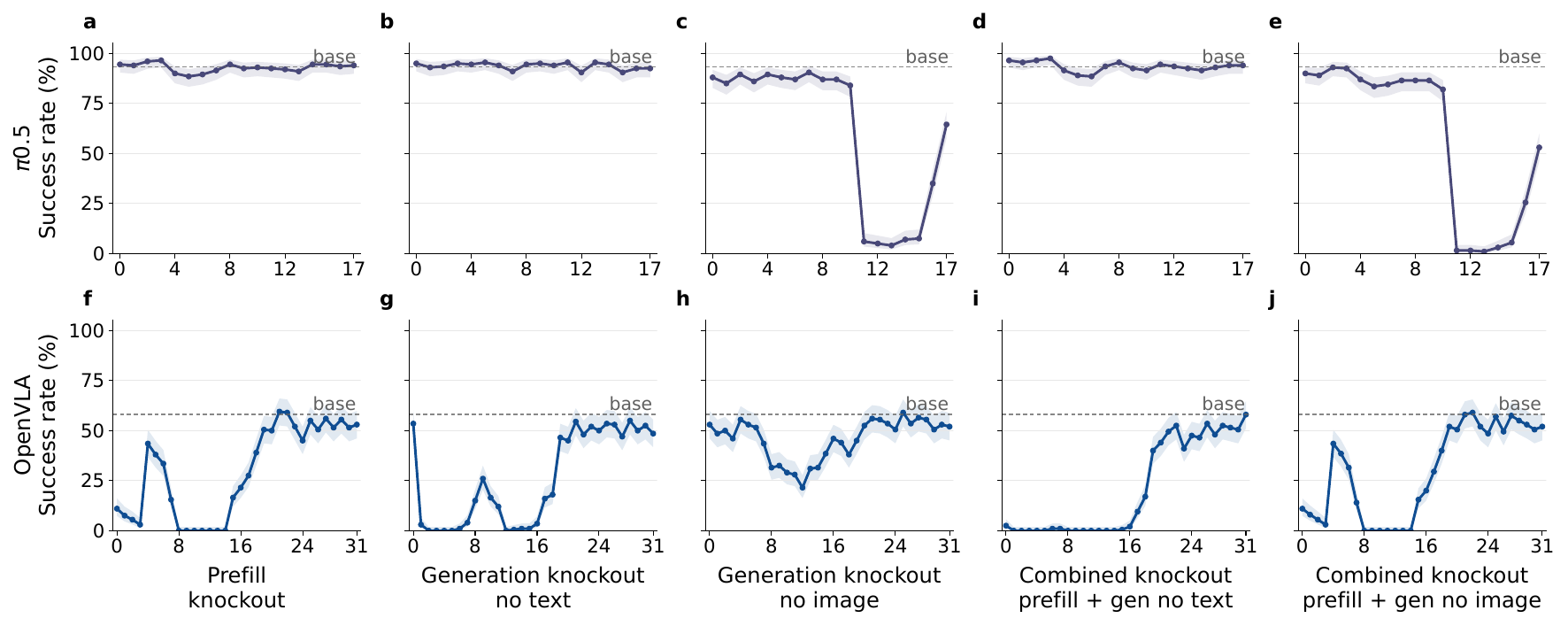}
    \vspace{-10pt}
    \caption{Attention knockout for $\pifive$ (top) and OpenVLA (bottom) on LIBERO-10. Similar observations for LIBERO-Goal, Object, and Spatial are provided in Figure \ref{fig:pi05_layerwise_window7} and \ref{fig:openvla_layerwise_window7}.}
    \label{fig:overview_knockout}
    \vspace{-0.1cm}
\end{figure*}
\paragraph{Evaluation Metrics.} Second, we probe how latent representations causally influence action generation. Specifically, we decompose inference into representation formation and action decoding, and perform targeted attention knockout. This allows us to determine whether a modality contributes by shaping the initial multimodal context, by directly conditioning the action tokens during decoding, or by both mechanisms. Further experimental details are provided in the Appendix~\ref{sec:appendix_knockout_protocol}.

\textbf{\textit{(1)~Cross-modal Interaction (Prefill).}}
The prefill stage encodes all input modalities (vision and language) to construct the initial internal representations. To isolate the role of cross-modal interaction, we block the attention between vision and language tokens (V–L attention), thereby enforcing independent modality processing. This allows us to assess how cross-modal integration during representation formation influences action generation. 

\textbf{\textit{(2)~Modality Dependency (Gen).}}
During generation, actions are produced autoregressively. We perform a modality knockout by blocking attention from action tokens to vision or language tokens, isolating the dependence of action decoding on multimodal context.

This design disentangles two roles of multimodal information in VLAs: (i) cross-modal interaction during representation encoding, and (ii) modality dependency during action decoding. By intervening at these two stages, we can distinguish whether performance gains arise from better multimodal understanding or more effective utilization of modality-specific information during generation. Three attention configurations are considered: \textbf{i.} Baseline, where all modalities are provided; \textbf{ii.} No Image, where visual attention is removed; \textbf{iii.} No Text, where language attention is excluded.
In this experiment, model performance is evaluated using success rate (\%) over 200 episodes (20 trials across 10 LIBERO-10 tasks).

\subsubsection{Attention Knockout}

\newcommand{\cmark}{\ding{51}}
\newcommand{\xmark}{\ding{55}}

\begin{table}[t]
\centering
\small
\setlength{\tabcolsep}{2pt}
\renewcommand{\arraystretch}{1.08}
\begin{tabular}{lccc|cccc}
\toprule
Model & P & G-T & G-I & LIBERO-10 & Goal & Spatial & Object \\
\midrule
\multirow{6}{*}{$\pifive$}
& \cmark & \cmark & \cmark & 93.5 & 96.0 & 98.5 & 99.5 \\
& \xmark & \cmark & \cmark & 0.0 & 11.5 & 77.0 & 71.5 \\
& \cmark & \xmark & \cmark & 39.0 & 96.5 & 99.0 & 98.0 \\
& \cmark & \cmark & \xmark & 0.0 & 4.0 & 0.0 & 0.0 \\
& \xmark & \xmark & \cmark & 71.5 & 10.5 & 70.5 & 46.0 \\
& \xmark & \cmark & \xmark & 0.0 & 0.0 & 0.0 & 0.0 \\
\midrule
\multirow{6}{*}{OpenVLA}
& \cmark & \cmark & \cmark & 58.0 & 74.5 & 75.5 & 74.0 \\
& \xmark & \cmark & \cmark & 0.0 & 0.0 & 0.0 & 0.0 \\
& \cmark & \xmark & \cmark & 0.0 & 0.0 & 0.0 & 0.0 \\
& \cmark & \cmark & \xmark & 1.0 & 16.0 & 44.0 & 32.5 \\
& \xmark & \xmark & \cmark & 0.0 & 0.0 & 0.0 & 0.0 \\
& \xmark & \cmark & \xmark & 0.0 & 0.0 & 0.0 & 0.0 \\
\bottomrule
\end{tabular}
\vspace{-5pt}
\caption{All-layer attention knockout results. Values show success rates. Rows correspond to the original settings used throughout the paper: Baseline, Prefill, Gen: no text, Gen: no image, Comb. + gen no text, and Comb. + gen no image. Here, `P' denotes the prefill pathway, while `G-T' and `G-I' denote text and image availability during generation, respectively. Each rate is computed over 200 episodes per suite.}
\label{tab:all_layer_knockout_main}
\vspace{-0.5cm}
\end{table}

We first apply all-layer attention knockout to measure the global necessity of each pathway, and then use layer-wise knockout to localize where the dependency is concentrated. Tab~\ref{tab:all_layer_knockout_main} reports the all-layer results across four LIBERO suites. For $\pifive$, removing visual access during generation is consistently destructive. \texttt{Gen:no image} reduces success to nearly zero across all datasets. In contrast, \texttt{Gen:no text} largely preserves performance on Goal, Object, and Spatial, although LIBERO-10 remains more sensitive. 
The prefill-stage interventions further reinforce this interpretation. Moreover, combining prefill no-V--L with generation no-image completely collapses performance. These results suggest that the most fragile decoding-time pathway in $\pifive$ is the visual-to-action route.


OpenVLA exhibits a different dependency pattern. All-layer \texttt{Gen:no text} reduces success to 0.00\% on every LIBERO suite, and prefill-stage image removal is equally destructive. In contrast, \texttt{Gen:no image} is harmful but does not uniformly collapse all datasets, suggesting that OpenVLA depends strongly on prompt-region access during decoding while also requiring visual grounding formed during prefill. Overall, these results indicate that \textit{$\pifive$ and OpenVLA exhibit distinct multimodal routing strategies during action generation: $\pifive$ primarily depends on visual-to-action pathways, whereas OpenVLA relies on both visual and language modalities \textbf{(Finding~3)}.}


Crucially, the same intervention is harmless on non-essential pathways: during generation, dropping the instruction tokens costs $\pifive$ only $1.3$ points and dropping the newline token costs OpenVLA $0.4$ points, against $95.9$ and $70.5$ points for the image and prompt pathways (Tab.~\ref{tab:attention_knockout_all_layers_summary_0526}). The collapses are therefore pathway-specific, and we read all knockout results as necessary under this defined intervention. We then use layer-wise knockout to localize these global effects. Fig.~\ref{fig:overview_knockout} shows that, on the LIBERO-10 task, $\pifive$ has a narrow visual bottleneck. \texttt{Gen:no image} drops to 4.00\% at the most vulnerable layer, while \texttt{Gen:no text} remains close to baseline across layers. Combining \texttt{prefill} with \texttt{Gen:no image} further sharpens this collapse around the same critical region. By contrast, OpenVLA shows broader vulnerable regions. \texttt{Gen:no text}, \texttt{prefill}, and their combined intervention reach 0.00\% across multiple layers. The full layer-wise results over window sizes and LIBERO suites are reported in the supplementary material. These results indicate that \textit{critical information is more broadly distributed in OpenVLA than in $\pifive$, whereas $\pifive$ routes action-critical visual information through a narrower bottleneck \textbf{(Finding~4)}.} 

\subsection{Behavioral Probes of Grounding and Shortcut Dependence}
\label{subsec:behavioral_probes}

\paragraph{Evaluation Metrics.} Finally, we investigate whether the previously identified representation shifts and causal pathways translate into observable rollout behavior. Specifically, while Sec.~\ref{subsec:cka_representation} examines how multimodal representations are reorganized and Sec.~\ref{subsec:attention_knockout} identifies the pathways required for action decoding, this section evaluates whether such pathways support robust grounding and instruction-conditioned behavior during closed-loop execution. We focus on three complementary behavioral probes:

\textbf{\textit{(1) Attention IoU and Patterns.}}
We first analyze action-to-image attention during LIBERO-10 rollouts. For each rollout step, we average action-query attention over visual patches to obtain an action-conditioned heatmap $\bar{A}$. Given a simulator-derived mask $M$ for task-relevant objects, robot regions, or their union, we compute three localization metrics. 
The high-attention patch IoU measures~\cite{IOU} whether thresholded high-confidence
attention patches overlap with task-relevant regions:
$\mathrm{IoU}_{90}(M)=
\frac{|\mathcal{H}_{90}(\bar{A})\cap M|}
{|\mathcal{H}_{90}(\bar{A})\cup M|},$
where $\mathcal{H}_{90}(\bar{A})=\{j:\bar{A}_j \ge q_{0.9}(\bar{A})\}$ denotes patches whose attention is larger than the 90th percentile.
The continuous attention mass measures how much attention is allocated to the region:
$
\mathrm{Mass}(M)=
\frac{\sum_{j\in M}\bar{A}_{j}}{\sum_{j}\bar{A}_{j}}.
$
Finally, the peak-hit rate measures whether the maximum-attention patch lies inside the target mask:
$
\mathrm{Hit}(M)=
\mathbf{1}\!\left[\arg\max_j \bar{A}_{j}\in M\right].
$
Because LIBERO-10 tasks are long-horizon instructions that typically contain two sequential subgoals, we further divide each rollout into two temporal phases by the first and second halves of the executed steps. 
This phase split provides an approximate but consistent proxy for the first and second instruction subgoals, allowing us to test whether visual attention evolves with the temporal structure of the task.

\textbf{\textit{(2) Visual Patch Masking.}} We mask target objects, gripper regions, robot bodies, and backgrounds using background replacement, black masking, and mosaic masking. Comparing success rates under these perturbations reveals whether the policy depends on target-object appearance, robot-object spatial relations, or broader scene context.

\textbf{\textit{(3) Input Editing.}} We edit the language instruction while preserving the task template, testing whether language can redirect policy behavior toward new targets. Failure to adapt indicates rigid dependence on the original visual-task configuration.



\subsubsection{Attention IoU and Patterns}

\begin{table}[t]
\centering
\small
\resizebox{\columnwidth}{!}{%
\fontsize{10pt}{10pt}
\begin{tabular}{llcccc}
\toprule
\multirow{2}{*}{Model} & \multirow{2}{*}{Stage}
& \multirow{2}{*}{Mass}
& \multicolumn{2}{c}{$\mathrm{IoU}_{90}$}
& \multirow{2}{*}{Hit} \\
\cmidrule(lr){4-5}
& & & Object & Robot+Object & \\
\midrule
\multirow{3}{*}{OpenVLA}
& Phase 1 & $0.5437$ & $0.1374$ & $0.1830$ & $0.9204$ \\
& Phase 2 & $0.5563$ & $0.1294$ & $0.1965$ & $0.9433$ \\
& Full    & $0.5882$ & $0.1504$ & $0.1965$ & $0.9597$ \\
\midrule
\multirow{3}{*}{$\pi_{0.5}$}
& Phase 1 & $0.6100$ & $0.1312$ & $0.2303$ & $0.6225$ \\
& Phase 2 & $0.5805$ & $0.1421$ & $0.2265$ & $0.6259$ \\
& Full    & $0.6328$ & $0.1379$ & $0.2233$ & $0.6349$ \\
\bottomrule
\end{tabular}%
}
\vspace{-5pt}
\caption{LIBERO-10 attention localization across temporal task phases. Metrics compare grounding on isolated target objects (\texttt{Object}) versus broader interaction regions (\texttt{Robot+Object}). \texttt{Mass} and \texttt{Hit} use the \texttt{Robot+Object} mask.}
\label{tab:libero10-phase-compact}
\vspace{-0.5cm}
\end{table}

\begin{figure*}[ht!]
    \includegraphics[width=\linewidth]{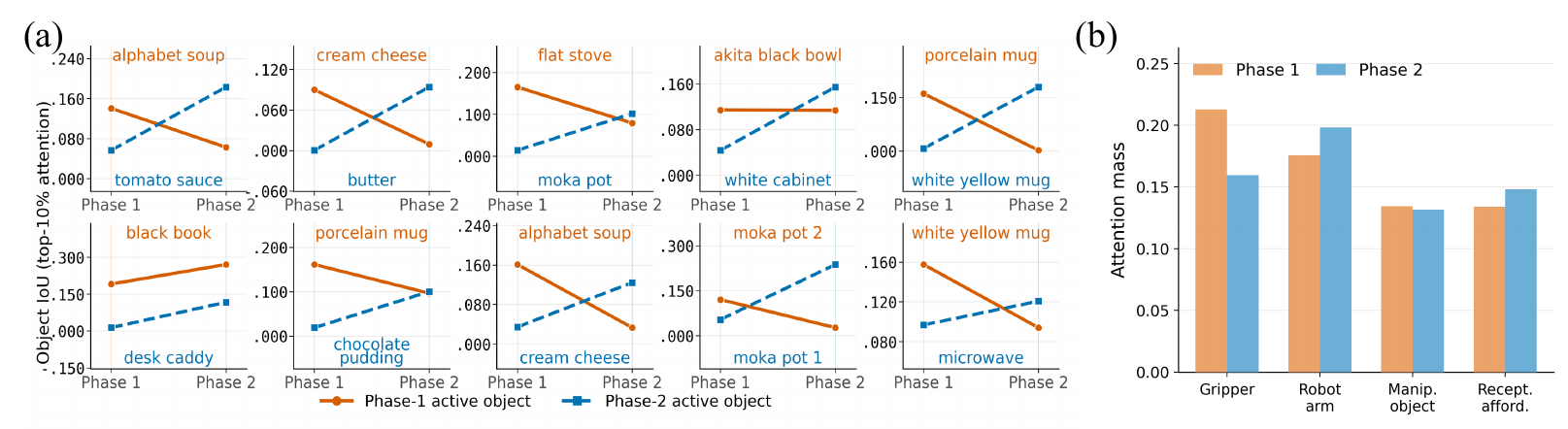}
    \vspace{-15pt}
    \caption{
    $\pifive$ attention IoU and mass on LIBERO-10. (a) Object IoU dynamically shifts between the first (Phase 1) and second (Phase 2) instruction subgoals. (b) Attention mass allocation over robot and object regions. These results indicate that VLA policies successfully generate visually grounded trajectories by tracking task-relevant objects over time. See similar observations in Fig.~\ref{fig:attention_openvla} for OpenVLA and Tab. \ref{tab:instruction_all} for instructions.}
    \label{fig:attention_iou}
\end{figure*}

As shown in Tab.~\ref{tab:libero10-phase-compact}, both models exhibit consistent overlap between high-attention patches and Robot+Object regions. 
Importantly, Robot+Object $\mathrm{IoU}_{90}$ is consistently higher than Object-only $\mathrm{IoU}_{90}$ for both models, suggesting that action attention is not routed solely to semantic object regions, but to broader robot-object interaction regions that include the gripper, robot arm, manipulated object, and task-relevant affordances. 
Beyond spatial localization, Fig.~\ref{fig:attention_iou} demonstrates that policies implicitly encode temporal task semantics. By splitting the two-subgoal LIBERO-10 tasks into Phase 1 and Phase 2, we observe that attention shifts consistently align with task progression. E.g., in the task ``turn on the stove and put the moka pot on it,'' the model's focus correctly transitions from the stove to the moka pot across phases. Further experimental details are provided in Appendix~\ref{app:appendix_attention_metrics}.

Despite both exhibiting spatial grounding, their routing mechanisms differ (as shown in Tab.~\ref{tab:libero10-phase-compact}). $\pifive$ distributes attention broadly (higher mass, lower peak-hit rates), while OpenVLA concentrates it sparsely (higher peak-hit rates, lower mass and IoU). Corroborating Sec.~\ref{subsec:attention_knockout}, this reflects $\pifive$'s dominant visual-to-action pathway compared to OpenVLA's reliance on both visual and prompt regions.
Overall, these observations indicate that \textit{the models exhibit spatially grounded and temporally structured action attention, routing action generation through robot-object interaction regions and shifting object-level attention across rollout phases \textbf{(Finding~5)}}.

\subsubsection{Visual Patch Masking}
\begin{table*}[t]
\centering
\scriptsize
\begin{tabular}{llccccc}
\toprule
Model & Setting & LIBERO-10 & LIBERO-Object & LIBERO-Spatial & LIBERO-Goal & Avg. Drop\\
\midrule

\multirow{15}{*}{$\pifive$}
& Baseline & 75.00 & 95.60 & 95.80 & 80.00 & - \\
\cmidrule{2-7}
& Target (BG) & 0.00 & 13.00 & 7.80 & 19.80 & \textbf{76.45} \\
& Target (Black) & 19.60 & 61.60 & 23.00 & 58.80 & 45.85 \\
& Target (Mosaic) & 47.20 & 74.00 & 50.20 & 58.40 & 29.15 \\
\cmidrule{2-7}
& Gripper (BG) & 8.40 & 3.20 & 64.00 & 59.20 & 52.90 \\
& Gripper (Black) & 52.60 & 97.40 & 94.80 & 74.80 & 6.70 \\
& Gripper (Mosaic) & 53.20 & 79.40 & 89.80 & 74.20 & 12.45 \\
\cmidrule{2-7}
& Robot (BG) & 3.40 & 26.40 & 47.60 & 44.00 & \underline{56.25} \\
& Robot (Black) & 30.20 & 93.60 & 95.00 & 69.80 & 14.45 \\
& Robot (Mosaic) & 39.20 & 74.60 & 89.60 & 68.60 & 18.60 \\
\cmidrule{2-7}
& Robot w/o Gripper (BG) & 52.60 & 91.80 & 97.40 & 76.80 & 6.95 \\
& Robot w/o Gripper (Black) & 58.40 & 92.80 & 99.00 & 76.40 & 4.85 \\
& Robot w/o Gripper (Mosaic) & 64.40 & 96.60 & 96.40 & 79.00 & 2.50 \\
\cmidrule{2-7}
& Background (Black) & 41.00 & 79.40 & 67.20 & 57.80 & 25.25 \\
& Background (Mosaic) & 57.80 & 86.60 & 93.00 & 78.00 & 7.75 \\
\midrule

\multirow{15}{*}{OpenVLA}
& Baseline & 54.33 & 70.00 & 79.67 & 74.00 & - \\
\cmidrule{2-7}
& Target (BG) & 5.00 & 0.33 & 36.67 & 45.00 & 47.75\\
& Target (Black) & 20.67 & 55.67 & 63.33 & 73.00 & 16.33\\
& Target (Mosaic) & 25.00 & 59.67 & 70.00 & 67.67 & 13.92 \\
\cmidrule{2-7}
& Gripper (BG) & 10.67 & 1.00 & 19.00 & 38.00 & 52.33 \\
& Gripper (Black) & 31.33 & 72.00 & 70.67 & 69.33 & 8.67 \\
& Gripper (Mosaic) & 20.33 & 3.67 & 53.00 & 63.00 & 34.50 \\
\cmidrule{2-7}
& Robot (BG) & 0.00 & 0.00 & 0.00 & 0.33 & \textbf{69.42} \\
& Robot (Black) & 0.67 & 4.00 & 3.33 & 6.00 & 66.00\\
& Robot (Mosaic) & 0.33 & 0.00 & 2.00 & 8.00 & \underline{66.92} \\
\cmidrule{2-7}
& Robot w/o Gripper (BG) & 10.00 & 34.33 & 22.00 & 25.67 & 46.50\\
& Robot w/o Gripper (Black) & 11.67 & 17.33 & 31.67 & 26.00 & 47.83 \\
& Robot w/o Gripper (Mosaic) & 16.00 & 15.67 & 49.67 & 38.33 & 39.58 \\
\cmidrule{2-7}
& Background (Black) & 26.33 & 42.67 & 47.33 & 41.67 & 30.00 \\
& Background (Mosaic) & 48.00 & 65.67 & 82.33 & 70.67 & 2.83 \\
\midrule

\multirow{15}{*}{OpenVLA-OFT}
& Baseline & 94.80 & 99.80 & 92.80 & 97.40 & - \\
\cmidrule{2-7}
& Target (BG) & 10.40 & 57.00 & 26.80 & 49.60 & \textbf{60.25} \\
& Target (Black) & 15.20 & 84.80 & 35.40 & 60.20 & \underline{47.30} \\
& Target (Mosaic) & 48.20 & 98.40 & 76.00 & 88.40 & 18.45 \\
\cmidrule{2-7}
& Gripper (BG) & 89.80 & 94.20 & 92.40 & 96.00 & 3.10 \\
& Gripper (Black) & 89.80 & 98.80 & 92.80 & 97.00 & 1.60 \\
& Gripper (Mosaic) & 95.00 & 100.00 & 93.80 & 97.60 & -0.40 \\
\cmidrule{2-7}
& Robot (BG) & 66.80 & 96.00 & 81.40 & 67.20 & 18.35 \\
& Robot (Black) & 79.00 & 96.80 & 86.80 & 87.20 & 8.75 \\
& Robot (Mosaic) & 84.40 & 97.60 & 91.60 & 94.60 & 4.15 \\
\cmidrule{2-7}
& Robot w/o Gripper (BG) & 85.40 & 98.60 & 90.40 & 90.60 & 4.95 \\
& Robot w/o Gripper (Black) & 84.00 & 99.80 & 91.00 & 92.60 & 4.35 \\
& Robot w/o Gripper (Mosaic) & 91.20 & 98.80 & 91.20 & 96.40 & 1.80 \\
\cmidrule{2-7}
& Background (Black) & 73.80 & 47.40 & 83.20 & 72.20 & 27.05 \\
& Background (Mosaic) & 84.60 & 95.00 & 88.20 & 97.60 & 4.85 \\

\bottomrule
\end{tabular}
\caption{Average success rates (\%) across LIBERO datasets for $\pifive$, OpenVLA, and OpenVLA-OFT under each masking strategy.} \label{tab:patch_mask_avg_merged}
\end{table*}

Next, we investigate visual perturbations. In Tab.~\ref{tab:patch_mask_avg_merged}, target-object background replacement causes the largest average drop for $\pifive$ (76.45) and OpenVLA-OFT (60.25). For OpenVLA, full-robot background replacement is more destructive (69.42) than target-object masking (47.75). These results indicate that target objects act as causal visual anchors for $\pifive$ and OpenVLA-OFT, whereas OpenVLA depends more critically on full-robot visibility.

Further perturbations expose model-specific visual shortcut dependencies. For $\pifive$, masking the gripper or full robot with background replacement causes substantial performance degradation, whereas masking the robot without the gripper is much less destructive, suggesting a stronger reliance on gripper-centered interaction geometry. Conversely, OpenVLA exhibits a brittle dependence on full-robot visibility. In addition, background masking reduces average success for all models, indicating that policies exploit broader scene layouts. And further experimental details and results are provided in Appendix~\ref{app:appendix_perturbation_protocol}. Together, these results reveal that
\textit{VLA policies rely on target objects, robot-object interaction regions, and scene-layout cues as causal visual anchors for action execution, revealing strong visual grounding but also model-specific visual shortcut dependence \textbf{(Finding~6)}}.





\begin{table*}[t]
\centering
\small
\begin{tabular}{p{6cm}ccp{7cm}}
\toprule
\textbf{Original Prompt} & \textbf{Orig.} & \textbf{Edit} & \textbf{Edited Prompt} \\
\midrule
put both the alphabet soup and the tomato sauce in the basket 
& 55 & 60 
& put both the alphabet soup and the ketchup in the basket \\

put both the cream cheese box and the butter in the basket 
& 100 & 90 
& put both the cream cheese box and the milk in the basket \\

turn on the stove and put the moka pot on it 
& 75 & 30 
& turn on the stove and put the frypan on it \\

put the black bowl in the bottom drawer of the cabinet and close it 
& 90 & 30 
& put the white bottle in the bottom drawer of the cabinet and close it \\

put the white mug on the left plate and put the yellow and white mug on the right plate 
& 80 & 70 
& put the red coffee mug on the left plate and put the yellow coffee mug on the right plate \\

pick up the book and place it in the back compartment of the caddy 
& 90 & 30 
& pick up the mug and place it in the back compartment of the caddy \\

put the white mug on the plate and put the chocolate pudding to the right of the plate 
& 75 & 100 
& put the red mug on the plate and put the chocolate pudding to the right of the plate \\
\bottomrule
\end{tabular}
\caption{Success rate (\%) comparison between original and edited prompts.}
\label{tab:text_edit}
\end{table*}
\subsubsection{Input Editing}

Finally, we evaluate whether the model follows fine-grained vision-language instruction variations. 
For text editing, Tab.~\ref{tab:text_edit} shows that replacing target objects in instructions does not consistently redirect policy behavior. Some edits cause large performance drops (e.g., moka pot$\rightarrow$frypan, 75\%$\rightarrow$30\%), while others have limited effect. Overall, the results suggest that \textit{text exerts a non-negligible but limited influence on action execution, and the policy often remains biased toward the original visual-task configuration even when edited semantics dictate a behavioral shift \textbf{(Finding~7)}.}

\section{Discussion and Outlook}
\label{sec:discussion}

In this section, we summarize the main experimental findings and discuss their implications for future research directions in the VLA domain.

\paragraph{Representation preservation and embodied modality adaptation.} 
First, our CKA alignment experiment reveals that $\pifive$ and OpenVLA exhibit distinct modality-specific adaptation dynamics at different stages. $\pifive$ induces more fluctuating cross-modal fusion dynamics than OpenVLA, resulting in less stable alignment patterns across layers. In addition, $\pifive$ primarily reorganizes textual representations into task-conditioned control features, whereas OpenVLA mainly restructures visual and cross-modal fusion representations.

Future VLA training should balance preserving vision-language priors with adapting them for robotic control. Insufficient adaptation limits visuomotor learning, while excessive adaptation may degrade semantic grounding and increase reliance on visual shortcuts.

\paragraph{Designing causal visual-language circuits for action decoding.} 
Second, attention knockout experiments reveal distinct control dependencies: $\pifive$ primarily relies on visual-to-action pathways, whereas OpenVLA depends on both visual grounding and prompt-region access. Layer-wise interventions further show that OpenVLA distributes control information more broadly across layers, highlighting the importance of reliable multimodal pathways for action generation.

Future VLA architectures may benefit from explicit VLA routing modules that regulate when and how visual and textual features are passed to action tokens. For example, informed by analyses of encoder layer fusion in sequence-to-sequence models~\cite{liu2020understanding}, mid-to-late layer fusion gates could preserve high-level semantic constraints until the action generation stage, while action-conditioned cross-attention modules could dynamically select task-relevant visual regions and instruction tokens.

\paragraph{From visual grounding to compositional semantic control.} Third, the model's attention is both spatially grounded in manipulation-relevant regions and temporally aligned with the sequential structure of the instruction. Besides, the policy often remains biased toward the original visual-task configuration even when edited semantics dictate a behavioral shift. These findings reveal a limitation in fine-grained instruction following: current VLA models can capture high-level task semantics, yet struggle to reliably incorporate subtle semantic modifications into low-level control behavior. As a result, edited instructions may fail to induce consistent compositional behavioral changes, particularly when they conflict with dominant visual or task-specific priors.

A natural future direction is to train VLA models with semantic-edit objectives, counterfactual instruction-image pairs, object-swap interventions, and contrastive behavior supervision, so that small but meaningful language changes become causally actionable in low-level control.

\section{Conclusion}
\label{sec:conclusion}

We present VLA-Trace, a systematic framework for analyzing representation and control in VLA models. By combining CKA, attention knockouts, attention localization, visual patch masking, and input editing, VLA-Trace connects representation dynamics with causal action dependencies and rollout behavior. 
Experiments on $\pifive$ and OpenVLA reveal divergent adaptation strategies and a shared limitation: semantic control remains partial and architecture-dependent despite strong benchmark performance. These findings suggest that future VLA models should better preserve visual, language, and cross-modal semantic pathways, and integrate them more reliably into action decoding for robust compositional control.

\section*{Limitations}
\label{sec:limitations}

This study has several limitations. First, attention knockout is a controlled necessity probe, and its induced failures should not be directly equated with natural out-of-distribution failures. It reveals which pathways are required under a specific intervention design, but it does not define all forms of robustness. Complementary interventions such as activation patching and representation restoration instead test sufficiency and mediation, which we leave to future work. Second, CKA measures sample-level representation geometry and should not be interpreted as an absolute score of grounding, perception, or language understanding. Its values depend on the selected dataset, pooling strategy, and checkpoint comparison; therefore, our claims emphasize relative layer profiles and checkpoint drift rather than absolute CKA magnitudes. Third, the current analysis does not include random-feature or permutation CKA baselines. Fourth, attention IoU depends on simulator masks and patch resolution, so it is best interpreted as a conservative localization measure rather than a complete account of visual grounding. Fifth, VLM-style grounding probes can be confounded by whether a VLA checkpoint still supports natural language localization outputs; degraded text generation can reflect output-format mismatch as well as grounding degradation. Sixth, $\pifive$ C2 is evaluated as a shared LIBERO-finetuned checkpoint across multiple suites, so suite-level differences reflect evaluation distribution effects rather than separate per-suite weight changes. Finally, this work focuses its complete evidence chain on $\pifive$ and OpenVLA; our broader implementation protocol defines extensions to X-VLA, OpenVLA-OFT, Calvin, SimplerEnv, LIBERO-Plus/Pro, and RoboTwin, but future work should complete these settings before making broader claims across the full VLA design space.
\section*{Ethics Statement}
This work analyzes existing VLA models using publicly available models and evaluation resources, without collecting personal data or involving human-subject studies. While our framework is intended to improve transparency and diagnosis, the analyzed models may still inherit biases, safety risks, and failure modes from their training data and design; therefore, real-world deployment requires careful evaluation, human oversight, and appropriate safety constraints.

\bibliography{acl}

\appendix
\section{Implementation Protocol Details}
\label{sec:appendix_sop}

\subsection{Model Interfaces and Input Templates}
\label{app:appendix_model_interfaces}
The VLA-Trace protocol tracks several VLA model families, including $\pifive$, X-VLA, OpenVLA, and OpenVLA-OFT. 
The main paper focuses on $\pifive$ and OpenVLA because they instantiate two distinct action-generation interfaces. 
$\pifive$ builds on a PaliGemma-style VLM and uses an action expert with flow-matching action generation. 
In the evaluated configuration, the transformer input contains visual tokens from the image observation, a BOS token, the task instruction, and a newline token. 
Although the original model interface includes proprioceptive state fields, the analyzed representation stream does not tokenize proprioceptive state into the transformer prompt. 
OpenVLA instead formulates action prediction autoregressively over action tokens, using image patches followed by a prompt template of the form \textit{``In: What action should the robot take to \{instruction\}? Out:''}. 
This interface distinction is important for rollout and attention-knockout experiments, where semantic instruction tokens must be separated from structural tokens such as BOS, newline markers, and system prompts. 
OpenVLA-OFT extends OpenVLA by replacing autoregressive action-token generation with continuous action regression during fine-tuning. 
This modification changes the action prediction interface and reduces dependence on sequential token decoding, making OpenVLA-OFT a useful supplementary setting for comparing how decoding paradigms influence representation dynamics and causal pathways.

\subsection{CKA Protocol}
\label{app:cka}
\paragraph{Prompt templates for CKA probes.}
For representation CKA, we use separate neutral text templates rather than the action-generation prompts used during rollout. 
This choice makes the open-domain and robot-domain probes share a comparable image-description form and avoids conflating representational alignment with action-query formatting. 
For COCO, captions are rendered as \texttt{A photo of \textless caption\textgreater}. 
For LIBERO, task instructions are rendered as \texttt{A photo of a robot \textless instruction\textgreater}, where the instruction is inserted after the word ``robot''. 
For OpenVLA, the same neutral textual content is wrapped with the model-required input format, yielding \textit{``In: ... Out:''}; for $\pifive$, the neutral text is used directly, with the PaliGemma loader adding the required image token internally. 
These templates are used only for the representation-alignment probe and should not be confused with the action-generation prompt templates used by $\pifive$ or OpenVLA during policy rollout and knockout experiments.

\paragraph{Pooled token views.}
For each sample and transformer layer, we extract hidden states over the full image--text sequence and partition them into visual and textual spans according to the model-specific token layout. 
Let $h^{\ell}_{i,t}$ denote the hidden state of token $t$ at layer $\ell$ for sample $i$, and let $\mathcal{I}$ and $\mathcal{T}$ denote the image-token and text-token index sets. 
We define three pooled views:
\[
v_i^\ell = \frac{1}{|\mathcal{I}|}\sum_{t\in\mathcal{I}} h^\ell_{i,t},
\]

\[
q_i^\ell = \frac{1}{|\mathcal{T}|}\sum_{t\in\mathcal{T}} h^\ell_{i,t},
\]

\[
j_i^\ell = \frac{1}{|\mathcal{I}|+|\mathcal{T}|}\sum_{t\in\mathcal{I}\cup\mathcal{T}} h^\ell_{i,t}.
\]
We refer to these vectors as \texttt{vision\_pooled}, \texttt{text\_pooled}, and \texttt{joint\_pooled}, respectively. 
Stacking them over all probe samples gives layer-wise representation matrices 
$V^\ell, Q^\ell, J^\ell \in \mathbb{R}^{N\times d}$, where $N$ is the number of probe examples and $d$ is the hidden dimension. 
For OpenVLA, the image span corresponds to the image-patch tokens and the text span corresponds to the subsequent textual prompt tokens. 
For PaliGemma-style $\pifive$ C0 extraction, the text span skips the BOS token following the image tokens.

\paragraph{CKA computation.}
We compute linear CKA over sample-wise pooled representations. 
Given two representation matrices $X,Y\in\mathbb{R}^{N\times d}$, we first center them along the sample dimension and compute
\[
\mathrm{CKA}(X,Y)=
\frac{\|X_c^\top Y_c\|_F^2}
{\|X_c^\top X_c\|_F \, \|Y_c^\top Y_c\|_F}.
\]
For layer-wise image--text alignment, we compute 
$\mathrm{CKA}(V^\ell, Q^\ell)$ independently at each layer $\ell$ within the same checkpoint. 
For checkpoint-drift analysis, we compare matched layers across two checkpoints for each pooled view:
\[
\mathrm{CKA}(V_a^\ell,V_b^\ell), \quad
\mathrm{CKA}(Q_a^\ell,Q_b^\ell), \quad
\mathrm{CKA}(J_a^\ell,J_b^\ell),
\]
and report the mean diagonal CKA over layers. 
Thus, high checkpoint-drift CKA indicates preservation of the corresponding representation subspace, whereas low CKA indicates stronger representational reorganization.


\subsection{Attention Knockout Protocol}
\label{sec:appendix_knockout_protocol}
We use attention knockout to test whether specific token pathways are causally required for action generation. All interventions are implemented by adding a large negative mask to selected attention logits before the softmax, while keeping model weights, input observations, and decoded action heads unchanged. For an attention layer with queries $Q$, keys $K$, values $V$, and hidden dimension $d$, the modified attention is
\begin{equation}
    \mathrm{Attn}(Q,K,V;M)
    =
    \mathrm{softmax}\left(\frac{QK^\top}{\sqrt{d}} + M\right)V,
\end{equation}
where $M_{ij}=0$ preserves attention from query token $i$ to key token $j$, and $M_{ij}=-\infty$ blocks that attention edge.

\paragraph{Intervention granularity.}
We evaluate three levels of knockout. 
First, \textit{all-layer knockout} applies the same source-blocking rule to every transformer layer, measuring the global necessity of a pathway. 
Second, \textit{single-layer knockout} applies the mask only at one target layer, measuring whether the dependency is localized. 
Third, to reduce the possibility that information bypasses a single blocked layer through residual connections, we use \textit{centered window knockout}. For a target layer $\ell$ and window size $w$, $w=3$ blocks layers $\ell-1$ to $\ell+1$, $w=5$ blocks $\ell-2$ to $\ell+2$, and $w=7$ blocks $\ell-3$ to $\ell+3$. We use these windows to distinguish distributed pathways from localized bottlenecks: if all-layer removal is destructive but every local window is mild, the pathway is likely distributed; if a narrow window causes a large drop, the corresponding layer range is treated as a candidate causal bottleneck.

\paragraph{Token partitions.}
The knockout masks are defined over role-specific token sets. 
Let $\mathcal{V}$ denote visual tokens, $\mathcal{L}$ semantic language tokens, $\mathcal{S}$ structural prompt tokens, and $\mathcal{A}$ action-query or action-token positions.
For $\pifive$, $\mathcal{V}$ contains visual tokens from the agent and wrist views, $\mathcal{L}$ contains the task-instruction tokens, $\mathcal{S}$ contains BOS and newline tokens, and $\mathcal{A}$ contains action-query or action-slot tokens used by the action expert. 
For OpenVLA and OpenVLA-OFT, $\mathcal{V}$ contains image patch tokens, $\mathcal{L}$ contains the natural-language task span, $\mathcal{S}$ contains system prompt and special tokens such as \textit{``In:''}, \textit{``Out:''}, question text, newline, BOS, and EOS where applicable, and $\mathcal{A}$ contains autoregressive action tokens.

\paragraph{Prefill-stage knockout.}
We separate inference into a \texttt{prefill} stage and a \texttt{generation} stage. 
The \texttt{prefill} stage builds the internal multimodal context before action decoding. 
For $\pifive$, whose VLM backbone uses bidirectional attention, we test N0 V--L prefill blocking by preventing visual and language tokens from directly attending to each other. 
With token groups ordered as $[\mathcal{L};\mathcal{V};\mathcal{A}]$, this can be represented as
\begin{equation}
M^\ell_{\mathrm{prefill:no\text{-}V\text{-}L}}
=
\begin{bmatrix}
\mathbf{LL} & -\infty & -- \\
-\infty & \mathbf{VV} & -- \\
\mathbf{AL} & \mathbf{AV} & \mathbf{AA}
\end{bmatrix},
\end{equation}
where unmasked blocks preserve the original attention pattern and $-\infty$ blocks suppress direct vision--language exchange. 
For OpenVLA-style autoregressive models, the main \texttt{prefill} intervention removes image access during context construction, testing whether visual grounding must be established before action-token decoding.

\paragraph{Generation-stage knockout.}
In the generation stage, action tokens are decoded from the prefilled multimodal context. 
We perform modality knockout by blocking attention from action positions to selected source tokens, thereby preventing the decoder from conditioning on a specific modality during action prediction. 
This intervention measures the extent to which generation depends on visual tokens, semantic instruction tokens, structural prompt tokens, or their combinations.

Let the token groups be ordered as $[\mathcal{L};\mathcal{V};\mathcal{A}]$, where $\mathcal{L}$ denotes language tokens, $\mathcal{V}$ denotes visual tokens, and $\mathcal{A}$ denotes action tokens. 
For example, the generation-time no-text intervention blocks action-to-language attention while preserving the remaining attention structure:
\begin{equation}
M^\ell_{\mathrm{gen:no\text{-}text}}
=
\begin{bmatrix}
\mathbf{LL} & \mathbf{LV} & -- \\
\mathbf{VL} & \mathbf{VV} & -- \\
-\infty & \mathbf{AV} & \mathbf{AA}
\end{bmatrix}.
\end{equation}
Here, the $-\infty$ block suppresses attention from action queries to language keys. 
Analogously, the no-image intervention replaces the $\mathbf{AV}$ block with $-\infty$, blocking action-to-vision attention:
\begin{equation}
M^\ell_{\mathrm{gen:no\text{-}image}}
=
\begin{bmatrix}
\mathbf{LL} & \mathbf{LV} & -- \\
\mathbf{VL} & \mathbf{VV} & -- \\
\mathbf{AL} & -\infty & \mathbf{AA}
\end{bmatrix}.
\end{equation}
For autoregressive OpenVLA-style models, these knockout masks are applied on top of the original causal mask, so they only remove source positions that would otherwise be visible to the current action token.

\begin{figure*}[ht!]
    \centering
    \includegraphics[width=\textwidth]{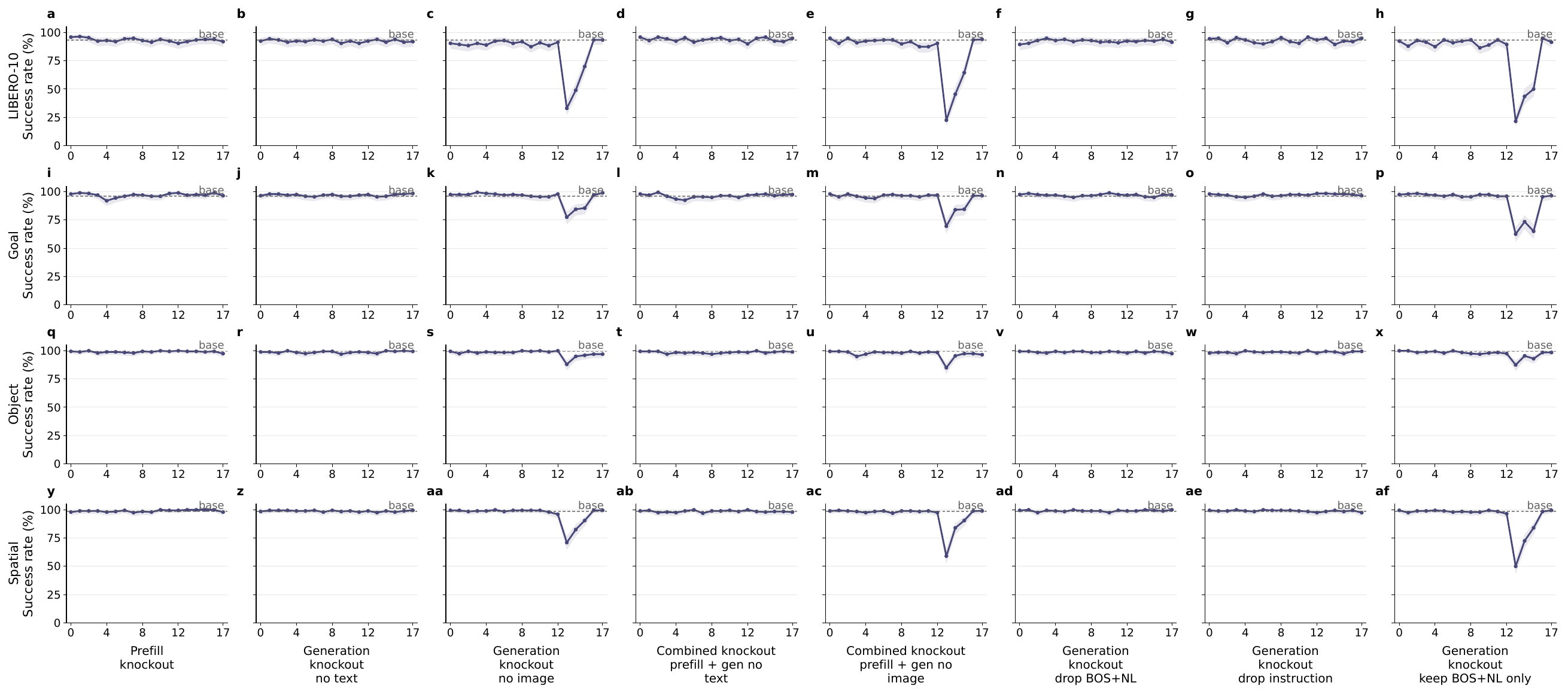}
    \caption{
   Layer-wise knockout results for \textbf{$\pifive$} on LIBERO-10, Goal, Object, and Spatial. Each point reports the success rate under a \textbf{3-layer} knockout window centered at the indicated layer.}
    \vspace{-8pt}
    \label{fig:pi05_layerwise_window3}
\end{figure*}

\begin{figure*}[ht!]
    \centering
    \includegraphics[width=\textwidth]{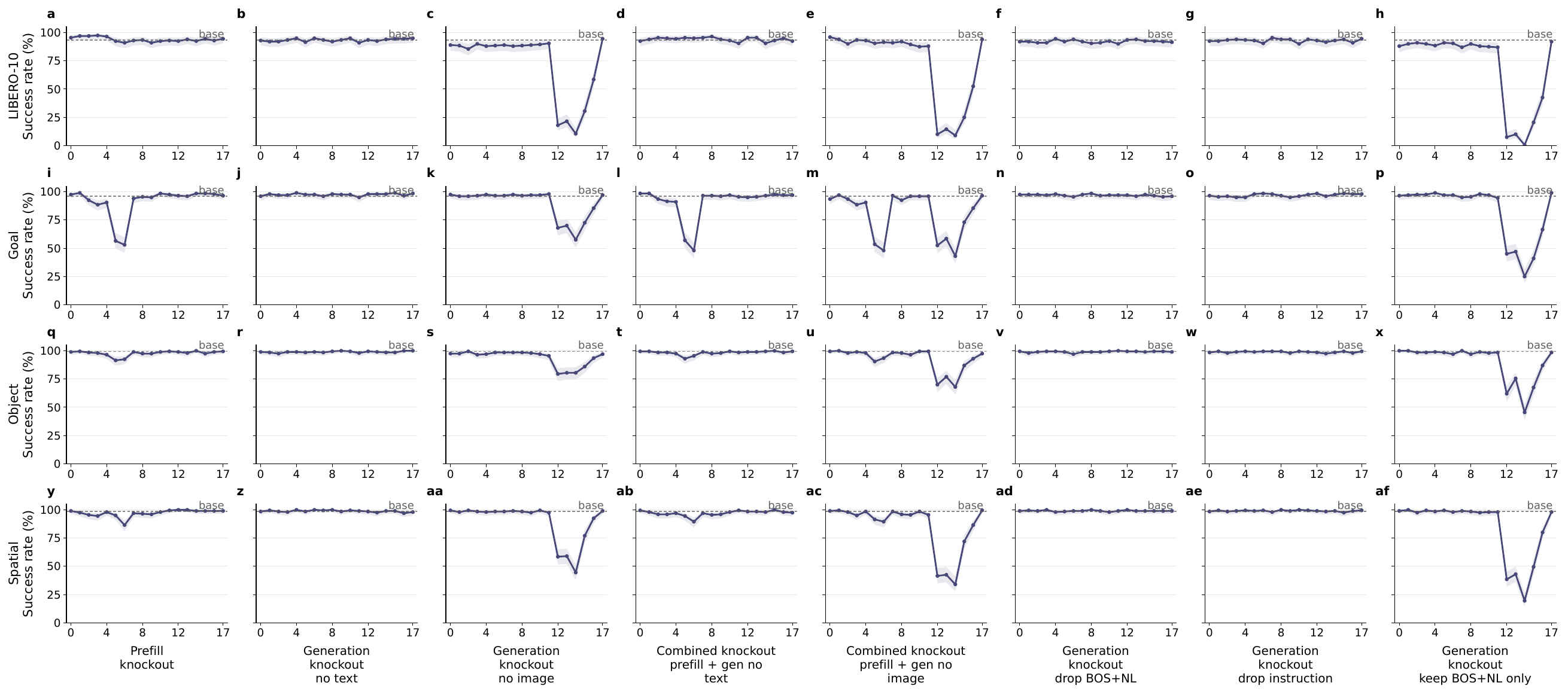}
    \caption{
    Layer-wise knockout results for \textbf{$\pifive$} on LIBERO-10, Goal, Object, and Spatial. Each point reports the success rate under a \textbf{5-layer} knockout window centered at the indicated layer.}
    \vspace{-8pt}
    \label{fig:pi05_layerwise_window5}
\end{figure*}

\begin{figure*}[ht!]
    \centering
    \includegraphics[width=\textwidth]{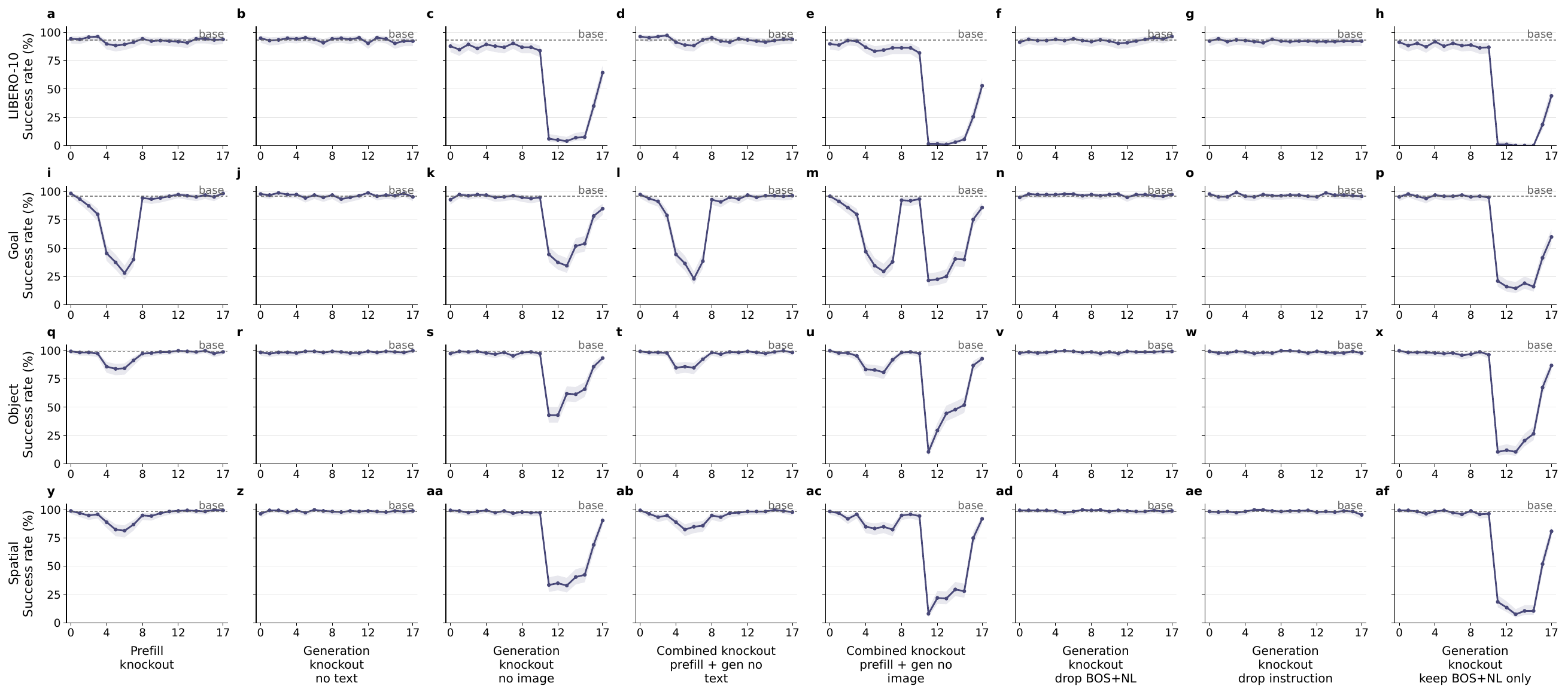}
    \caption{
   Layer-wise knockout results for \textbf{$\pifive$} on LIBERO-10, Goal, Object, and Spatial. Each point reports the success rate under a \textbf{7-layer} knockout window centered at the indicated layer.}
    \vspace{-8pt}
    \label{fig:pi05_layerwise_window7}
\end{figure*}

\begin{figure*}[ht!]
    \centering
    \includegraphics[width=\textwidth]{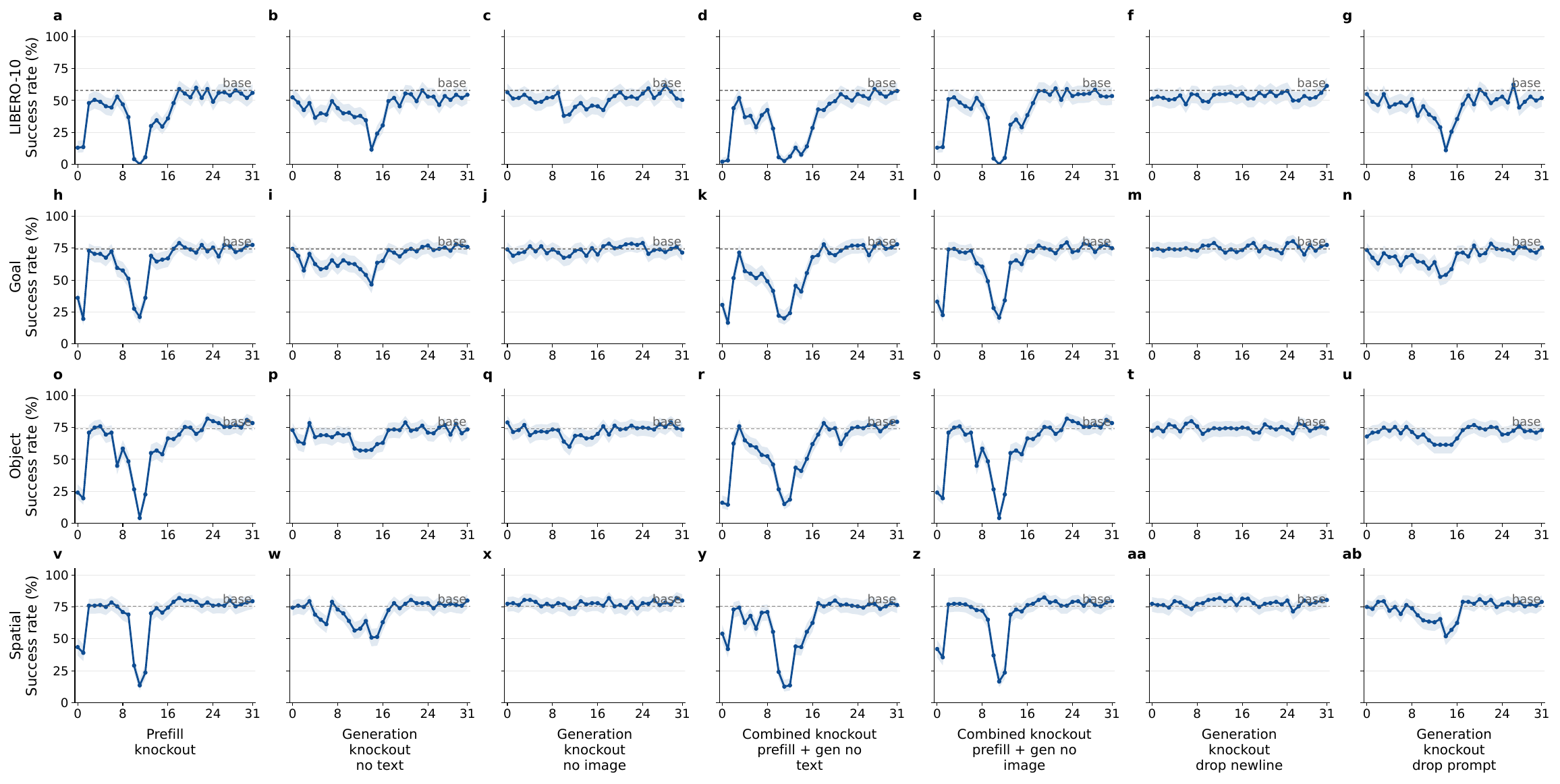}
    \caption{
   Layer-wise knockout results for \textbf{OpenVLA} on LIBERO-10, Goal, Object, and Spatial. Each point reports the success rate under a \textbf{3-layer} knockout window centered at the indicated layer.}
    \label{fig:openvla_layerwise_window3}
\end{figure*}

\begin{figure*}[ht!]
    \centering
    \includegraphics[width=\textwidth]{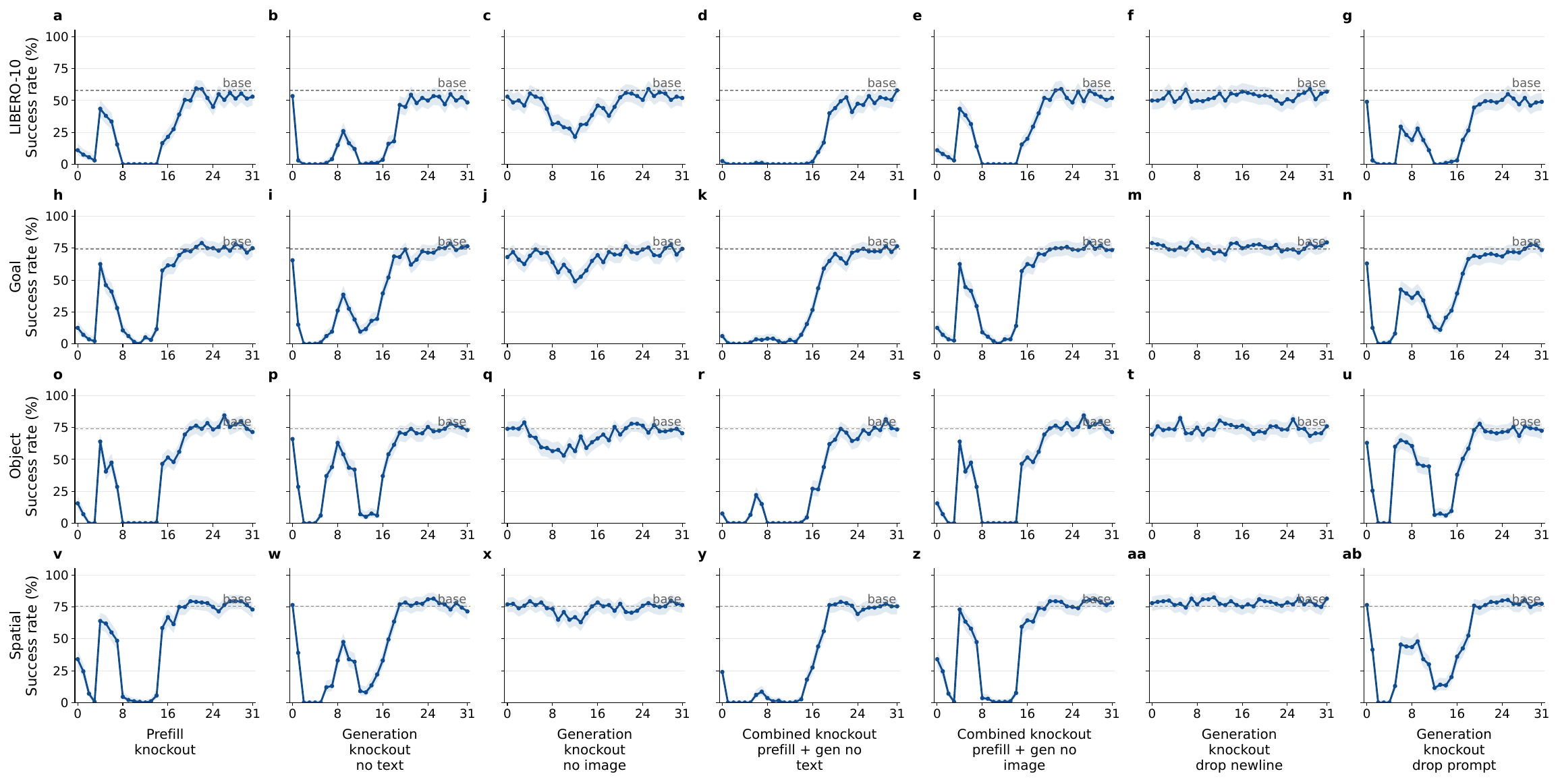}
    \caption{
   Layer-wise knockout results for \textbf{OpenVLA} on LIBERO-10, Goal, Object, and Spatial. Each point reports the success rate under a \textbf{7-layer} knockout window centered at the indicated layer.}
    \label{fig:openvla_layerwise_window7}
\end{figure*}

\begin{figure*}[ht!]
    \centering
    \includegraphics[width=\textwidth]{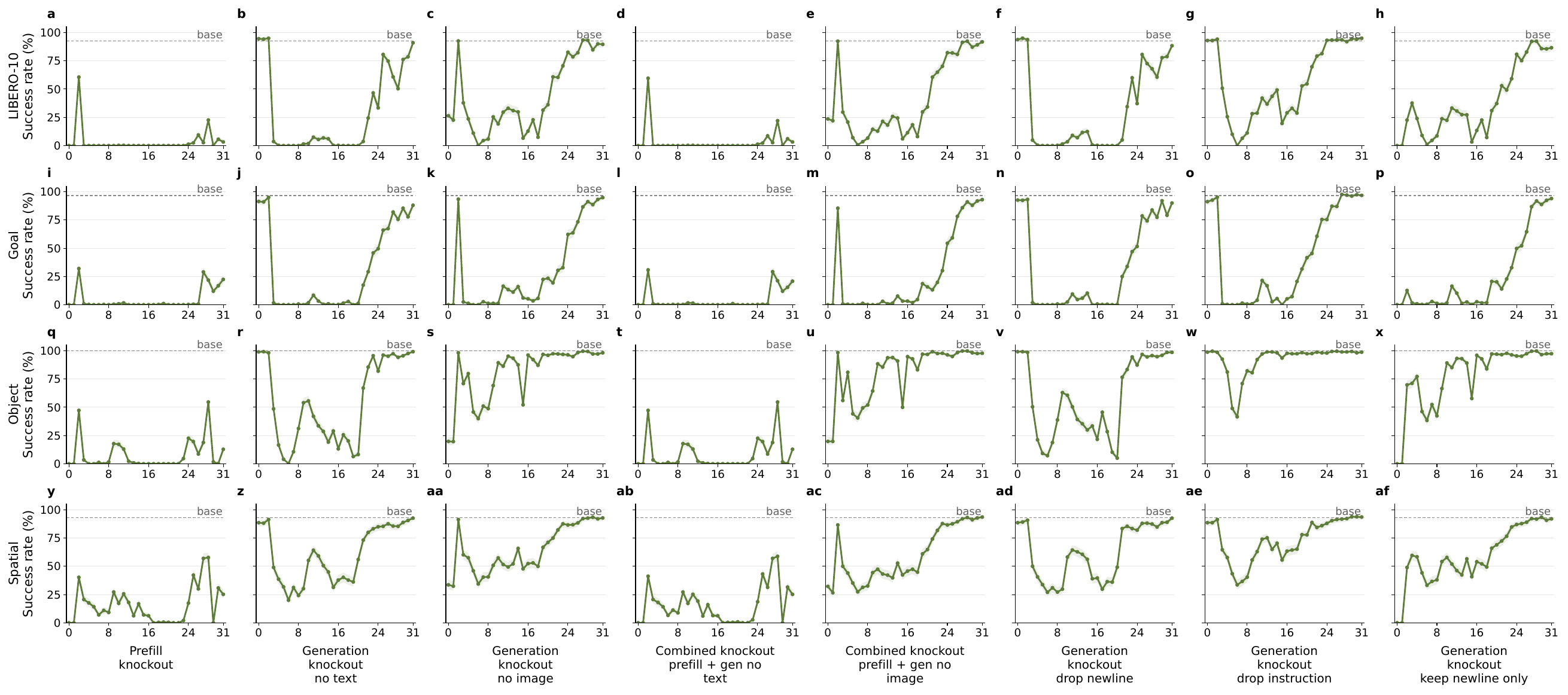}
    \caption{
   Layer-wise knockout results for \textbf{OpenVLA-OFT} on LIBERO-10, Goal, Object, and Spatial. Each point reports the success rate under a \textbf{3-layer} knockout window centered at the indicated layer.}
    \vspace{-8pt}
    \label{fig:openvla_oft_layerwise_window3}
\end{figure*}

\begin{figure*}[ht!]
    \centering
    \includegraphics[width=\textwidth]{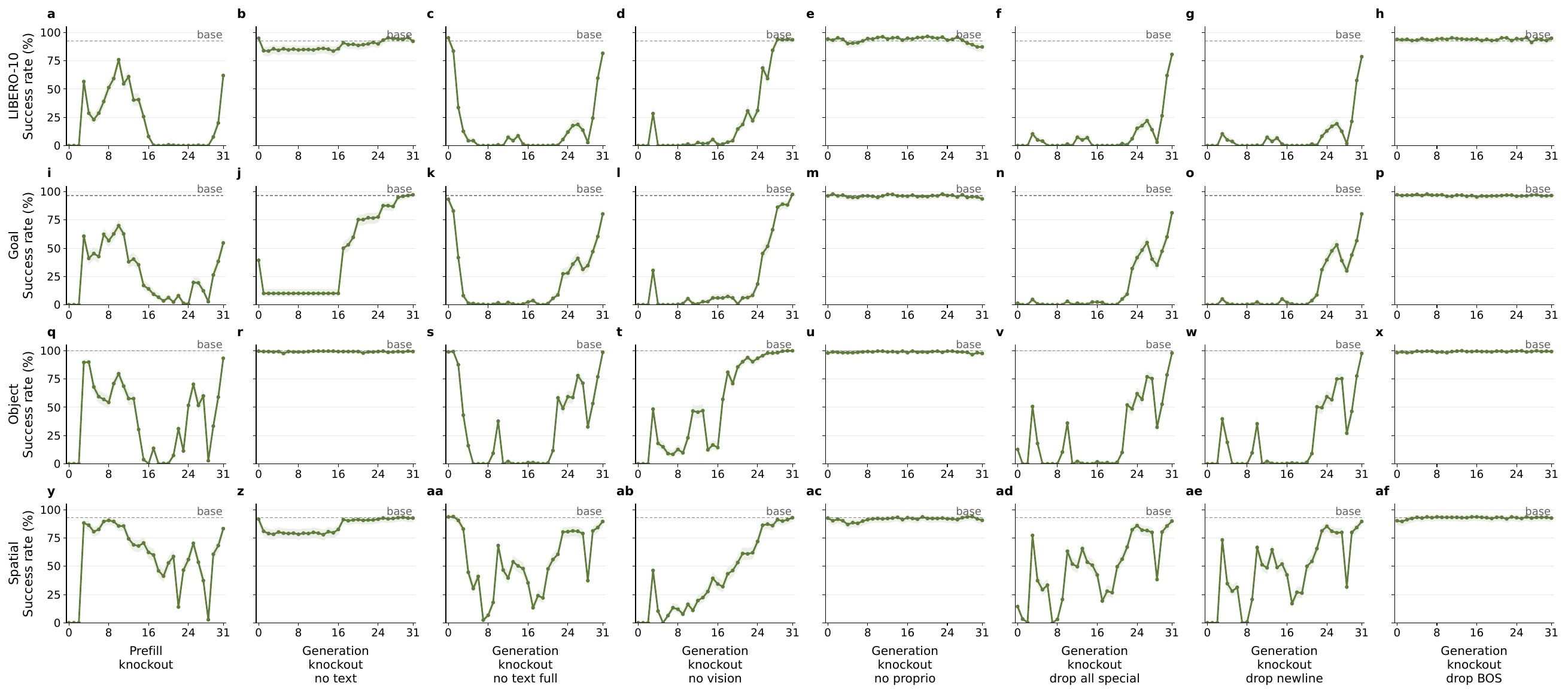}
    \caption{
   Layer-wise knockout results for \textbf{OpenVLA-OFT} on LIBERO-10, Goal, Object, and Spatial. Each point reports the success rate under a \textbf{5-layer} knockout window centered at the indicated layer.}
    \vspace{-8pt}
    \label{fig:openvla_oft_layerwise_window5}
\end{figure*}

\begin{figure*}[ht!]
    \centering
    \includegraphics[width=\textwidth]{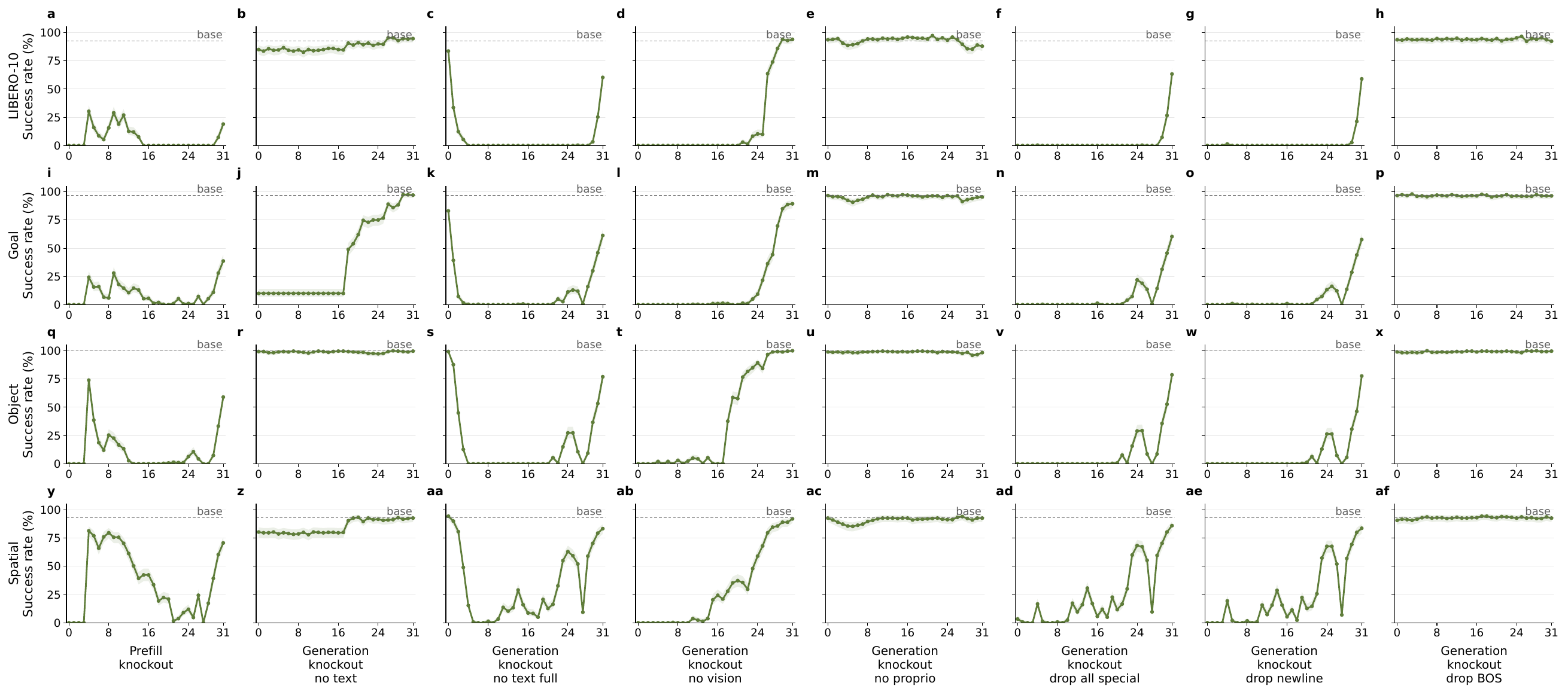}
    \caption{
   Layer-wise knockout results for \textbf{OpenVLA-OFT} on LIBERO-10, Goal, Object, and Spatial. Each point reports the success rate under a \textbf{7-layer} knockout window centered at the indicated layer.}
    \vspace{-8pt}
    \label{fig:openvla_oft_layerwise_window7}
\end{figure*}

\begin{figure*}[ht!]
    \centering
    \includegraphics[width=\textwidth]{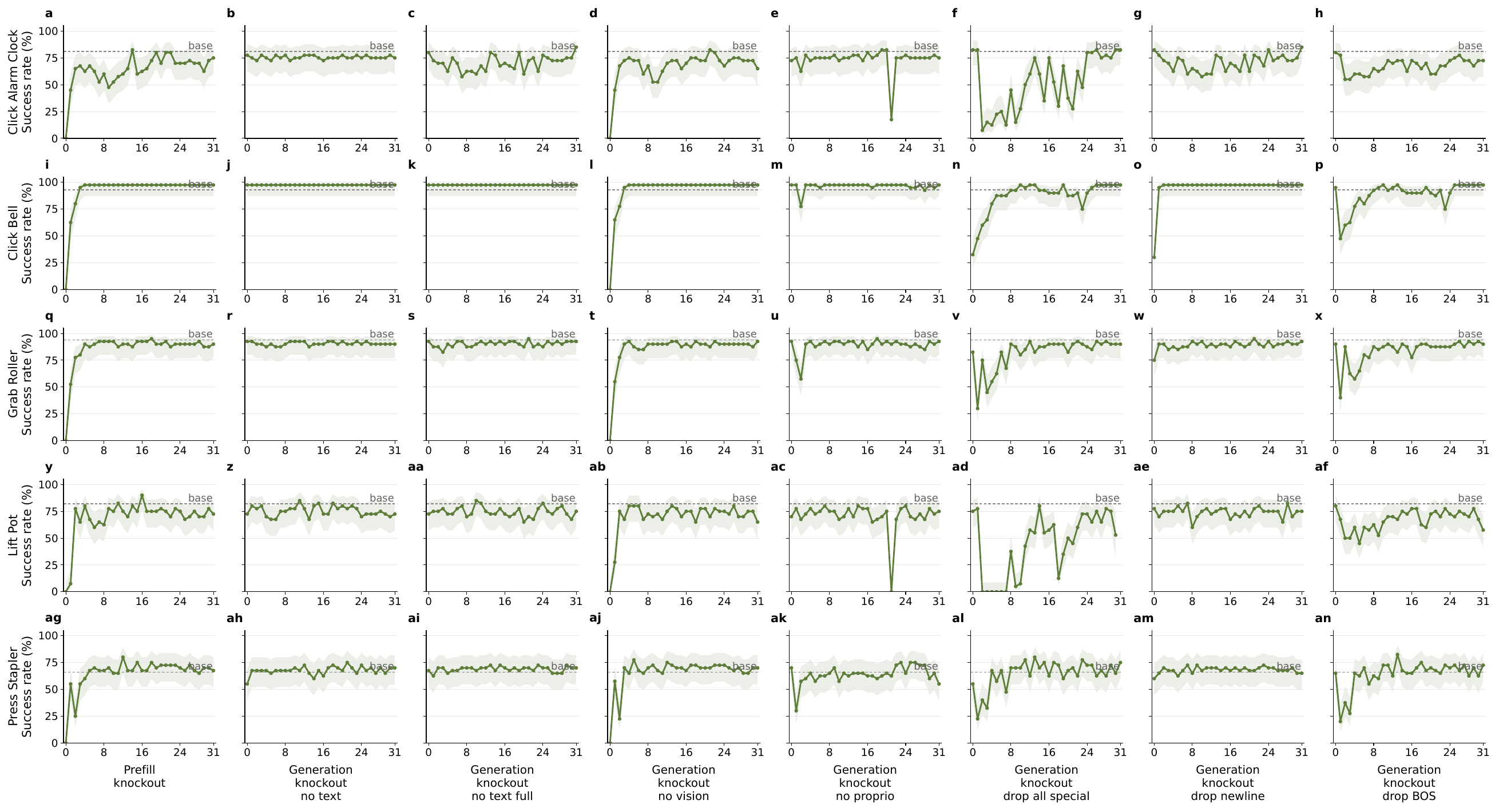}
    \caption{
    Layer-wise knockout results for \textbf{OpenVLA-OFT} on RoboTwin tasks. Each row corresponds to one RoboTwin task, and each point reports the success rate under a \textbf{1-layer} knockout window at the indicated layer.}
    \label{fig:openvla_oft_layerwise_robotwin_window1}
\end{figure*}

\begin{figure*}[ht!]
    \centering
    \includegraphics[width=\textwidth]{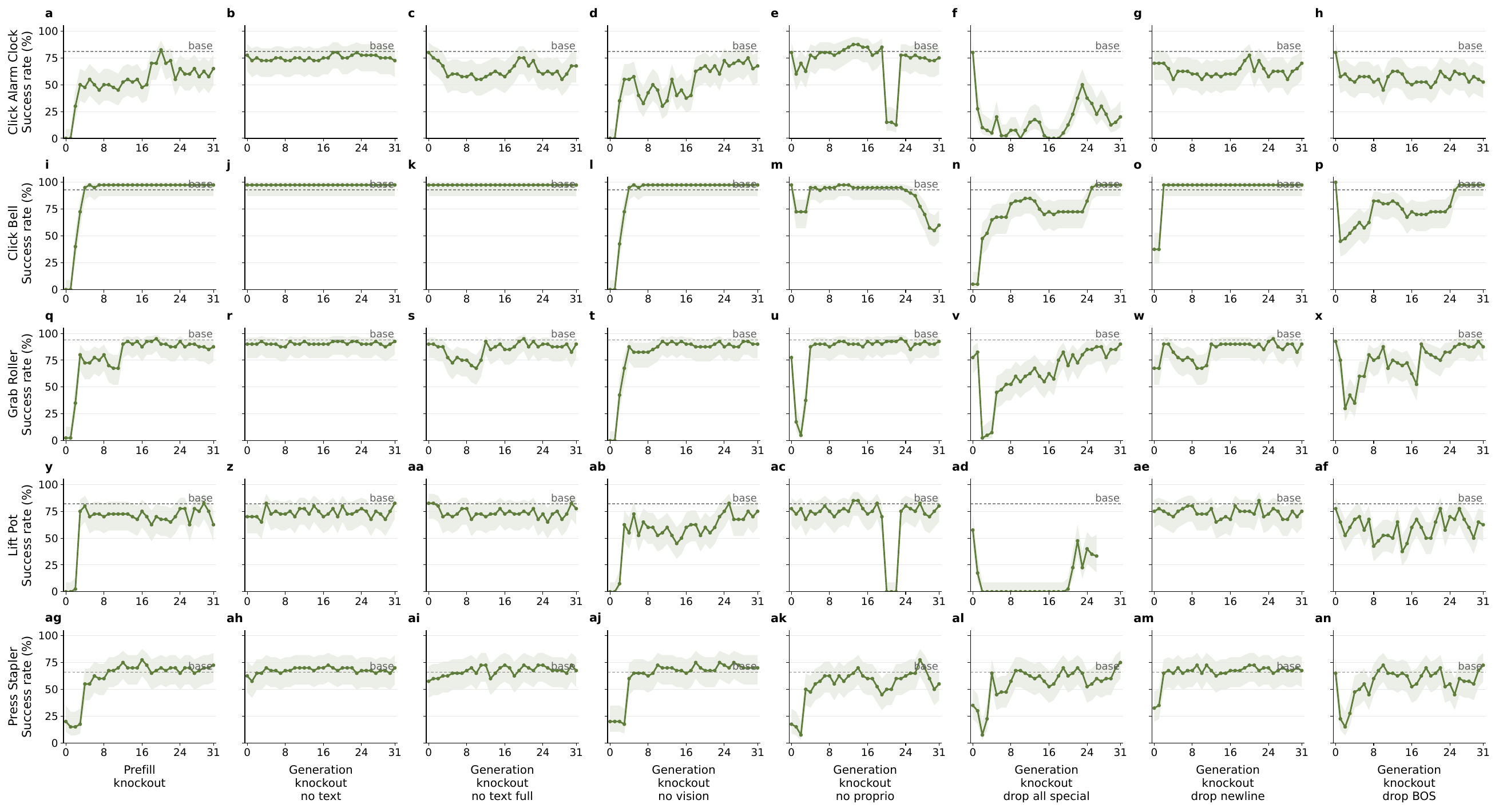}
    \caption{
    Layer-wise knockout results for \textbf{OpenVLA-OFT} on RoboTwin tasks. Each row corresponds to one RoboTwin task, and each point reports the success rate under a \textbf{3-layer} knockout window at the indicated layer.}
    \label{fig:openvla_oft_layerwise_robotwin_window3}
\end{figure*}

\paragraph{Additional experimental results.}
Beyond the all-layer knockout results reported in the main paper, we provide full layer-wise (Tab.~\ref{tab:attention_knockout_all_layers_summary_0526}) and windowed knockout sweeps to assess whether each dependency is distributed across the network or concentrated in a narrow layer range. 
We use all-layer knockout to establish global pathway necessity, layer-wise and centered-window knockout to localize vulnerable regions, and semantic/structural token splits to avoid conflating task-language grounding with dependence on prompt-formatting tokens. 
Fig.~\ref{fig:pi05_layerwise_window3}--\ref{fig:pi05_layerwise_window7} report $\pifive$ the results of the four LIBERO suites under 3-, 5-, and 7-layer windows; Figures~\ref{fig:openvla_layerwise_window3}--\ref{fig:openvla_layerwise_window7} report OpenVLA's results; Fig.~\ref{fig:openvla_oft_layerwise_window3}--\ref{fig:openvla_oft_layerwise_window7} extend the analysis to OpenVLA-OFT; and Fig.~\ref{fig:openvla_oft_layerwise_robotwin_window1}--\ref{fig:openvla_oft_layerwise_robotwin_window3} show OpenVLA-OFT results on RoboTwin tasks. 
Together, these results provide consistent evidence about whether action generation depends primarily on visual tokens, semantic language tokens, structural prompt tokens, prefill-stage visual grounding, or localized layer bottlenecks across model families and evaluation suites.

\subsection{Attention Localization Metrics and Visualizations}
\label{app:appendix_attention_metrics}
We complement the knockout analysis with attention-localization diagnostics that measure where action-relevant attention is spatially allocated during policy execution. All localization metrics are computed on simulator-aligned image patch grids. For each rollout step, we extract attention from action-query or action-token positions to image patches and average over action positions to obtain an action-conditioned heatmap $\bar{A}$. Let $\bar{A}_{j}$ denote the attention assigned to image patch $j$, and let $M$ denote a simulator-derived binary mask for task-relevant objects, gripper/robot regions, or their union. We report three complementary localization metrics.

First, continuous attention mass measures the fraction of action attention assigned to the target region:
\begin{equation}
\mathrm{Mass}(M) =
\frac{\sum_{j \in M} \bar{A}_{j}}
{\sum_{j} \bar{A}_{j}}.
\end{equation}

Second, to evaluate whether high-confidence attention regions spatially overlap with task-relevant regions, we compute a 90th-percentile thresholded attention IoU. 
Let
\begin{equation}
\tau_{90} = q_{0.9}(\bar{A}),
\end{equation}
where $q_{0.9}(\bar{A})$ denotes the 90th percentile of the heatmap values. 
We define the high-attention patch set as
\begin{equation}
\mathcal{H}_{90}(\bar{A})
=
\{j:\bar{A}_{j} \ge \tau_{90}\}.
\end{equation}
The thresholded IoU is then
\begin{equation}
\mathrm{IoU}_{90}(M)
=
\frac{
|\mathcal{H}_{90}(\bar{A}) \cap M|
}{
|\mathcal{H}_{90}(\bar{A}) \cup M|
}.
\end{equation}

Third, peak-hit rate measures whether the single maximum-attention patch falls inside the target mask:
\begin{equation}
\mathrm{Hit}(M)
=
\mathbf{1}\!\left[\arg\max_j \bar{A}_{j} \in M\right].
\end{equation}

These metrics capture different aspects of visual grounding: $\mathrm{Mass}$ measures total attention allocation to a region, $\mathrm{IoU}_{90}$ measures spatial overlap between the target mask and attention regions above the 90th-percentile threshold, and $\mathrm{Hit}$ measures whether the strongest attended patch is task-relevant.

\paragraph{Temporal phase split.}
LIBERO-10 tasks often involve long-horizon instructions with multiple sequential subgoals. To analyze whether visual attention evolves with the temporal structure of the task, we divide each rollout into two phases using the first and second halves of the executed trajectory. This split provides a simple and consistent proxy for early versus late subgoals without requiring manual step-level annotation. We compute all localization metrics separately for each phase and for each mask type.

\begin{figure*}[ht!]
    \centering
    \includegraphics[width=\textwidth]{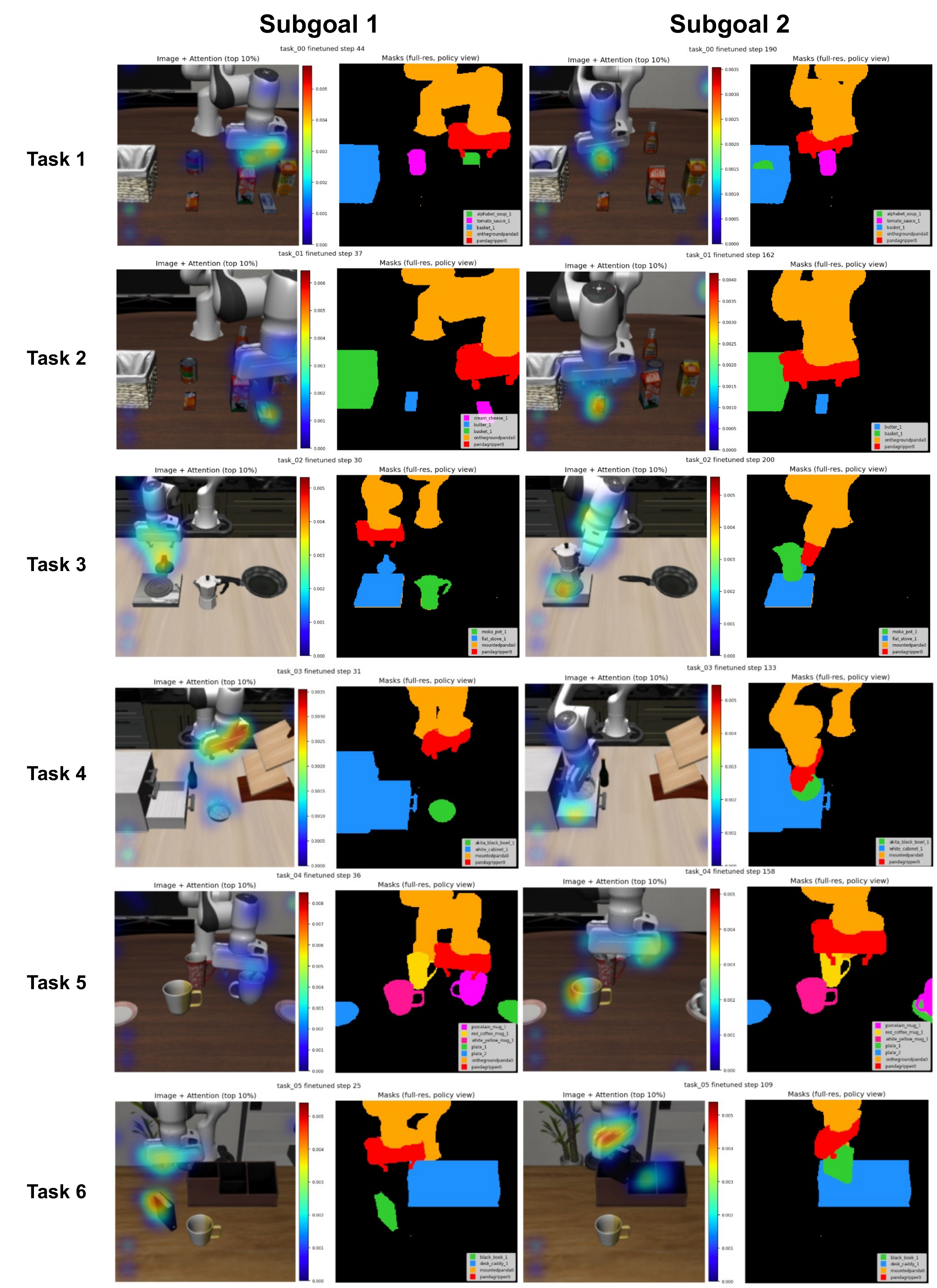}
    \caption{
    Schematic illustration of attention-IoU computation. We project action-conditioned attention onto the image patch grid, keep the top 10\% highest-attention patches, and compute IoU against a simulator-derived target mask. 
    The same mask is also used to compute continuous attention mass and peak-hit rate.
    }
    \label{fig:attention_iou_schematic}
\end{figure*}

\paragraph{IoU visualization schematic.}
To make the localization metrics visually interpretable, we include a schematic visualization of the IoU computation in Fig.~\ref{fig:attention_iou_schematic}. 
The action-conditioned attention map is first projected onto the image patch grid. 
We then threshold the map at the 90th percentile to obtain the high-attention patch set $\mathcal{H}_{90}(\bar{A})$ and compare it with the simulator-derived target mask $M$. The overlap region contributes to the numerator of $\mathrm{IoU}_{90}$, while the union of high-attention and target patches forms the denominator.

\begin{figure*}[ht!]
    \centering
    \includegraphics[width=\textwidth]{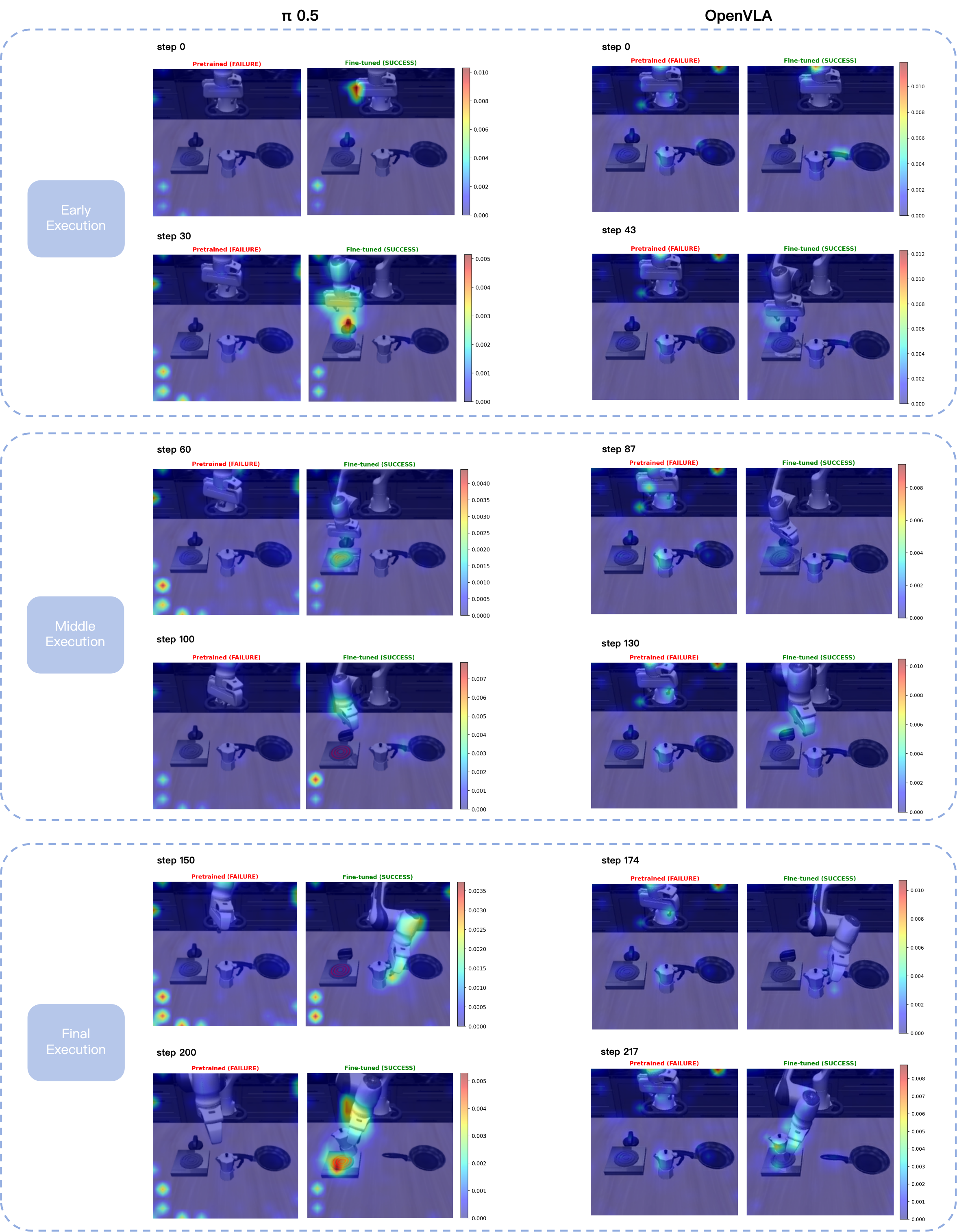}
    \caption{Qualitative action-to-image attention visualizations across rollout stages. We compare pretrained and fine-tuned $\pifive$ and OpenVLA models across early, middle, and final execution stages. Finetuning shifts action attention from diffuse or background-biased regions toward task-relevant robot-object interaction regions, indicating stronger visually grounded action generation during manipulation.}
    \label{fig:a2i}
\end{figure*}

\begin{figure*}[ht!]
    \centering
    \includegraphics[width=\textwidth]{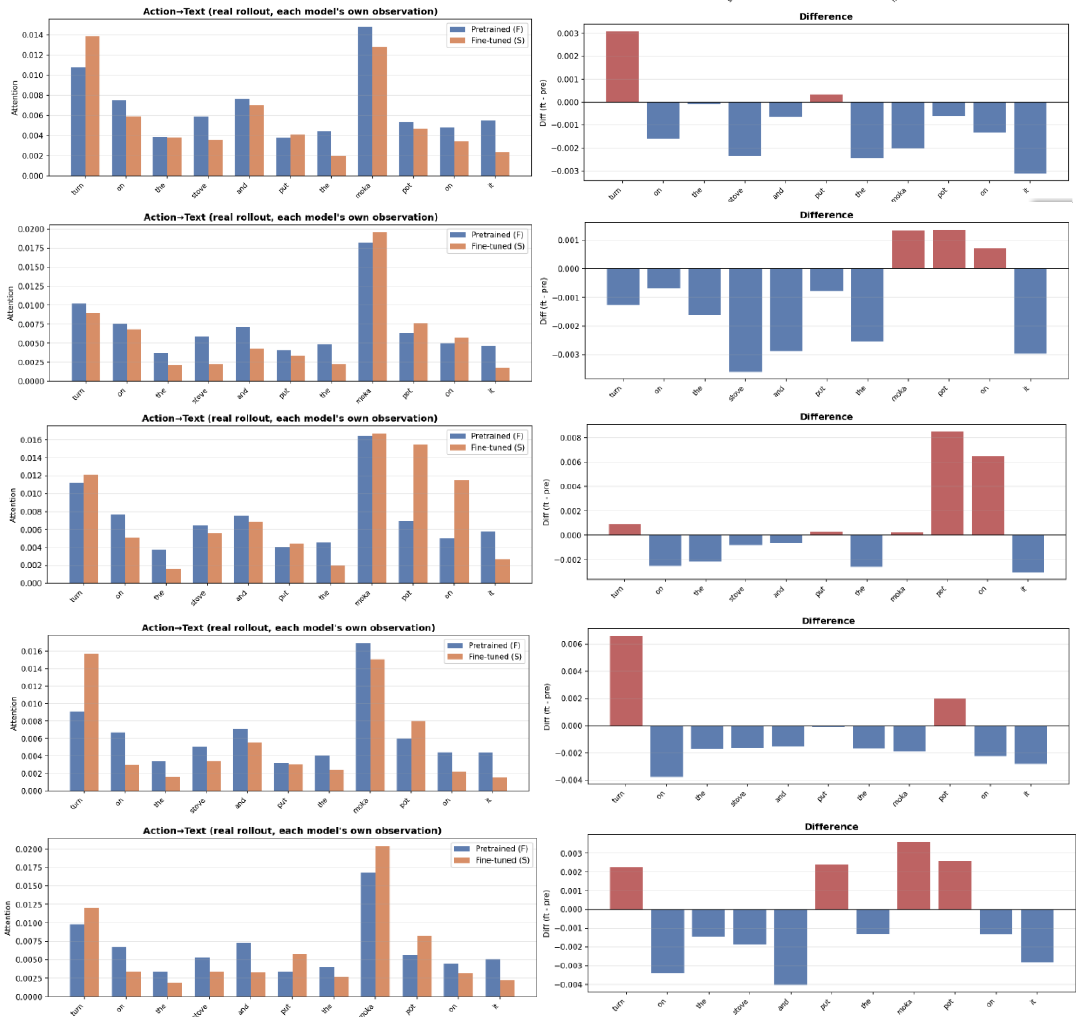}
    \caption{Visualization of action-to-text attention at different timesteps (top to bottom: steps 0, 30, 60, 100, and 250).}
    \label{fig:a2i_1}
\end{figure*}

\begin{figure*}[ht!]
    \centering
    \includegraphics[width=\textwidth]{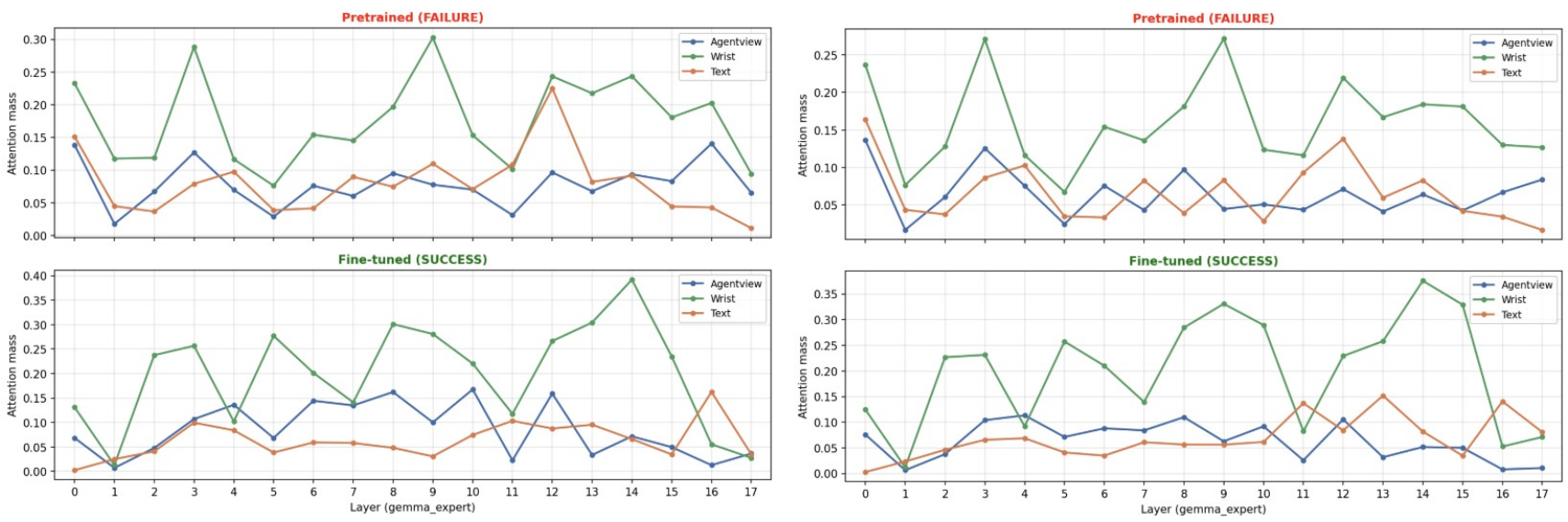}
    \caption{Visualization of layer-wise modality attention (left: step 30, right: step 150) (Top: pretrained, bottom: fine-tuned).}
    \label{fig:a2i_4}
\end{figure*}

\paragraph{Qualitative action-to-image attention.}
In addition to the quantitative localization metrics, we visualize action-conditioned image attention across execution stages. Fig.~\ref{fig:a2i} compares pretrained and fine-tuned $\pifive$ and OpenVLA checkpoints over early, middle, and final rollout steps. And Figures~\ref{fig:a2i_1} and~\ref{fig:a2i_4} provide more qualitative attention visualizations, showing the temporal evolution of action-to-text attention and the layer-wise redistribution of modality attention before and after fine-tuning. These visualizations are intended as qualitative complements to the mask-based metrics, which show how action attention moves over the scene during manipulation, while the IoU, mass, and hit-rate metrics quantify whether this attention overlaps with simulator-derived task regions. 

\begin{figure*}[ht!]
    \centering
    \includegraphics[width=0.97\textwidth]{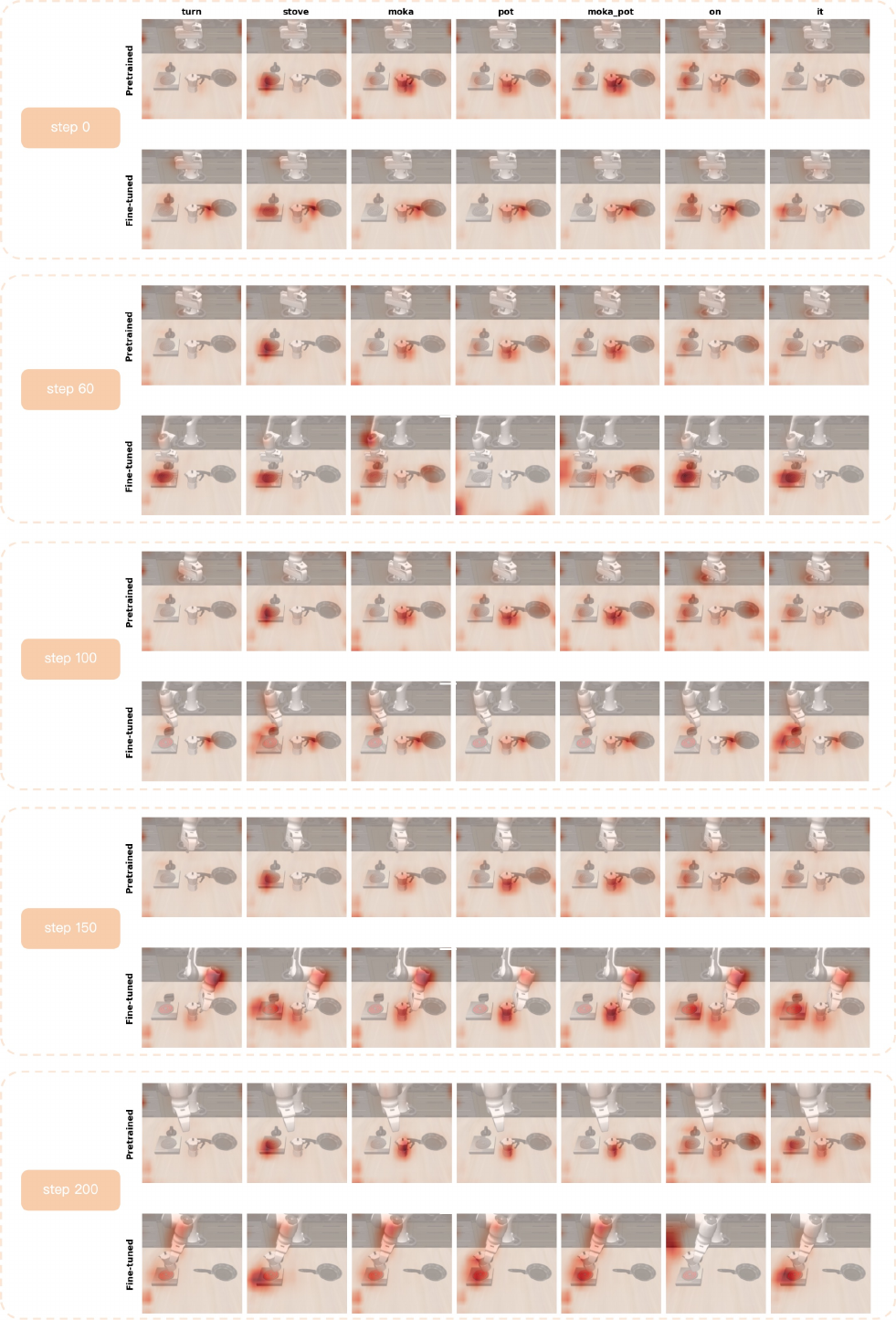}
    \caption{Token-wise text-to-image attention for pretrained and fine-tuned $\pifive$ across execution steps. Columns correspond to instruction tokens, and rows compare checkpoints. Each heatmap shows how the selected text token attends to image patches.}
    \label{fig:pi05_t2i}
\end{figure*}

\begin{figure*}[ht!]
    \centering
    \includegraphics[width=\textwidth]{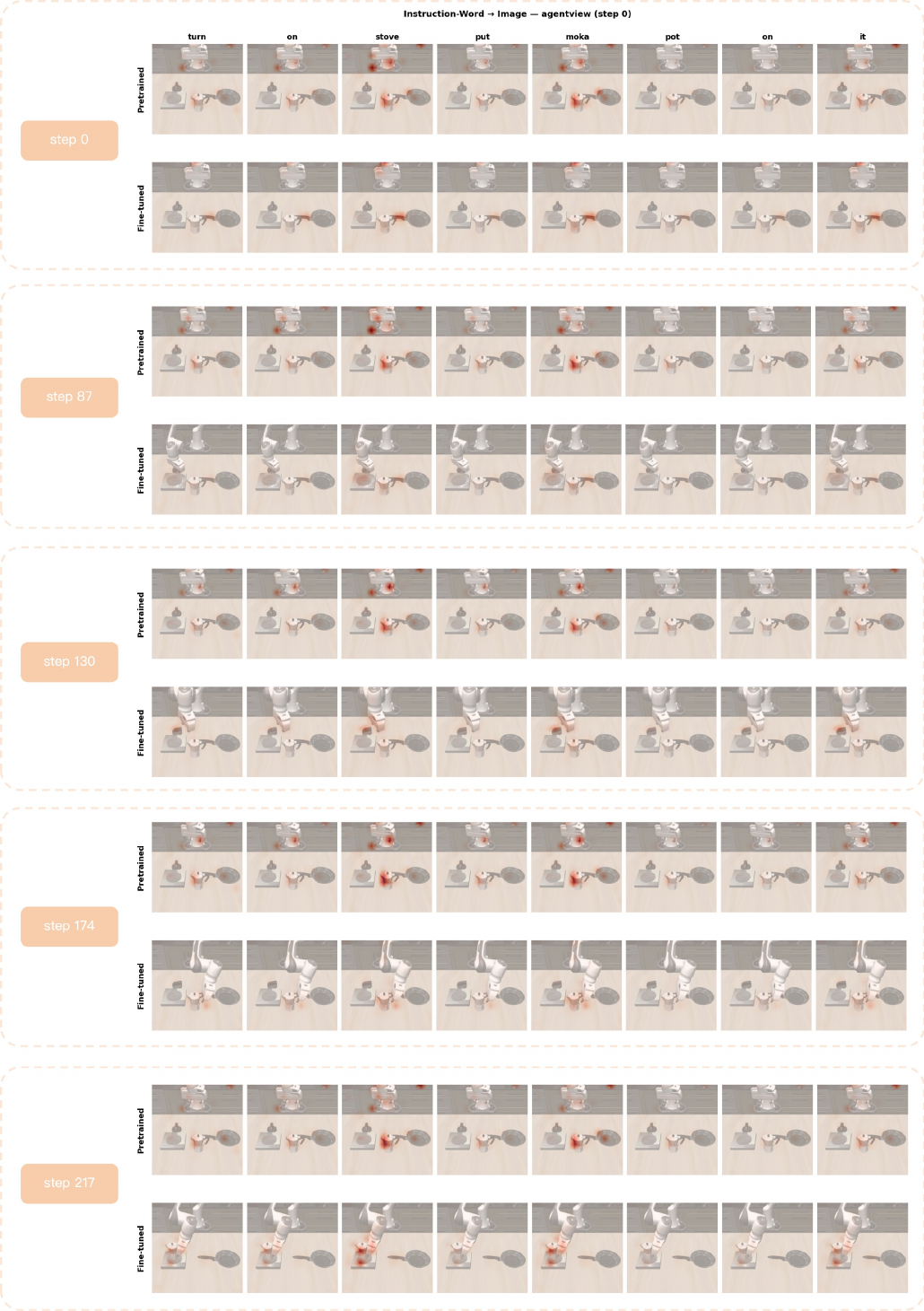}
    \caption{Token-wise text-to-image attention for pretrained and fine-tuned OpenVLA across execution steps. Columns correspond to instruction tokens, and rows compare checkpoints. Each heatmap shows how the selected text token attends to image patches.}
    \label{fig:openvla_t2i}
\end{figure*}

\paragraph{Token-wise text-to-image attention.}
We also visualize text-to-image attention to inspect how individual instruction tokens attend to visual patches. For each displayed instruction token, we project its attention over image patches back to the image grid. Fig.~\ref{fig:pi05_t2i} and~\ref{fig:openvla_t2i} compare pretrained and fine-tuned checkpoints for $\pifive$ and OpenVLA, respectively. 
These plots help separate action-conditioned visual grounding from token-level language-to-image grounding: action-to-image attention measures which visual regions are used for decoding actions, whereas text-to-image attention shows how instruction tokens are visually grounded during representation formation.

\subsection{Visual Perturbation and Editing Protocol}
\label{app:appendix_perturbation_protocol}

We use visual perturbation to test whether the image regions highlighted by attention analyses are also causally required for action execution. For each evaluation episode, we construct region masks over the input observation and replace selected regions before feeding the edited image to the policy. Model weights, task instructions, and the rollout environment are otherwise unchanged. We evaluate four primary region types: target objects, gripper regions, robot body regions, and background. Target-object masks are derived from task-relevant object annotations or simulator-aligned segmentation masks. Gripper masks cover the end-effector region, robot-body masks cover the visible robot morphology, and full-robot masks combine gripper and body regions. Background masks cover image regions outside the annotated task objects and robot regions. When multiple camera views are used, the same perturbation type is applied to the corresponding visible regions in each view.

\paragraph{Mask replacement styles.}
We consider multiple replacement styles because they remove different visual cues. 
In \textit{background-color replacement}, pixels inside the selected mask are replaced by a local or estimated background color. This removes object appearance and weakens boundary cues, making it the most aggressive test of whether the policy depends on the removed region. 
In \textit{black masking}, pixels are replaced by a uniform black value. This removes texture and color but can preserve a coarse silhouette through the artificial mask boundary. 
In \textit{mosaic masking}, the selected region is replaced by a low-resolution or block-wise mosaic pattern. This disrupts fine-grained appearance while preserving partial spatial occupancy and coarse geometry. 
Comparing these styles helps distinguish dependence on visual identity, object geometry, gripper-object spatial relations, and broader scene-layout cues.

\paragraph{Region-specific perturbations.}
Target-object masking tests whether the manipulated object is a causal visual anchor rather than merely an attended region. Gripper masking tests whether the policy depends on end-effector pose and local contact geometry. Robot-body and full-robot masking test whether policies use broader robot configuration cues beyond the gripper. Background masking tests whether policies rely on global scene layout, support surfaces, or contextual correlations outside explicitly task-relevant objects. Because these regions are not semantically equivalent, we interpret performance drops relative to the masked region and replacement style rather than treating all perturbations as generic image corruption.

\paragraph{Additional experimental results.}
The main paper reports averaged perturbation effects across masking strategies. 
In the supplement, we provide task-level success rates for all evaluated model families, including $\pifive$, OpenVLA, OpenVLA-OFT, and X-VLA, to show how visual perturbation sensitivity varies across architectures and task distributions. 
Tab.~\ref{tab:mask1_merged} reports results on LIBERO-10, Tab.~\ref{tab:mask2_merged} reports results on LIBERO-Object, Tab.~\ref{tab:mask3_merged} reports results on LIBERO-Spatial, and Tab.~\ref{tab:mask_goal_merged} reports results on LIBERO-Goal. We additionally report X-VLA perturbation results on the Simpler benchmark in Tab.~\ref{tab:mask4}, providing an out-of-LIBERO check of the same masking protocol. 
Together, these tables provide a task-level view of how target-object, gripper, robot, and background perturbations affect different VLA architectures across the LIBERO benchmark suites.




\begin{table*}[t]
\centering
\small
\setlength{\tabcolsep}{4pt}
\begin{tabular}{llccccc}
\toprule
Model & Setting & LIBERO-10 & Goal & Spatial & Object & Avg. \\
\midrule
\multirow{9}{*}{$\pifive$}
& Baseline & 93.5 (0.0) & 96.0 (0.0) & 98.5 (0.0) & 99.5 (0.0) & 96.9 (0.0) \\
& Gen: drop BOS+newline & 57.5 (-36.0) & 97.5 (+1.5) & 99.0 (+0.5) & 99.5 (0.0) & 88.4 (-8.5) \\
& Gen: drop instruction & 86.0 (-7.5) & 97.0 (+1.0) & 100.0 (+1.5) & 99.5 (0.0) & 95.6 (-1.3) \\
& Gen: keep BOS+newline only & 0.0 (-93.5) & 1.5 (-94.5) & 0.0 (-98.5) & 0.0 (-99.5) & 0.4 (-96.5) \\
& Gen: no image & 0.0 (-93.5) & 4.0 (-92.0) & 0.0 (-98.5) & 0.0 (-99.5) & 1.0 (-95.9) \\
& Gen: no text & 39.0 (-54.5) & 96.5 (+0.5) & 99.0 (+0.5) & 98.0 (-1.5) & 83.1 (-13.8) \\
& Prefill & 0.0 (-93.5) & 11.5 (-84.5) & 77.0 (-21.5) & 71.5 (-28.0) & 40.0 (-56.9) \\
& Comb. + gen no image & 0.0 (-93.5) & 0.0 (-96.0) & 0.0 (-98.5) & 0.0 (-99.5) & 0.0 (-96.9) \\
& Comb. + gen no text & 71.5 (-22.0) & 10.5 (-85.5) & 70.5 (-28.0) & 46.0 (-53.5) & 49.6 (-47.3) \\
\midrule
\multirow{8}{*}{OpenVLA}
& Baseline & 58.0 (0.0) & 74.5 (0.0) & 75.5 (0.0) & 74.0 (0.0) & 70.5 (0.0) \\
& Gen: drop newline & 53.5 (-4.5) & 75.5 (+1.0) & 80.0 (+4.5) & 71.5 (-2.5) & 70.1 (-0.4) \\
& Gen: drop prompt, keep newline & 0.0 (-58.0) & 0.0 (-74.5) & 0.0 (-75.5) & 0.0 (-74.0) & 0.0 (-70.5) \\
& Gen: no image & 1.0 (-57.0) & 16.0 (-58.5) & 44.0 (-31.5) & 32.5 (-41.5) & 23.4 (-47.1) \\
& Gen: no text & 0.0 (-58.0) & 0.0 (-74.5) & 0.0 (-75.5) & 0.0 (-74.0) & 0.0 (-70.5) \\
& Prefill & 0.0 (-58.0) & 0.0 (-74.5) & 0.0 (-75.5) & 0.0 (-74.0) & 0.0 (-70.5) \\
& Comb. + gen no image & 0.0 (-58.0) & 0.0 (-74.5) & 0.0 (-75.5) & 0.0 (-74.0) & 0.0 (-70.5) \\
& Comb. + gen no text & 0.0 (-58.0) & 0.0 (-74.5) & 0.0 (-75.5) & 0.0 (-74.0) & 0.0 (-70.5) \\
\bottomrule
\end{tabular}
\caption{All-layer attention knockout results across LIBERO suites. Each cell reports success rate (\%) and the absolute change relative to the suite-specific all-layer baseline in parentheses. For $\pifive$, \texttt{Prefill} blocks V--L attention; for OpenVLA, \texttt{Prefill} removes image access. \texttt{Comb.} combines the corresponding prefill intervention with the specified generation-stage intervention.}
\label{tab:attention_knockout_all_layers_summary_0526}
\end{table*}

\begin{figure*}[t]
    \includegraphics[width=\linewidth]{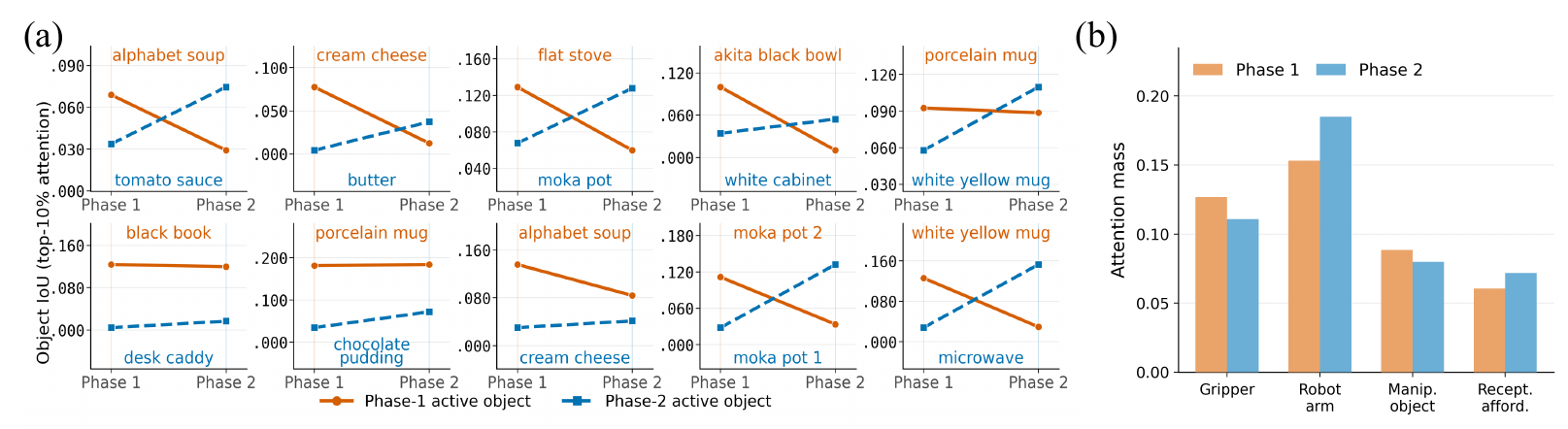}
    \caption{OpenVLA attention IoU and mass on LIBERO-10. (a) Object IoU dynamically shifts between the first (Phase 1) and second (Phase 2) instruction subgoals. (b) Attention mass allocation over robot and object regions. These results indicate that VLA policies successfully generate visually grounded trajectories by tracking task-relevant objects over time.}
    \label{fig:attention_openvla}
\end{figure*}

\begin{table*}[t]
\centering
\scriptsize
\setlength{\tabcolsep}{5pt}
\begin{tabular}{llccccccccccc}
\toprule
Model & Setting & Avg & t0 & t1 & t2 & t3 & t4 & t5 & t6 & t7 & t8 & t9 \\
\midrule

\multirow{15}{*}{$\pifive$}
& Baseline & 75.00 & 56.00 & 98.00 & 64.00 & 90.00 & 76.00 & 86.00 & 70.00 & 96.00 & 58.00 & 56.00 \\
\cmidrule{2-13}
& Target (BG) & 0.00 & 0.00 & 0.00 & 0.00 & 0.00 & 0.00 & 0.00 & 0.00 & 0.00 & 0.00 & 0.00 \\
& Target (Black) & 19.60 & 12.00 & 64.00 & 20.00 & 2.00 & 0.00 & 60.00 & 0.00 & 38.00 & 0.00 & 0.00 \\
& Target (Mosaic) & 47.20 & 82.00 & 96.00 & 34.00 & 8.00 & 22.00 & 94.00 & 22.00 & 58.00 & 54.00 & 2.00 \\
\cmidrule{2-13}
& Gripper (BG) & 8.40 & 6.00 & 0.00 & 0.00 & 4.00 & 18.00 & 54.00 & 0.00 & 2.00 & 0.00 & 0.00 \\
& Gripper (Black) & 52.60 & 60.00 & 96.00 & 6.00 & 60.00 & 46.00 & 68.00 & 52.00 & 78.00 & 24.00 & 36.00 \\
& Gripper (Mosaic) & 53.20 & 62.00 & 90.00 & 2.00 & 56.00 & 72.00 & 84.00 & 82.00 & 60.00 & 14.00 & 10.00 \\
\cmidrule{2-13}
& Robot (BG) & 3.40 & 0.00 & 0.00 & 0.00 & 2.00 & 12.00 & 20.00 & 0.00 & 0.00 & 0.00 & 0.00 \\
& Robot (Black) & 30.20 & 50.00 & 90.00 & 0.00 & 32.00 & 8.00 & 34.00 & 34.00 & 34.00 & 0.00 & 20.00 \\
& Robot (Mosaic) & 39.20 & 54.00 & 64.00 & 0.00 & 32.00 & 64.00 & 38.00 & 72.00 & 66.00 & 2.00 & 0.00 \\
\cmidrule{2-13}
& Robot w/o Gripper (BG) & 52.60 & 70.00 & 96.00 & 0.00 & 36.00 & 66.00 & 44.00 & 44.00 & 94.00 & 44.00 & 32.00 \\
& Robot w/o Gripper (Black) & 58.40 & 80.00 & 94.00 & 0.00 & 84.00 & 58.00 & 58.00 & 52.00 & 90.00 & 26.00 & 42.00 \\
& Robot w/o Gripper (Mosaic) & 64.40 & 66.00 & 98.00 & 4.00 & 92.00 & 72.00 & 62.00 & 78.00 & 86.00 & 58.00 & 28.00 \\
\cmidrule{2-13}
& Background (Black) & 41.00 & 56.00 & 70.00 & 0.00 & 14.00 & 52.00 & 84.00 & 52.00 & 76.00 & 0.00 & 6.00 \\
& Background (Mosaic) & 57.80 & 60.00 & 98.00 & 8.00 & 94.00 & 56.00 & 88.00 & 54.00 & 44.00 & 42.00 & 34.00 \\
\midrule

\multirow{15}{*}{OpenVLA}
& Baseline & 54.33 & 50.00 & 83.33 & 56.67 & 36.67 & 50.00 & 70.00 & 36.67 & 76.67 & 43.33 & 40.00 \\
\cmidrule{2-13}
& Target (BG) & 5.00 & 0.00 & 0.00 & 0.00 & 0.00 & 0.00 & 50.00 & 0.00 & 0.00 & 0.00 & 0.00 \\
& Target (Black) & 20.67 & 33.33 & 43.33 & 10.00 & 0.00 & 26.67 & 73.33 & 3.33 & 16.67 & 0.00 & 0.00 \\
& Target (Mosaic) & 25.00 & 16.67 & 60.00 & 6.67 & 3.33 & 16.67 & 76.67 & 10.00 & 50.00 & 6.67 & 3.33 \\
\cmidrule{2-13}
& Gripper (BG) & 10.67 & 10.00 & 6.67 & 23.33 & 10.00 & 10.00 & 3.33 & 40.00 & 3.33 & 0.00 & 0.00 \\
& Gripper (Black) & 31.33 & 23.33 & 46.67 & 36.67 & 16.67 & 30.00 & 36.67 & 30.00 & 53.33 & 3.33 & 36.67 \\
& Gripper (Mosaic) & 20.33 & 33.33 & 23.33 & 50.00 & 6.67 & 20.00 & 30.00 & 13.33 & 20.00 & 6.67 & 0.00 \\
\cmidrule{2-13}
& Robot (BG) & 0.00 & 0.00 & 0.00 & 0.00 & 0.00 & 0.00 & 0.00 & 0.00 & 0.00 & 0.00 & 0.00 \\
& Robot (Black) & 0.67 & 0.00 & 0.00 & 0.00 & 0.00 & 0.00 & 6.67 & 0.00 & 0.00 & 0.00 & 0.00 \\
& Robot (Mosaic) & 0.33 & 3.33 & 0.00 & 0.00 & 0.00 & 0.00 & 0.00 & 0.00 & 0.00 & 0.00 & 0.00 \\
\cmidrule{2-13}
& Robot w/o Gripper (BG) & 10.00 & 10.00 & 13.33 & 0.00 & 0.00 & 26.67 & 43.33 & 6.67 & 0.00 & 0.00 & 0.00 \\
& Robot w/o Gripper (Black) & 11.67 & 13.33 & 16.67 & 0.00 & 0.00 & 23.33 & 46.67 & 3.33 & 10.00 & 3.33 & 0.00 \\
& Robot w/o Gripper (Mosaic) & 16.00 & 30.00 & 23.33 & 0.00 & 0.00 & 16.67 & 36.67 & 23.33 & 30.00 & 0.00 & 0.00 \\
\cmidrule{2-13}
& Background (Black) & 26.33 & 30.00 & 46.67 & 10.00 & 16.67 & 20.00 & 53.33 & 10.00 & 36.67 & 6.67 & 33.33 \\
& Background (Mosaic) & 48.00 & 46.67 & 73.33 & 30.00 & 36.67 & 56.67 & 76.67 & 56.67 & 53.33 & 20.00 & 30.00 \\
\midrule

\multirow{15}{*}{OpenVLA-OFT}
& Baseline & 94.80 & 98.00 & 96.00 & 100.00 & 94.00 & 98.00 & 100.00 & 82.00 & 100.00 & 88.00 & 92.00 \\
\cmidrule{2-13}
& Target (BG) & 10.40 & 0.00 & 0.00 & 0.00 & 0.00 & 0.00 & 100.00 & 0.00 & 4.00 & 0.00 & 0.00 \\
& Target (Black) & 15.20 & 10.00 & 16.00 & 2.00 & 0.00 & 0.00 & 100.00 & 0.00 & 24.00 & 0.00 & 0.00 \\
& Target (Mosaic) & 48.20 & 86.00 & 98.00 & 0.00 & 78.00 & 14.00 & 100.00 & 4.00 & 96.00 & 0.00 & 6.00 \\
\cmidrule{2-13}
& Gripper (BG) & 89.80 & 92.00 & 100.00 & 92.00 & 94.00 & 100.00 & 98.00 & 86.00 & 86.00 & 64.00 & 86.00 \\
& Gripper (Black) & 89.80 & 92.00 & 98.00 & 98.00 & 90.00 & 96.00 & 94.00 & 84.00 & 98.00 & 60.00 & 88.00 \\
& Gripper (Mosaic) & 95.00 & 98.00 & 94.00 & 92.00 & 98.00 & 98.00 & 96.00 & 92.00 & 98.00 & 88.00 & 96.00 \\
\cmidrule{2-13}
& Robot (BG) & 66.80 & 68.00 & 66.00 & 54.00 & 82.00 & 88.00 & 80.00 & 50.00 & 88.00 & 30.00 & 62.00 \\
& Robot (Black) & 79.00 & 86.00 & 92.00 & 66.00 & 86.00 & 96.00 & 84.00 & 52.00 & 94.00 & 52.00 & 82.00 \\
& Robot (Mosaic) & 84.40 & 90.00 & 96.00 & 70.00 & 100.00 & 92.00 & 88.00 & 92.00 & 98.00 & 30.00 & 88.00 \\
\cmidrule{2-13}
& Robot w/o Gripper (BG) & 85.40 & 88.00 & 88.00 & 94.00 & 90.00 & 96.00 & 98.00 & 54.00 & 100.00 & 58.00 & 88.00 \\
& Robot w/o Gripper (Black) & 84.00 & 92.00 & 94.00 & 96.00 & 84.00 & 98.00 & 96.00 & 46.00 & 98.00 & 56.00 & 80.00 \\
& Robot w/o Gripper (Mosaic) & 91.20 & 92.00 & 96.00 & 94.00 & 100.00 & 98.00 & 98.00 & 94.00 & 98.00 & 50.00 & 92.00 \\
\cmidrule{2-13}
& Background (Black) & 73.80 & 100.00 & 92.00 & 86.00 & 90.00 & 92.00 & 84.00 & 62.00 & 96.00 & 8.00 & 28.00 \\
& Background (Mosaic) & 84.60 & 92.00 & 98.00 & 80.00 & 92.00 & 84.00 & 100.00 & 76.00 & 100.00 & 40.00 & 84.00 \\
\midrule

\multirow{15}{*}{X-VLA}
& Baseline & 96.00 & 92.00 & 98.00 & 100.00 & 92.00 & 96.00 & 96.00 & 94.00 & 98.00 & 96.00 & 98.00 \\
\cmidrule{2-13}
& Target (BG) & 46.40 & 70.00 & 46.00 & 86.00 & 26.00 & 22.00 & 80.00 & 38.00 & 28.00 & 40.00 & 28.00 \\
& Target (Black) & 78.60 & 82.00 & 98.00 & 92.00 & 72.00 & 26.00 & 94.00 & 72.00 & 86.00 & 100.00 & 64.00 \\
& Target (Mosaic) & 91.40 & 94.00 & 100.00 & 92.00 & 96.00 & 76.00 & 84.00 & 94.00 & 90.00 & 90.00 & 98.00 \\
\cmidrule{2-13}
& Gripper (BG) & 90.80 & 80.00 & 90.00 & 96.00 & 88.00 & 96.00 & 92.00 & 86.00 & 96.00 & 94.00 & 90.00 \\
& Gripper (Black) & 96.60 & 98.00 & 100.00 & 100.00 & 92.00 & 92.00 & 92.00 & 98.00 & 96.00 & 100.00 & 98.00 \\
& Gripper (Mosaic) & 95.60 & 94.00 & 98.00 & 100.00 & 88.00 & 98.00 & 90.00 & 94.00 & 96.00 & 98.00 & 100.00 \\
\cmidrule{2-13}
& Robot (BG) & 43.33 & 60.00 & 66.67 & 23.33 & 16.67 & 36.67 & 53.33 & 46.67 & 76.67 & 6.67 & 46.67 \\
& Robot (Black) & 83.40 & 78.00 & 100.00 & 80.00 & 80.00 & 92.00 & 84.00 & 96.00 & 90.00 & 68.00 & 66.00 \\
& Robot (Mosaic) & 96.20 & 96.00 & 100.00 & 100.00 & 96.00 & 100.00 & 86.00 & 92.00 & 96.00 & 100.00 & 96.00 \\
\cmidrule{2-13}
& Robot w/o Gripper (BG) & 80.40 & 90.00 & 100.00 & 92.00 & 82.00 & 100.00 & 84.00 & 92.00 & 98.00 & 14.00 & 52.00 \\
& Robot w/o Gripper (Black) & 88.20 & 92.00 & 100.00 & 92.00 & 80.00 & 96.00 & 84.00 & 94.00 & 100.00 & 86.00 & 58.00 \\
& Robot w/o Gripper (Mosaic) & 96.80 & 98.00 & 100.00 & 100.00 & 100.00 & 98.00 & 88.00 & 92.00 & 100.00 & 94.00 & 98.00 \\
\cmidrule{2-13}
& Background (Black) & 89.00 & 94.00 & 92.00 & 96.00 & 84.00 & 96.00 & 36.00 & 94.00 & 100.00 & 100.00 & 98.00 \\
& Background (Mosaic) & 89.40 & 94.00 & 98.00 & 98.00 & 92.00 & 94.00 & 34.00 & 92.00 & 98.00 & 96.00 & 98.00 \\

\bottomrule
\end{tabular}
\caption{Success rates (\%) across tasks and masking strategies on \textbf{LIBERO-10} for $\pifive$, OpenVLA, OpenVLA-OFT and X-VLA. The instruction is provided in Tab.~\ref{tab:instruction_all}.} \label{tab:mask1_merged}
\end{table*}

\begin{table*}[t]
\centering
\scriptsize
\setlength{\tabcolsep}{4.5pt}
\begin{tabular}{llccccccccccc}
\toprule
Model & Setting & Avg & t0 & t1 & t2 & t3 & t4 & t5 & t6 & t7 & t8 & t9 \\
\midrule

\multirow{15}{*}{$\pifive$}
& Baseline & 95.60 & 96.00 & 98.00 & 78.00 & 98.00 & 90.00 & 100.00 & 100.00 & 96.00 & 100.00 & 100.00  \\
\cmidrule{2-13}
& Target (BG) & 13.00 & 8.00 & 0.00 & 0.00 & 0.00 & 0.00 & 0.00 & 52.00 & 0.00 & 70.00 & 0.00 \\
& Target (Black) & 61.60 & 16.00 & 94.00 & 44.00 & 94.00 & 94.00 & 24.00 & 72.00 & 90.00 & 88.00 & 0.00 \\
& Target (Mosaic) & 74.00 & 58.00 & 94.00 & 22.00 & 86.00 & 96.00 & 14.00 & 90.00 & 90.00 & 92.00 & 98.00 \\
\cmidrule{2-13}
& Gripper (BG) & 3.20 & 2.00 & 2.00 & 0.00 & 0.00 & 24.00 & 0.00 & 0.00 & 0.00 & 4.00 & 0.00 \\
& Gripper (Black) & 97.40 & 100.00 & 100.00 & 86.00 & 96.00 & 100.00 & 98.00 & 98.00 & 96.00 & 100.00 & 100.00 \\
& Gripper (Mosaic) & 79.40 & 100.00 & 96.00 & 24.00 & 76.00 & 100.00 & 26.00 & 100.00 & 100.00 & 98.00 & 74.00 \\
\cmidrule{2-13}
& Robot (BG) & 26.40 & 8.00 & 52.00 & 8.00 & 24.00 & 64.00 & 0.00 & 10.00 & 58.00 & 40.00 & 0.00 \\
& Robot (Black)& 93.60 & 94.00 & 98.00 & 90.00 & 84.00 & 92.00 & 98.00 & 90.00 & 100.00 & 96.00 & 94.00 \\
& Robot (Mosaic) & 74.60 & 100.00 & 96.00 & 26.00 & 76.00 & 100.00 & 20.00 & 100.00 & 100.00 & 94.00 & 34.00 \\
\cmidrule{2-13}
& Robot w/o Gripper (BG) & 91.80 & 94.00 & 98.00 & 90.00 & 76.00 & 66.00 & 98.00 & 100.00 & 96.00 & 100.00 & 100.00 \\
& Robot w/o Gripper (Black) & 92.80 & 94.00 & 100.00 & 90.00 & 88.00 & 66.00 & 98.00 & 96.00 & 100.00 & 100.00 & 96.00 \\
& Robot w/o Gripper (Mosaic) & 96.60 & 100.00 & 98.00 & 80.00 & 98.00 & 96.00 & 100.00 & 98.00 & 98.00 & 100.00 & 98.00 \\
\cmidrule{2-13}
& Background (Black) & 79.40 & 66.00 & 100.00 & 80.00 & 78.00 & 100.00 & 68.00 & 62.00 & 68.00 & 88.00 & 84.00 \\
& Background (Mosaic) & 86.60 & 96.00 & 96.00 & 22.00 & 78.00 & 100.00 & 88.00 & 92.00 & 98.00 & 100.00 & 96.00 \\
\midrule

\multirow{15}{*}{OpenVLA}
& Baseline & 70.00 & 63.33 & 76.67 & 70.00 & 46.67 & 83.33 & 73.33 & 63.33 & 76.67 & 60.00 & 86.67 \\
\cmidrule{2-13}
& Target (BG) & 0.33 & 0.00 & 0.00 & 0.00 & 0.00 & 0.00 & 0.00 & 3.33 & 0.00 & 0.00 & 0.00 \\
& Target (Black) & 55.67 & 83.33 & 46.67 & 73.33 & 43.33 & 63.33 & 80.00 & 50.00 & 73.33 & 20.00 & 23.33 \\
& Target (Mosaic) & 59.67 & 70.00 & 50.00 & 56.67 & 26.67 & 76.67 & 63.33 & 63.33 & 86.67 & 60.00 & 43.33 \\
\cmidrule{2-13}
& Gripper (BG) & 1.00 & 0.00 & 0.00 & 6.67 & 3.33 & 0.00 & 0.00 & 0.00 & 0.00 & 0.00 & 0.00 \\
& Gripper (Black) & 72.00 & 73.33 & 40.00 & 80.00 & 56.67 & 80.00 & 80.00 & 86.67 & 96.67 & 43.33 & 83.33 \\
& Gripper (Mosaic) & 3.67 & 0.00 & 0.00 & 0.00 & 0.00 & 0.00 & 0.00 & 0.00 & 0.00 & 0.00 & 36.67 \\
\cmidrule{2-13}
& Robot (BG) & 0.00 & 0.00 & 0.00 & 0.00 & 0.00 & 0.00 & 0.00 & 0.00 & 0.00 & 0.00 & 0.00 \\
& Robot (Black) & 4.00 & 0.00 & 0.00 & 0.00 & 0.00 & 0.00 & 36.67 & 0.00 & 0.00 & 0.00 & 3.33 \\
& Robot (Mosaic) & 0.00 & 0.00 & 0.00 & 0.00 & 0.00 & 0.00 & 0.00 & 0.00 & 0.00 & 0.00 & 0.00 \\
\cmidrule{2-13}
& Robot w/o Gripper (BG) & 34.33 & 73.33 & 33.33 & 43.33 & 56.67 & 33.33 & 6.67 & 0.00 & 6.67 & 10.00 & 80.00 \\
& Robot w/o Gripper (Black) & 17.33 & 0.00 & 16.67 & 36.67 & 26.67 & 0.00 & 23.33 & 0.00 & 0.00 & 0.00 & 70.00 \\
& Robot w/o Gripper (Mosaic) & 15.67 & 50.00 & 13.33 & 10.00 & 30.00 & 3.33 & 6.67 & 0.00 & 6.67 & 0.00 & 36.67 \\
\cmidrule{2-13}
& Background (Black) & 42.67 & 63.33 & 23.33 & 43.33 & 40.00 & 66.67 & 0.00 & 6.67 & 73.33 & 50.00 & 60.00 \\
& Background (Mosaic) & 65.67 & 76.67 & 56.67 & 83.33 & 53.33 & 93.33 & 63.33 & 40.00 & 76.67 & 40.00 & 73.33 \\
\midrule

\multirow{15}{*}{OpenVLA-OFT}
& Baseline & 99.80 & 100.00 & 100.00 & 98.00 & 100.00 & 100.00 & 100.00 & 100.00 & 100.00 & 100.00 & 100.00 \\
\cmidrule{2-13}
& Target (BG) & 57.00 & 94.00 & 0.00 & 16.00 & 46.00 & 38.00 & 80.00 & 94.00 & 12.00 & 96.00 & 94.00 \\
& Target (Black) & 84.80 & 70.00 & 32.00 & 100.00 & 96.00 & 100.00 & 100.00 & 98.00 & 80.00 & 74.00 & 98.00 \\
& Target (Mosaic) & 98.40 & 94.00 & 96.00 & 98.00 & 98.00 & 100.00 & 98.00 & 100.00 & 100.00 & 100.00 & 100.00 \\
\cmidrule{2-13}
& Gripper (BG) & 94.20 & 100.00 & 100.00 & 100.00 & 94.00 & 66.00 & 100.00 & 96.00 & 88.00 & 100.00 & 98.00 \\
& Gripper (Black) & 98.80 & 96.00 & 100.00 & 100.00 & 98.00 & 100.00 & 100.00 & 96.00 & 100.00 & 98.00 & 100.00 \\
& Gripper (Mosaic) & 100.00 & 100.00 & 100.00 & 100.00 & 100.00 & 100.00 & 100.00 & 100.00 & 100.00 & 100.00 & 100.00 \\
\cmidrule{2-13}
& Robot (BG) & 96.00 & 98.00 & 100.00 & 100.00 & 98.00 & 92.00 & 100.00 & 92.00 & 86.00 & 96.00 & 98.00 \\
& Robot (Black) & 96.80 & 98.00 & 98.00 & 96.00 & 98.00 & 96.00 & 98.00 & 92.00 & 96.00 & 100.00 & 96.00 \\
& Robot (Mosaic) & 97.60 & 100.00 & 92.00 & 100.00 & 98.00 & 96.00 & 100.00 & 96.00 & 96.00 & 98.00 & 100.00 \\
\cmidrule{2-13}
& Robot w/o Gripper (BG) & 98.60 & 100.00 & 100.00 & 98.00 & 100.00 & 100.00 & 98.00 & 98.00 & 100.00 & 100.00 & 92.00 \\
& Robot w/o Gripper (Black) & 99.80 & 100.00 & 100.00 & 100.00 & 100.00 & 100.00 & 100.00 & 98.00 & 100.00 & 100.00 & 100.00 \\
& Robot w/o Gripper (Mosaic) & 98.80 & 100.00 & 98.00 & 100.00 & 96.00 & 100.00 & 98.00 & 98.00 & 100.00 & 100.00 & 98.00 \\
\cmidrule{2-13}
& Background (Black) & 47.40 & 96.00 & 2.00 & 2.00 & 16.00 & 86.00 & 52.00 & 70.00 & 24.00 & 100.00 & 26.00 \\
& Background (Mosaic) & 95.00 & 100.00 & 86.00 & 98.00 & 100.00 & 96.00 & 88.00 & 94.00 & 98.00 & 100.00 & 90.00 \\

\midrule

\multirow{15}{*}{X-VLA}
& Baseline & 98.60 & 100.00 & 100.00 & 100.00 & 96.00 & 98.00 & 94.00 & 100.00 & 98.00 & 100.00 & 100.00 \\
\cmidrule{2-13}
& Target (BG) & 94.20 & 98.00 & 100.00 & 64.00 & 100.00 & 100.00 & 88.00 & 100.00 & 98.00 & 98.00 & 96.00 \\
& Target (Black) & 97.40 & 100.00 & 100.00 & 100.00 & 98.00 & 100.00 & 76.00 & 100.00 & 100.00 & 100.00 & 100.00 \\
& Target (Mosaic) & 98.40 & 100.00 & 100.00 & 100.00 & 94.00 & 100.00 & 90.00 & 100.00 & 100.00 & 100.00 & 100.00 \\
\cmidrule{2-13}
& Gripper (BG) & 95.60 & 98.00 & 100.00 & 100.00 & 98.00 & 100.00 & 62.00 & 100.00 & 98.00 & 100.00 & 100.00 \\
& Gripper (Black) & 97.80 & 100.00 & 100.00 & 100.00 & 96.00 & 100.00 & 84.00 & 100.00 & 98.00 & 100.00 & 100.00 \\
& Gripper (Mosaic) & 97.20 & 100.00 & 100.00 & 100.00 & 100.00 & 100.00 & 74.00 & 100.00 & 98.00 & 100.00 & 100.00 \\
\cmidrule{2-13}
& Robot (BG) & 63.20 & 96.00 & 42.00 & 96.00 & 82.00 & 62.00 & 30.00 & 42.00 & 78.00 & 22.00 & 82.00 \\
& Robot (Black) & 97.80 & 100.00 & 100.00 & 100.00 & 96.00 & 100.00 & 82.00 & 100.00 & 100.00 & 100.00 & 100.00 \\
& Robot (Mosaic) & 97.40 & 100.00 & 100.00 & 100.00 & 100.00 & 98.00 & 78.00 & 100.00 & 98.00 & 100.00 & 100.00 \\
\cmidrule{2-13}
& Robot w/o Gripper (BG) & 93.20 & 100.00 & 92.00 & 98.00 & 92.00 & 100.00 & 90.00 & 80.00 & 98.00 & 88.00 & 94.00 \\
& Robot w/o Gripper (Black) & 98.40 & 100.00 & 100.00 & 98.00 & 94.00 & 100.00 & 96.00 & 100.00 & 98.00 & 100.00 & 98.00 \\
& Robot w/o Gripper (Mosaic) & 98.40 & 100.00 & 100.00 & 100.00 & 96.00 & 100.00 & 90.00 & 100.00 & 98.00 & 100.00 & 100.00 \\
\cmidrule{2-13}
& Background (Black) & 91.00 & 100.00 & 82.00 & 40.00 & 92.00 & 100.00 & 98.00 & 100.00 & 98.00 & 100.00 & 100.00 \\
& Background (Mosaic) & 98.40 & 98.00 & 100.00 & 98.00 & 94.00 & 100.00 & 96.00 & 100.00 & 98.00 & 100.00 & 100.00 \\

\bottomrule
\end{tabular}
\caption{Success rates (\%) across tasks and masking strategies on \textbf{LIBERO-Object} for $\pifive$, OpenVLA, OpenVLA-OFT, and X-VLA. The instruction is provided in Tab.~\ref{tab:instruction_all}.} \label{tab:mask2_merged}
\end{table*}

\begin{table*}[t]
\centering
\scriptsize
\setlength{\tabcolsep}{5pt}
\begin{tabular}{llccccccccccc}
\toprule
Model & Setting & Avg & t0 & t1 & t2 & t3 & t4 & t5 & t6 & t7 & t8 & t9 \\
\midrule

\multirow{15}{*}{$\pifive$}
& Baseline & 95.80 & 100.00 & 92.00 & 100.00 & 100.00 & 88.00 & 100.00 & 100.00 & 96.00 & 86.00 & 96.00 \\
\cmidrule{2-13}
& Target (BG) & 7.80 & 0.00 & 0.00 & 0.00 & 0.00 & 8.00 & 0.00 & 0.00 & 52.00 & 0.00 & 18.00 \\
& Target (Black) & 23.00 & 44.00 & 12.00 & 64.00 & 6.00 & 0.00 & 2.00 & 24.00 & 64.00 & 0.00 & 14.00 \\
& Target (Mosaic) & 50.20 & 74.00 & 64.00 & 46.00 & 62.00 & 20.00 & 16.00 & 78.00 & 76.00 & 20.00 & 46.00 \\
\cmidrule{2-13}
& Gripper (BG) & 64.00 & 76.00 & 66.00 & 100.00 & 64.00 & 20.00 & 72.00 & 84.00 & 98.00 & 60.00 & 0.00 \\
& Gripper (Black) & 94.80 & 100.00 & 94.00 & 100.00 & 96.00 & 78.00 & 98.00 & 100.00 & 100.00 & 90.00 & 92.00 \\
& Gripper (Mosaic) & 89.80 & 100.00 & 94.00 & 100.00 & 92.00 & 58.00 & 94.00 & 100.00 & 98.00 & 90.00 & 72.00 \\
\cmidrule{2-13}
& Robot (BG) & 47.60 & 78.00 & 2.00 & 92.00 & 44.00 & 4.00 & 68.00 & 94.00 & 86.00 & 8.00 & 0.00 \\
& Robot (Black)& 95.00 & 100.00 & 98.00 & 98.00 & 94.00 & 84.00 & 98.00 & 100.00 & 96.00 & 94.00 & 88.00 \\
& Robot (Mosaic) & 89.60 & 100.00 & 98.00 & 100.00 & 82.00 & 66.00 & 94.00 & 98.00 & 100.00 & 88.00 & 70.00 \\
\cmidrule{2-13}
& Robot w/o Gripper (BG) & 97.40 & 98.00 & 100.00 & 98.00 & 100.00 & 92.00 & 96.00 & 100.00 & 100.00 & 92.00 & 98.00 \\
& Robot w/o Gripper (Black) & 99.00 & 98.00 & 98.00 & 100.00 & 100.00 & 98.00 & 100.00 & 100.00 & 100.00 & 96.00 & 100.00 \\
& Robot w/o Gripper (Mosaic) & 96.40 & 100.00 & 96.00 & 96.00 & 100.00 & 92.00 & 96.00 & 100.00 & 94.00 & 92.00 & 98.00 \\
\cmidrule{2-13}
& Background (Black) & 67.20 & 96.00 & 46.00 & 94.00 & 58.00 & 54.00 & 30.00 & 100.00 & 92.00 & 26.00 & 76.00 \\
& Background (Mosaic) & 93.00 & 100.00 & 96.00 & 100.00 & 96.00 & 68.00 & 100.00 & 98.00 & 94.00 & 84.00 & 94.00 \\
\midrule

\multirow{15}{*}{OpenVLA}
& Baseline & 79.67 & 86.67 & 90.00 & 80.00 & 100.00 & 76.67 & 46.67 & 86.67 & 76.67 & 83.33 & 70.00 \\
\cmidrule{2-13}
& Target (BG) & 36.67 & 26.67 & 83.33 & 43.33 & 36.67 & 3.33 & 0.00 & 66.67 & 70.00 & 16.67 & 20.00 \\
& Target (Black) & 63.33 & 60.00 & 93.33 & 63.33 & 86.67 & 63.33 & 3.33 & 80.00 & 90.00 & 50.00 & 43.33 \\
& Target (Mosaic) & 70.00 & 83.33 & 93.33 & 76.67 & 83.33 & 60.00 & 30.00 & 86.67 & 80.00 & 63.33 & 43.33 \\
\cmidrule{2-13}
& Gripper (BG) & 19.00 & 23.33 & 0.00 & 10.00 & 26.67 & 13.33 & 0.00 & 23.33 & 40.00 & 53.33 & 0.00 \\
& Gripper (Black) & 70.67 & 80.00 & 70.00 & 93.33 & 86.67 & 56.67 & 13.33 & 90.00 & 80.00 & 73.33 & 63.33 \\
& Gripper (Mosaic) & 53.00 & 56.67 & 73.33 & 46.67 & 90.00 & 6.67 & 0.00 & 73.33 & 60.00 & 76.67 & 46.67 \\
\cmidrule{2-13}
& Robot (BG) & 0.00 & 0.00 & 0.00 & 0.00 & 0.00 & 0.00 & 0.00 & 0.00 & 0.00 & 0.00 & 0.00 \\
& Robot (Black) & 3.33 & 0.00 & 13.33 & 0.00 & 0.00 & 3.33 & 0.00 & 0.00 & 0.00 & 16.67 & 0.00 \\
& Robot (Mosaic) & 2.00 & 0.00 & 0.00 & 0.00 & 0.00 & 0.00 & 0.00 & 0.00 & 0.00 & 16.67 & 3.33 \\
\cmidrule{2-13}
& Robot w/o Gripper (BG) & 22.00 & 13.33 & 16.67 & 46.67 & 43.33 & 26.67 & 0.00 & 10.00 & 0.00 & 23.33 & 40.00 \\
& Robot w/o Gripper (Black) & 31.67 & 46.67 & 46.67 & 33.33 & 70.00 & 46.67 & 0.00 & 33.33 & 6.67 & 23.33 & 10.00 \\
& Robot w/o Gripper (Mosaic) & 49.67 & 43.33 & 76.67 & 73.33 & 60.00 & 66.67 & 0.00 & 33.33 & 43.33 & 63.33 & 36.67 \\
\cmidrule{2-13}
& Background (Black) & 47.33 & 43.33 & 3.33 & 46.67 & 100.00 & 56.67 & 3.33 & 83.33 & 23.33 & 70.00 & 43.33 \\
& Background (Mosaic) & 82.33 & 96.67 & 96.67 & 80.00 & 86.67 & 66.67 & 46.67 & 93.33 & 90.00 & 90.00 & 76.67 \\
\midrule

\multirow{15}{*}{OpenVLA-OFT}
& Baseline & 92.80 & 100.00 & 98.00 & 100.00 & 100.00 & 96.00 & 46.00 & 100.00 & 94.00 & 96.00 & 98.00 \\
\cmidrule{2-13}
& Target (BG) & 26.80 & 26.00 & 0.00 & 10.00 & 26.00 & 56.00 & 0.00 & 30.00 & 18.00 & 10.00 & 92.00 \\
& Target (Black) & 35.40 & 14.00 & 54.00 & 0.00 & 58.00 & 42.00 & 2.00 & 44.00 & 18.00 & 32.00 & 90.00 \\
& Target (Mosaic) & 76.00 & 92.00 & 90.00 & 92.00 & 84.00 & 90.00 & 6.00 & 96.00 & 38.00 & 76.00 & 96.00 \\
\cmidrule{2-13}
& Gripper (BG) & 92.40 & 100.00 & 100.00 & 100.00 & 100.00 & 100.00 & 36.00 & 98.00 & 96.00 & 96.00 & 98.00 \\
& Gripper (Black) & 92.80 & 100.00 & 100.00 & 98.00 & 100.00 & 100.00 & 46.00 & 98.00 & 96.00 & 94.00 & 96.00 \\
& Gripper (Mosaic) & 93.80 & 100.00 & 98.00 & 100.00 & 100.00 & 98.00 & 54.00 & 100.00 & 92.00 & 96.00 & 100.00 \\
\cmidrule{2-13}
& Robot (BG) & 81.40 & 86.00 & 84.00 & 100.00 & 94.00 & 88.00 & 30.00 & 82.00 & 90.00 & 66.00 & 94.00 \\
& Robot (Black) & 86.80 & 88.00 & 94.00 & 98.00 & 98.00 & 98.00 & 30.00 & 100.00 & 94.00 & 74.00 & 94.00 \\
& Robot (Mosaic) & 91.60 & 98.00 & 100.00 & 100.00 & 100.00 & 90.00 & 38.00 & 100.00 & 94.00 & 96.00 & 100.00 \\
\cmidrule{2-13}
& Robot w/o Gripper (BG) & 90.40 & 90.00 & 100.00 & 100.00 & 96.00 & 94.00 & 48.00 & 100.00 & 92.00 & 90.00 & 94.00 \\
& Robot w/o Gripper (Black) & 91.00 & 96.00 & 100.00 & 100.00 & 100.00 & 98.00 & 40.00 & 100.00 & 92.00 & 90.00 & 94.00 \\
& Robot w/o Gripper (Mosaic) & 91.20 & 98.00 & 100.00 & 96.00 & 100.00 & 98.00 & 34.00 & 100.00 & 96.00 & 94.00 & 96.00 \\
\cmidrule{2-13}
& Background (Black) & 83.20 & 96.00 & 88.00 & 100.00 & 100.00 & 82.00 & 0.00 & 100.00 & 94.00 & 86.00 & 86.00 \\
& Background (Mosaic) & 88.20 & 100.00 & 86.00 & 100.00 & 98.00 & 92.00 & 24.00 & 100.00 & 94.00 & 94.00 & 94.00 \\

\midrule

\multirow{15}{*}{X-VLA}
& Baseline & 98.00 & 98.00 & 100.00 & 100.00 & 100.00 & 90.00 & 98.00 & 100.00 & 94.00 & 100.00 & 100.00 \\
\cmidrule{2-13}
& Target (BG) & 79.80 & 66.00 & 84.00 & 64.00 & 92.00 & 84.00 & 86.00 & 90.00 & 58.00 & 82.00 & 92.00 \\
& Target (Black) & 90.20 & 56.00 & 94.00 & 100.00 & 100.00 & 90.00 & 98.00 & 98.00 & 78.00 & 94.00 & 94.00 \\
& Target (Mosaic) & 89.80 & 92.00 & 90.00 & 74.00 & 100.00 & 96.00 & 88.00 & 100.00 & 74.00 & 94.00 & 90.00 \\
\cmidrule{2-13}
& Gripper (BG) & 97.00 & 92.00 & 100.00 & 98.00 & 100.00 & 94.00 & 98.00 & 100.00 & 94.00 & 100.00 & 94.00 \\
& Gripper (Black) & 97.20 & 100.00 & 100.00 & 100.00 & 98.00 & 90.00 & 96.00 & 100.00 & 94.00 & 98.00 & 96.00 \\
& Gripper (Mosaic) & 96.80 & 100.00 & 100.00 & 100.00 & 100.00 & 90.00 & 96.00 & 98.00 & 92.00 & 98.00 & 94.00 \\
\cmidrule{2-13}
& Robot (BG) & 80.80 & 86.00 & 88.00 & 94.00 & 98.00 & 88.00 & 12.00 & 80.00 & 72.00 & 100.00 & 90.00 \\
& Robot (Black) & 96.40 & 100.00 & 100.00 & 100.00 & 100.00 & 96.00 & 92.00 & 98.00 & 88.00 & 98.00 & 92.00 \\
& Robot (Mosaic) & 96.20 & 100.00 & 98.00 & 100.00 & 100.00 & 88.00 & 96.00 & 98.00 & 96.00 & 94.00 & 92.00 \\
\cmidrule{2-13}
& Robot w/o Gripper (BG) & 98.00 & 100.00 & 100.00 & 100.00 & 100.00 & 94.00 & 100.00 & 92.00 & 96.00 & 98.00 & 100.00 \\
& Robot w/o Gripper (Black) & 97.60 & 100.00 & 100.00 & 100.00 & 100.00 & 90.00 & 96.00 & 100.00 & 90.00 & 100.00 & 100.00 \\
& Robot w/o Gripper (Mosaic) & 97.20 & 100.00 & 100.00 & 100.00 & 98.00 & 88.00 & 100.00 & 100.00 & 92.00 & 98.00 & 96.00 \\
\cmidrule{2-13}
& Background (Black) & 97.60 & 100.00 & 98.00 & 100.00 & 98.00 & 94.00 & 98.00 & 100.00 & 92.00 & 100.00 & 96.00 \\
& Background (Mosaic) & 97.60 & 100.00 & 100.00 & 100.00 & 100.00 & 84.00 & 98.00 & 100.00 & 94.00 & 100.00 & 100.00 \\

\bottomrule
\end{tabular}
\caption{Success rates (\%) across tasks and masking strategies on \textbf{LIBERO-Spatial} for $\pifive$, OpenVLA, OpenVLA-OFT, and X-VLA. The instruction is provided in Tab.~\ref{tab:instruction_all}.} \label{tab:mask3_merged}
\end{table*}

\begin{table*}[htbp]
\centering
\scriptsize
\setlength{\tabcolsep}{5pt}
\begin{tabular}{llccccccccccc}
\toprule
Model & Setting & Avg & t0 & t1 & t2 & t3 & t4 & t5 & t6 & t7 & t8 & t9 \\
\midrule

\multirow{15}{*}{$\pifive$}
& Baseline & 80.00 & 8.00 & 100.00 & 94.00 & 46.00 & 100.00 & 100.00 & 100.00 & 100.00 & 100.00 & 52.00 \\
\cmidrule{2-13}
& Target(BG) & 19.80 & 6.00 & 12.00 & 52.00 & 2.00 & 10.00 & 6.00 & 14.00 & 68.00 & 20.00 & 8.00 \\
& Target(Black) & 58.80 & 4.00 & 70.00 & 92.00 & 2.00 & 76.00 & 54.00 & 54.00 & 100.00 & 56.00 & 80.00 \\
& Target(Mosaic) & 58.40 & 6.00 & 32.00 & 92.00 & 8.00 & 12.00 & 98.00 & 92.00 & 100.00 & 94.00 & 50.00 \\
\cmidrule{2-13}
& Gripper(BG) & 59.20 & 0.00 & 96.00 & 68.00 & 6.00 & 96.00 & 44.00 & 90.00 & 100.00 & 88.00 & 4.00 \\
& Gripper(Black) & 74.80 & 0.00 & 100.00 & 90.00 & 28.00 & 100.00 & 100.00 & 100.00 & 100.00 & 100.00 & 30.00 \\
& Gripper(Mosaic) & 74.20 & 2.00 & 100.00 & 92.00 & 16.00 & 100.00 & 100.00 & 100.00 & 100.00 & 100.00 & 32.00 \\
\cmidrule{2-13}
& Robot(BG) & 44.00 & 0.00 & 92.00 & 10.00 & 0.00 & 94.00 & 34.00 & 58.00 & 52.00 & 100.00 & 0.00 \\
& Robot(Black) & 69.80 & 6.00 & 100.00 & 78.00 & 18.00 & 100.00 & 98.00 & 98.00 & 100.00 & 100.00 & 0.00 \\
& Robot(Mosaic) & 68.60 & 6.00 & 100.00 & 40.00 & 18.00 & 100.00 & 92.00 & 98.00 & 100.00 & 100.00 & 32.00 \\
\cmidrule{2-13}
& Robot w/o Gripper(BG) & 76.80 & 2.00 & 100.00 & 76.00 & 46.00 & 100.00 & 100.00 & 100.00 & 100.00 & 100.00 & 44.00 \\
& Robot w/o Gripper(Black) & 76.40 & 8.00 & 100.00 & 90.00 & 32.00 & 100.00 & 100.00 & 100.00 & 100.00 & 100.00 & 34.00 \\
& Robot w/o Gripper(Mosaic) & 79.00 & 18.00 & 100.00 & 76.00 & 44.00 & 100.00 & 100.00 & 96.00 & 100.00 & 100.00 & 56.00 \\
\cmidrule{2-13}
& Background(Black) & 57.80 & 16.00 & 100.00 & 20.00 & 16.00 & 100.00 & 40.00 & 58.00 & 100.00 & 100.00 & 28.00 \\
& Background(Mosaic) & 78.00 & 2.00 & 100.00 & 90.00 & 36.00 & 100.00 & 100.00 & 96.00 & 100.00 & 100.00 & 56.00 \\
\midrule

\multirow{15}{*}{OpenVLA}
& Baseline & 74.00 & 53.33 & 93.33 & 80.00 & 43.33 & 93.33 & 80.00 & 63.33 & 93.33 & 83.33 & 56.67 \\
\cmidrule{2-13}
& Target(BG) & 45.00 & 53.33 & 60.00 & 30.00 & 46.67 & 73.33 & 73.33 & 3.33 & 46.67 & 26.67 & 36.67 \\
& Target(Black) & 73.00 & 73.33 & 83.33 & 83.33 & 46.67 & 83.33 & 73.33 & 50.00 & 93.33 & 76.67 & 66.67 \\
& Target(Mosaic) & 67.67 & 50.00 & 80.00 & 73.33 & 63.33 & 80.00 & 76.67 & 43.33 & 100.00 & 66.67 & 43.33 \\
\cmidrule{2-13}
& Gripper(BG) & 38.00 & 33.33 & 26.67 & 33.33 & 3.33 & 86.67 & 63.33 & 50.00 & 20.00 & 43.33 & 20.00 \\
& Gripper(Black) & 69.33 & 50.00 & 73.33 & 83.33 & 13.33 & 90.00 & 73.33 & 80.00 & 86.67 & 86.67 & 56.67 \\
& Gripper(Mosaic) & 63.00 & 20.00 & 60.00 & 63.33 & 30.00 & 100.00 & 70.00 & 80.00 & 96.67 & 63.33 & 46.67 \\
\cmidrule{2-13}
& Robot(BG) & 0.33 & 0.00 & 3.33 & 0.00 & 0.00 & 0.00 & 0.00 & 0.00 & 0.00 & 0.00 & 0.00 \\
& Robot(Black) & 6.00 & 0.00 & 3.33 & 0.00 & 0.00 & 3.33 & 0.00 & 16.67 & 36.67 & 0.00 & 0.00 \\
& Robot(Mosaic) & 8.00 & 0.00 & 0.00 & 3.33 & 0.00 & 40.00 & 10.00 & 6.67 & 20.00 & 0.00 & 0.00 \\
\cmidrule{2-13}
& Robot w/o Gripper(BG) & 25.67 & 10.00 & 63.33 & 13.33 & 0.00 & 50.00 & 0.00 & 23.33 & 66.67 & 30.00 & 0.00 \\
& Robot w/o Gripper(Black) & 26.00 & 13.33 & 70.00 & 30.00 & 0.00 & 26.67 & 10.00 & 16.67 & 76.67 & 16.67 & 0.00 \\
& Robot w/o Gripper(Mosaic) & 38.33 & 6.67 & 70.00 & 53.33 & 10.00 & 80.00 & 30.00 & 13.33 & 66.67 & 53.33 & 0.00 \\
\cmidrule{2-13}
& Background(Black) & 41.67 & 33.33 & 56.67 & 3.33 & 13.33 & 63.33 & 86.67 & 43.33 & 80.00 & 26.67 & 10.00 \\
& Background(Mosaic) & 70.67 & 56.67 & 100.00 & 63.33 & 30.00 & 96.67 & 86.67 & 50.00 & 96.67 & 76.67 & 50.00 \\
\midrule

\multirow{15}{*}{OpenVLA-OFT}
& Baseline & 97.40 & 98.00 & 100.00 & 94.00 & 96.00 & 100.00 & 96.00 & 94.00 & 100.00 & 100.00 & 96.00 \\
\cmidrule{2-13}
& Target (BG) & 49.60 & 98.00 & 82.00 & 68.00 & 38.00 & 72.00 & 0.00 & 4.00 & 34.00 & 18.00 & 82.00 \\
& Target (Black) & 60.20 & 98.00 & 90.00 & 94.00 & 46.00 & 86.00 & 0.00 & 2.00 & 86.00 & 0.00 & 100.00 \\
& Target (Mosaic) & 88.40 & 98.00 & 96.00 & 96.00 & 82.00 & 98.00 & 68.00 & 74.00 & 78.00 & 96.00 & 98.00 \\
\cmidrule{2-13}
& Gripper (BG) & 97.20 & 100.00 & 100.00 & 90.00 & 90.00 & 100.00 & 96.00 & 98.00 & 100.00 & 100.00 & 98.00\\
& Gripper (Black) & 97.00 & 98.00 & 98.00 & 98.00 & 90.00 & 98.00 & 100.00 & 96.00 & 98.00 & 96.00 & 98.00 \\
& Gripper (Mosaic) & 97.60 & 98.00 & 100.00 & 98.00 & 90.00 & 100.00 & 98.00 & 94.00 & 100.00 & 100.00 & 98.00 \\
\cmidrule{2-13}
& Robot (BG) & 67.20 & 94.00 & 50.00 & 96.00 & 16.00 & 100.00 & 70.00 & 60.00 & 18.00 & 92.00 & 76.00 \\
& Robot (Black) & 87.20 & 100.00 & 76.00 & 92.00 & 62.00 & 98.00 & 96.00 & 70.00 & 96.00 & 96.00 & 86.00 \\
& Robot (Mosaic) & 94.60 & 98.00 & 90.00 & 100.00 & 80.00 & 100.00 & 98.00 & 88.00 & 100.00 & 100.00 & 92.00 \\
\cmidrule{2-13}
& Robot w/o Gripper (BG) & 90.60 & 90.00 & 96.00 & 94.00 & 72.00 & 98.00 & 92.00 & 78.00 & 96.00 & 100.00 & 90.00 \\
& Robot w/o Gripper (Black) & 92.60 & 100.00 & 94.00 & 96.00 & 78.00 & 96.00 & 94.00 & 78.00 & 100.00 & 96.00 & 94.00 \\
& Robot w/o Gripper (Mosaic) & 96.40 & 92.00 & 100.00 & 100.00 & 92.00 & 98.00 & 98.00 & 90.00 & 100.00 & 100.00 & 94.00 \\
\cmidrule{2-13}
& Background (Black) & 72.20 & 86.00 & 98.00 & 34.00 & 52.00 & 96.00 & 96.00 & 56.00 & 22.00 & 100.00 & 82.00 \\
& Background (Mosaic) & 97.60 & 98.00 & 100.00 & 96.00 & 94.00 & 100.00 & 100.00 & 90.00 & 98.00 & 100.00 & 100.00 \\

\midrule

\multirow{15}{*}{X-VLA}
& Baseline & 97.20 & 100.00 & 100.00 & 94.00 & 96.00 & 98.00 & 92.00 & 100.00 & 100.00 & 98.00 & 94.00 \\
\cmidrule{2-13}
& Target (BG) & 83.80 & 100.00 & 94.00 & 76.00 & 98.00 & 96.00 & 4.00 & 84.00 & 100.00 & 98.00 & 88.00 \\
& Target (Black) & 84.40 & 100.00 & 100.00 & 94.00 & 90.00 & 92.00 & 0.00 & 86.00 & 100.00 & 88.00 & 94.00 \\
& Target (Mosaic) & 95.60 & 100.00 & 96.00 & 94.00 & 94.00 & 98.00 & 82.00 & 98.00 & 100.00 & 100.00 & 94.00 \\
\cmidrule{2-13}
& Gripper (BG) & 96.20 & 100.00 & 100.00 & 100.00 & 96.00 & 96.00 & 94.00 & 100.00 & 100.00 & 94.00 & 82.00 \\
& Gripper (Black) & 97.40 & 100.00 & 98.00 & 98.00 & 96.00 & 98.00 & 94.00 & 100.00 & 100.00 & 98.00 & 92.00 \\
& Gripper (Mosaic) & 98.40 & 100.00 & 100.00 & 98.00 & 96.00 & 98.00 & 98.00 & 100.00 & 100.00 & 100.00 & 94.00 \\
\cmidrule{2-13}
& Robot (BG) & 55.80 & 28.00 & 78.00 & 80.00 & 34.00 & 72.00 & 6.00 & 100.00 & 100.00 & 52.00 & 8.00 \\
& Robot (Black) & 93.40 & 96.00 & 98.00 & 92.00 & 98.00 & 100.00 & 62.00 & 100.00 & 100.00 & 96.00 & 92.00 \\
& Robot (Mosaic) & 95.20 & 98.00 & 98.00 & 96.00 & 96.00 & 98.00 & 74.00 & 100.00 & 100.00 & 98.00 & 94.00 \\
\cmidrule{2-13}
& Robot w/o Gripper (BG) & 92.40 & 96.00 & 100.00 & 92.00 & 94.00 & 100.00 & 52.00 & 100.00 & 100.00 & 98.00 & 92.00 \\
& Robot w/o Gripper (Black) & 94.00 & 94.00 & 98.00 & 98.00 & 98.00 & 100.00 & 58.00 & 100.00 & 100.00 & 100.00 & 94.00 \\
& Robot w/o Gripper (Mosaic) & 95.80 & 100.00 & 98.00 & 92.00 & 92.00 & 98.00 & 82.00 & 100.00 & 100.00 & 100.00 & 96.00 \\
\cmidrule{2-13}
& Background (Black) & 94.60 & 100.00 & 100.00 & 100.00 & 98.00 & 100.00 & 56.00 & 98.00 & 100.00 & 100.00 & 94.00 \\
& Background (Mosaic) & 98.00 & 100.00 & 100.00 & 100.00 & 94.00 & 98.00 & 92.00 & 100.00 & 100.00 & 100.00 & 96.00 \\

\bottomrule
\end{tabular}
\caption{Success rates (\%) across tasks and masking strategies on \textbf{LIBERO-Goal} for $\pifive$, OpenVLA, OpenVLA-OFT, X-VLA. The instruction is provided in Tab.~\ref{tab:instruction_all}.} \label{tab:mask_goal_merged}
\end{table*}








\begin{table}[htbp]
\centering
\small
\setlength{\tabcolsep}{3pt}
\renewcommand{\arraystretch}{1.5}
\begin{tabular}{lccccc}
\toprule
Setting & Avg & t1 & t2 & t3 & t4 \\
\hline

Baseline & 79.17 & 79.17 & 70.83 & 95.83 & 70.83 \\
\hline

Target(BG) & 27.08 & 50.00 & 8.33 & 50.00 & 0.00 \\
Target(Black) & 44.79 & 50.00 & 33.33 & 95.83 & 0.00 \\
Target(Mosaic) & 72.92 & 66.67 & 70.83 & 87.50 & 66.67 \\ \hline

Gripper(BG) & 78.12 & 79.17 & 75.00 & 91.67 & 66.67 \\
Gripper(Black) & 75.00 & 83.33 & 66.67 & 91.67 & 58.33 \\
Gripper(Mosaic) & 75.00 & 83.33 & 70.83 & 91.67 & 54.17 \\ \hline

Robot(BG) & 76.04 & 75.00 & 70.83 & 95.83 & 62.50 \\
Robot(Black) & 76.04 & 87.50 & 70.83 & 87.50 & 58.33 \\ 
Robot(Mosaic) & 77.08 & 83.33 & 66.67 & 91.67 & 66.67 \\ \hline

Background(BG) & 59.38 & 83.33 & 0.00 & 100.00 & 54.17 \\ 
Background(Black) & 58.33 & 66.67 & 0.00 & 95.83 & 70.83 \\
Background(Mosaic) & 71.88 & 70.83 & 54.17 & 87.50 & 75.00 \\
\bottomrule
\end{tabular}
\caption{Performance of X-VLA under different masking strategies on \textbf{Simpler}.} \label{tab:mask4}
\end{table}

\section{Instruction}
\begin{table*}[t]
\centering
\small
\setlength{\tabcolsep}{4pt}
\renewcommand{\arraystretch}{1.12}
\resizebox{\textwidth}{!}{
\begin{tabular}{llccclll}
\toprule
\textbf{Model} & \textbf{Backbone} & \textbf{Size} & \textbf{Layers} & \textbf{Mask} & \textbf{Decoding} & \textbf{observation} & \textbf{Bench Coverage } \\
\midrule

$\pifive$   & PaliGemma  & 3B    & 18  & Bi-dir  & Flow Matching  & RGB + Lang.            & LIBERO, Calvin \\
X-VLA       & Florence-2 & 0.9B  & 24  & Bi-dir  & Flow Matching  & RGB + Lang. + Soft\_P. & LIBERO, Clv, Simp., Rbt2 \\
OpenVLA     & Prismatic  & 7B    & 32  & Causal  & AutoRegressive & RGB + Lang.            & LIBERO, Clv, Simpler \\
OpenVLA-OFT & Prismatic  & 7B    & 32  & Bi-dir  & L1 Regression  & RGB + Lang.+ Prop.     & LIBERO, Clv, RoboTwin2.0 \\

\bottomrule
\end{tabular}
}
\caption{
Overview of main and supplementary VLA model and simulator coverage.
We conduct the full mechanistic diagnosis on $\pifive$ and OpenVLA using LIBERO, and include X-VLA, OpenVLA-OFT, RoboTwin2.0, CALVIN, and Simpler as supplementary settings for broader validation. 
Lang. = language. LIB. = LIBERO; Cal. = CALVIN; Simp. = Simpler; Rbt2 = RoboTwin2.0.
}

\label{tab:vlas}
\end{table*}


We provide the full set of task instructions used in our benchmark evaluation to make the experimental coverage explicit and reproducible. 
Tab.~\ref{tab:instruction_all} summarizes the instructions for LIBERO, RoboTwin2.0, and SimplerEnv-BridgeV2. 
The LIBERO suites cover complementary manipulation capabilities: LIBERO-10 evaluates long-horizon compositional manipulation, LIBERO-Object focuses on object-centric understanding, LIBERO-Spatial probes spatial reasoning, and LIBERO-Goal evaluates goal-conditioned control. 
For RoboTwin2.0, we evaluate a subset of 5 out of 50 tasks, focusing on representative bimanual manipulation scenarios. 
SimplerEnv-BridgeV2 evaluates real-to-sim transfer on Bridge-style manipulation tasks. 
Tab.~\ref{tab:calvin_tasks} separately lists the CALVIN tasks, which cover sequential multi-skill manipulation, including object rotation, pushing, grasping, lifting, stacking, container interaction, drawer operation, and light control.

\begin{table*}[t]
\centering
\small
\begin{tabular}{lllccc}
\toprule
Model & Stage & Region & Attention Mass & IoU top-10\% & Peak Hit \\
\midrule
\multirow{9}{*}{OpenVLA}
& \multirow{3}{*}{Phase 1}
  & Gripper   & $0.2098 \pm 0.0638$ & $0.2198 \pm 0.0702$ & $0.4903$ \\
& & Robot Arm & $0.2406 \pm 0.0592$ & $0.1006 \pm 0.0480$ & $0.3131$ \\
& & Object    & $0.2039 \pm 0.0741$ & $0.1374 \pm 0.0902$ & $0.2975$ \\
\cmidrule(lr){2-6}
& \multirow{3}{*}{Phase 2}
  & Gripper   & $0.1930 \pm 0.0616$ & $0.2166 \pm 0.0700$ & $0.5148$ \\
& & Robot Arm & $0.2708 \pm 0.0646$ & $0.1209 \pm 0.0468$ & $0.3479$ \\
& & Object    & $0.1965 \pm 0.0748$ & $0.1294 \pm 0.0796$ & $0.3389$ \\
\cmidrule(lr){2-6}
& \multirow{3}{*}{Full}
  & Gripper   & $0.2014 \pm 0.0632$ & $0.2182 \pm 0.0701$ & $0.5026$ \\
& & Robot Arm & $0.2557 \pm 0.0638$ & $0.1108 \pm 0.0485$ & $0.3305$ \\
& & Object    & $0.2475 \pm 0.0863$ & $0.1504 \pm 0.0842$ & $0.3599$ \\
\midrule
\multirow{9}{*}{$\pi_{0.5}$}
& \multirow{3}{*}{Phase 1}
  & Gripper   & $0.2076 \pm 0.1064$ & $0.2650 \pm 0.1395$ & $0.1887$ \\
& & Robot Arm & $0.1812 \pm 0.0863$ & $0.1008 \pm 0.0768$ & $0.0540$ \\
& & Object    & $0.2213 \pm 0.0868$ & $0.1312 \pm 0.0884$ & $0.3798$ \\
\cmidrule(lr){2-6}
& \multirow{3}{*}{Phase 2}
  & Gripper   & $0.1560 \pm 0.0931$ & $0.2057 \pm 0.1324$ & $0.1487$ \\
& & Robot Arm & $0.2042 \pm 0.0649$ & $0.1130 \pm 0.0696$ & $0.0372$ \\
& & Object    & $0.2204 \pm 0.0700$ & $0.1421 \pm 0.0793$ & $0.4401$ \\
\cmidrule(lr){2-6}
& \multirow{3}{*}{Full}
  & Gripper   & $0.1817 \pm 0.1032$ & $0.2353 \pm 0.1392$ & $0.1687$ \\
& & Robot Arm & $0.1927 \pm 0.0772$ & $0.1069 \pm 0.0736$ & $0.0456$ \\
& & Object    & $0.2584 \pm 0.0755$ & $0.1379 \pm 0.0738$ & $0.4206$ \\
\bottomrule
\end{tabular}%
\caption{LIBERO-10 attention grounding for gripper, robot arm, and object regions. All values are pooled over task-step samples. Gripper is the union of robot masks whose instance name contains ``gripper''; Robot Arm is the union of the remaining robot masks; Object is the current subtask focus union. Full is recomputed over the full rollout using the union of both subtask focus sets.}
\label{tab:libero10-robot-parts-object}
\end{table*}

\begin{table*}[t]
\centering
\small
\begin{tabular}{p{0.18\textwidth}p{0.34\textwidth}p{0.40\textwidth}}
\toprule
$\pifive$ evidence & Quantitative observation & Interpretation \\
\midrule
Rollout contrast & Pretrained model: 0\% success and 520-step failures; fine-tuned model: 100\% success and 231-step completion on the representative rollout. & Finetuning changes attention from diffuse scene reading to executable manipulation tracking. \\
View allocation & In the layer-wise attention summary, action queries allocate roughly 19\% mass to the wrist view and 9\% to the agent view, after accounting for the BOS-dominated text sink. & $\pifive$ uses the closer manipulation view more strongly, consistent with reliance on local contact and affordance cues. \\
Layer routing & Early layers L0--L3 mainly perform coarse input reading; visual use increases through L8--L13; L14 shows the strongest image--action coupling; L15--L17 mainly propagate action information. & The visual grounding signal is not monotonically stronger in later layers, but peaks near the layer where image removal is most damaging. \\
Attention sink & BOS self-attention reaches 0.81, and query groups send approximately 28--45\% of their mass to BOS; other text tokens allocate about 70\% mass within the text/BOS loop and only about 28\% to images. & Raw text attention can overstate semantic language use; special tokens must be separated before interpreting instruction grounding. \\
Action slots & The ten action slots have nearly symmetric attention distributions under bidirectional self-attention. & $\pifive$ encodes the action chunk as a coupled block, so visual routing affects the whole action segment rather than a single scalar action. \\
\bottomrule
\end{tabular}
\caption{Additional $\pifive$ attention-routing evidence from the layer-wise visualization analysis. These statistics complement the IoU and mass results by explaining where the attention signal is routed inside the transformer.}
\label{tab:pi05_attention_routing}
\end{table*}

\begin{table*}[t]
\centering
\footnotesize
\renewcommand{\arraystretch}{1.12}

\begin{tabularx}{\textwidth}
{p{2.2cm} p{0.7cm} X p{3.0cm}}
\toprule
\textbf{Task Family} &
\textbf{ID} &
\textbf{Instruction} &
\textbf{Capability} \\
\midrule

\multirow{10}{*}{\makecell[l]{\textbf{LIBERO-10}\\Long-horizon}}
&t0& Put both the alphabet soup and tomato sauce into the basket. & Multi-object\\
&t1& Put both the cream cheese box and butter into the basket. & Multi-object\\
&t2& Turn on the stove and put the moka pot on it. & Sequential reasoning\\
&t3& Put the black bowl into the bottom drawer and close it. & Long-horizon\\
&t4& Put mugs onto corresponding plates. & Spatial planning\\
&t5& Place the book into the back compartment of the caddy. & Manipulation\\
&t6& Put mug on plate and pudding beside it. & Spatial reasoning\\
&t7& Put alphabet soup and cream cheese into basket. & Compositional\\
&t8& Put both moka pots on the stove. & Multi-object\\
&t9& Put mug into microwave and close it. & Long-horizon\\

\midrule

\multirow{10}{*}{\makecell[l]{\textbf{LIBERO-Object}\\Object-centric}}
&t0& Pick up alphabet soup and place it in basket. & Object recognition\\
&t1& Pick up cream cheese and place it in basket. & Object grounding\\
&t2& Pick up salad dressing and place it in basket. & Visual grounding\\
&t3& Pick up BBQ sauce and place it in basket. & Recognition\\
&t4& Pick up ketchup and place it in basket. & Recognition\\
&t5& Pick up tomato sauce and place it in basket. & Recognition\\
&t6& Pick up butter and place it in basket. & Visual identification\\
&t7& Pick up milk and place it in basket. & Category generalization\\
&t8& Pick up chocolate pudding and place it in basket. & Recognition\\
&t9& Pick up orange juice and place it in basket. & Generalization\\

\midrule

\multirow{10}{*}{\makecell[l]{\textbf{LIBERO-Spatial}\\Spatial reasoning}}
&t0& Pick bowl between plate and ramekin onto plate. & Relative position\\
&t1& Pick bowl next to ramekin onto plate. & Spatial reasoning\\
&t2& Pick bowl from table center onto plate. & Localization\\
&t3& Pick bowl on cookie box onto plate. & Spatial grounding\\
&t4& Pick bowl in top drawer onto plate. & 3D localization\\
&t5& Pick bowl on ramekin onto plate. & Relative relation\\
&t6& Pick bowl beside cookie box onto plate. & Spatial grounding\\
&t7& Pick bowl on stove onto plate. & Spatial reasoning\\
&t8& Pick bowl beside plate onto plate. & Relative relation\\
&t9& Pick bowl on cabinet onto plate. & Localization\\

\midrule

\multirow{10}{*}{\makecell[l]{\textbf{LIBERO-Goal}\\Goal-directed}}
&t0& Open the middle drawer of cabinet. & Goal execution\\
&t1& Put bowl on stove. & Goal reaching\\
&t2& Put wine bottle on cabinet. & Manipulation\\
&t3& Open top drawer and put bowl inside. & Multi-step planning\\
&t4& Put bowl on cabinet. & Goal completion\\
&t5& Push plate to front of stove. & Motion control\\
&t6& Put cream cheese into bowl. & Object interaction\\
&t7& Turn on stove. & Goal execution\\
&t8& Put bowl on plate. & Goal reaching\\
&t9& Put wine bottle on rack. & Target reaching\\

\midrule

\multirow{5}{*}{\makecell[l]{\textbf{RoboTwin2.0}\\ Dual-arm}}
&t0& Click the alarm clock's center of the top side button on the table. & Button clicking \\
&t1& Click the bell's top center on the table. & Target clicking \\
&t2& Use both arms to grab the roller on the table. & Bimanual grasping \\
&t3& Use both arms to lift the pot. & Bimanual lifting \\
&t4& Use one arm to press the stapler. & Single-arm pressing \\

\midrule

\multirow{4}{*}{\makecell[l]{\textbf{SimplerEnv}\\Real-to-Sim}}
&t0& Put carrot on plate. & Object placement \\
&t1& Put eggplant into yellow basket. & Goal manipulation \\
&t2& Put the spoon on the towel. & Spatial reasoning \\
&t3& Stack the green block on the yellow block. & Stacking manipulation \\

\bottomrule
\end{tabularx}

\caption{
Complete overview of LIBERO, RoboTwin2.0, Simpler benchmark tasks.
LIBERO-10 evaluates long-horizon compositional manipulation,
LIBERO-Object focuses on object-centric understanding,
LIBERO-Spatial measures spatial reasoning, and LIBERO-Goal evaluates goal-conditioned control.
RoboTwin2.0 subset (5/50 tasks) evaluates bimanual manipulation on five selected tasks.
SimplerEnv-BridgeV2 tests real-to-sim transfer on Bridge-style manipulation tasks.
}
\label{tab:instruction_all}
\end{table*}

\begin{table*}[t]
\centering
\footnotesize
\renewcommand{\arraystretch}{1.12}
\setlength{\tabcolsep}{4pt}

\begin{tabularx}{\textwidth}{p{2.1cm} p{0.8cm} X p{2.9cm}}
\toprule
\textbf{Task Family} & \textbf{ID} & \textbf{Instruction} & \textbf{Capability} \\
\midrule

\multirow{6}{*}{\makecell[l]{\textbf{Block}\\\textbf{Rotation}}}
&t0 & Take the red block and rotate it to the right. & Block rotation \\
&t1 & Take the red block and rotate it to the left. & Block rotation \\
&t2 & Take the blue block and rotate it to the right. & Block rotation \\
&t3 & Take the blue block and rotate it to the left. & Block rotation \\
&t4 & Take the pink block and rotate it to the right. & Block rotation \\
&t5 & Take the pink block and rotate it to the left. & Block rotation \\

\midrule

\multirow{6}{*}{\makecell[l]{\textbf{Block}\\\textbf{Pushing}}}
&t6 & Push the red block to the right. & Object pushing \\
&t7 & Push the red block to the left. & Object pushing \\
&t8 & Push the blue block to the right. & Object pushing \\
&t9 & Push the blue block to the left. & Object pushing \\
&t10 & Push the pink block to the right. & Object pushing \\
&t11 & Push the pink block to the left. & Object pushing \\

\midrule

\multirow{4}{*}{\makecell[l]{\textbf{Slider }\\\textbf{Drawer}}}
&t12 & Push the sliding door to the left side. & Slider manipulation \\
&t13 & Push the sliding door to the right side. & Slider manipulation \\
&t14 & Pull the handle to open the drawer. & Drawer manipulation \\
&t15 & Push the handle to close the drawer. & Drawer manipulation \\

\midrule

\multirow{3}{*}{\makecell[l]{\textbf{Grasp}\\\textbf{\& Lift}}}
&t16 & Grasp and lift the red block. & Grasp and lift \\
&t17 & Grasp and lift the blue block. & Grasp and lift \\
&t18 & Grasp and lift the pink block. & Grasp and lift \\

\midrule

\multirow{3}{*}{\makecell[l]{\textbf{Lift from}\\\textbf{Cabinet}}}
&t19 & Lift the red block from the sliding cabinet. & Lift from container \\
&t20 & Lift the blue block from the sliding cabinet. & Lift from container \\
&t21 & Lift the pink block from the sliding cabinet. & Lift from container \\

\midrule

\multirow{3}{*}{\makecell[l]{\textbf{Lift from}\\\textbf{Drawer}}}
&t22 & Take the red block from the drawer. & Lift from drawer \\
&t23 & Take the blue block from the drawer. & Lift from drawer \\
&t24 & Take the pink block from the drawer. & Lift from drawer \\

\midrule

\multirow{2}{*}{\makecell[l]{\textbf{Place }\\\textbf{Store}}}
&t25 & Store the grasped block in the sliding cabinet. & Place into container \\
&t26 & Store the grasped block in the drawer. & Place into drawer \\

\midrule

\makecell[l]{\textbf{Push}\\\textbf{into Drawer}}
&t27 & Slide the block until it falls into the drawer. & Push into drawer \\

\midrule

\multirow{2}{*}{\makecell[l]{\textbf{Stack }\\\textbf{Unstack}}}
&t28 & Stack the grasped block. & Stacking \\
&t29 & Remove the stacked block. & Unstacking \\

\midrule

\multirow{4}{*}{\makecell[l]{\textbf{Light}\\\textbf{Control}}}
&t30 & Use the switch to turn on the light bulb. & Light control \\
&t31 & Use the switch to turn off the light bulb. & Light control \\
&t32 & Press the button to turn on the LED light. & LED control \\
&t33 & Press the button to turn off the LED light. & LED control \\

\bottomrule
\end{tabularx}

\caption{
Overview of CALVIN manipulation tasks.
CALVIN evaluates sequential multi-skill manipulation abilities, including object rotation,
pushing, grasping, lifting, stacking, container interaction, drawer operation, and light control.
}
\label{tab:calvin_tasks}
\end{table*}

\end{document}